\documentclass{article}

\PassOptionsToPackage{numbers, compress}{natbib}
 \usepackage[preprint]{neurips_2026}

\usepackage[utf8]{inputenc} 
\usepackage[T1]{fontenc}    
\usepackage{hyperref}       
\usepackage{url}            
\usepackage{booktabs}       
\usepackage{amsfonts}       
\usepackage{nicefrac}       
\usepackage{microtype}      
\usepackage{xcolor}         

\usepackage{amsmath}
\usepackage{amssymb}
\usepackage{mathtools}
\usepackage{amsthm}
\usepackage{thm-restate}

\usepackage{xcolor}
\usepackage{comment}
\usepackage{xspace}
\newcommand{\algname}{\texttt{P2P}\xspace}
\usepackage{mathrsfs}
\usepackage{mathtools}
\usepackage{dirtytalk}
\usepackage[breakable]{tcolorbox}
\usepackage{algorithm}
\usepackage{algorithmic}
\usepackage{subcaption}

\usepackage{array}
\theoremstyle{plain}
\newtheorem{theorem}{Observation}[section]
\newtheorem{proposition}[theorem]{Proposition}

\theoremstyle{definition}
\newtheorem{definition}[theorem]{Definition}

\theoremstyle{remark}
\newtheorem{remark}[theorem]{Remark}
\title{Prompts to Proxies: Emulating Human Preferences via a Compact LLM Ensemble}

\author{%
  Bingchen Wang\thanks{Corresponding author.} \\
  Independent Researcher\\
  \texttt{bw2506@columbia.edu} \\
  \And
  Zi-Yu Khoo \\
  AI Singapore \\
  \texttt{ziyu@aisingapore.org} \\
  \And
  Jingtan Wang \\
  School of Computing \\
  National University of Singapore \\
   \texttt{jingtan.w@u.nus.edu} \\
}

\begin{document}

\maketitle

\begin{abstract}
  Large language models are increasingly used as proxies for human subjects in social science research, yet external validity requires matching the response distributions of target human populations. We study population-level survey alignment: reconstructing aggregate survey responses from limited public survey data, without individual-level demographic profiles or model finetuning. We formalize this problem as preference reconstruction: rather than matching proxy agents to demographic profiles, we construct a functional basis of proxy agents and recover population preferences by weighted aggregation. We instantiate this idea via \textit{Prompts to Proxies} (\algname), a two-stage inference-time system. Stage 1 uses structured attribute-based prompting with entropy-guided adaptive sampling to construct a diverse proxy pool spanning the latent preference space. Stage 2 employs L1-regularized regression to select a compact weighted ensemble matching observed target-population responses. Across 14 American Trends Panel waves, \algname achieves an average test MSE of 0.014 at approximately 0.8 USD per survey, improving over prompting and demographic-conditioning baselines. On the World Values Survey, cross-locale transfer experiments show that basis expressiveness can outperform locale matching on average, while locale-specific generation helps on culturally divergent questions. A stress test against an SFT-aligned survey model shows competitive performance using less than 3\% of the training data. These results position preference reconstruction as a lightweight, externally verifiable alternative for survey-based population alignment. Code is available at \url{https://github.com/bingchen-wang/P2P}.
\end{abstract}

\section{Introduction}
Large language models (LLMs) hold considerable promise as proxies for human subjects in social science research. From agent-based modeling of community dynamics~\cite{park2023simulacra, zhao2024competeAI}, to simulating survey responses~\cite{Argyle_Busby_Fulda_Gubler_Rytting_Wingate_2023,park2024generativeagentsimulations1000} and prototyping policy interventions~\cite{hou2025societygenerativeagentssimulate}, synthetic agents afford scalability, reproducibility, and cost advantages unmatched by traditional human samples \cite{horton_homosilicus_2023, NBERw32381}. However, realizing this promise hinges on solving a fundamental validity challenge: the preferences expressed by synthetic agents must faithfully reflect those of the target human population, at least within the topic domain. Without such alignment, conclusions drawn from synthetic experiments lack external validity~\cite{shadish_experimental_2002}, limiting the scientific value of LLM-based social simulation.

This alignment challenge departs from conventional AI safety framing centered on steering a model toward a single set of desirable values, but shares motivation with the emerging notion of pluralistic alignment~\cite{chen2025pal,feng-etal-2024-modular,Sorensen_pluralistic2024}\footnote{See Appendix~\ref{sec:lit} for an extensive review on this topic.}. Here, however, the target is externally verifiable \textit{distributional fidelity}, namely reproducing the heterogeneity of aggregate response patterns observed in a documented target population. In reality, this challenge is compounded by an invisible resource barrier: while well-resourced labs can deploy supervised fine-tuning to train personalized models on large annotated corpora, small academic institutions and independent researchers often face limited compute, limited data, and ethical and practical hurdles that preclude access to sensitive personal information. Existing approaches offer unsatisfying trade-offs: off-the-shelf LLMs trained via supervised fine-tuning (SFT)~\citep{tan-etal-2024-democratizing} and reinforcement learning from human feedback (RLHF)~\citep{bai2022traininghelpfulharmlessassistant, deepRLHF_christiano2017} are easy to deploy but tend to collapse toward central tendencies in their training data, obscuring the preference heterogeneity~\citep{feng-etal-2024-modular,kirk2024understanding,Lee_aligning_2024,slocum2025diverse}; demographic conditioning methods require sensitive personal data and labor-intensive profile construction, yet still fail to guarantee distributional fidelity~\cite{Argyle_Busby_Fulda_Gubler_Rytting_Wingate_2023, castricato_2025_persona, park2024generativeagentsimulations1000, Bisbee_Clinton_Dorff_Kenkel_Larson_2024}.

In this work, we address this gap by introducing \textit{preference reconstruction theory}, a framework that formalizes population preference alignment as a representation learning problem, inspired by revealed preference theory in economics in that preferences are treated as latent and observable only through responses (Section~\ref{sec:theory}). The core claim is that the relevant object is not a demographic match between human and proxy agents, but a response-expressive functional basis—a set of proxy agents whose weighted aggregate responses can reconstruct population-level preferences. We focus on social survey data as the alignment signal, as they are publicly available, transparently documented, and designed to represent well-defined populations, yet typically withholding sensitive demographic information that would enable individual-level profile matching. We operationalize this theory via \algname, a two-stage inference-time system. Stage 1 uses structured attribute-based prompting and entropy-guided adaptive sampling to construct a diverse proxy pool with different response patterns, aimed to span the latent preference space of a topic domain. Stage 2 employs L1-regularized regression to select a compact weighted ensemble of agents whose aggregate responses match those of the target human population. We validate the design through simulation studies and evaluate \algname on the American Trends Panel and World Values Survey, demonstrating robust performance on held-out questions across topics and locales at low cost. Our contributions are as follows:
\begin{itemize}
    \item \textbf{Preference reconstruction formulation.} We formulate population preference alignment as preference reconstruction: constructing a response-expressive proxy basis and recovering population preferences by weighted aggregation.
    \item \textbf{Two-stage inference-time system.}  We instantiate this formulation via Prompts to Proxies (\algname), combining structured attribute-based prompting, entropy-based adaptive sampling and L1-regularized regression, without fine-tuning or individual-level demographic data. 
    \item \textbf{Empirical validation.} Across 14 ATP waves and the WVS, \algname shows 41\% improvement over prompting baselines, competes with an SFT-aligned survey model at less than 3\% of the training data, and suggests that basis expressiveness can matter more than locale matching.
\end{itemize}

\section{Modeling Preferences via Functional Bases}\label{sec:theory}
\paragraph{Alignment as representation learning}
We formalize human-AI preference alignment as a representation learning problem, inspired by the revealed preference theory~\citep{microeconomic_theory,revealed_preference_samuelson}. Let $\mathcal{P}$ denote the latent human preference space. An \textit{individual} preference $p\in\mathcal{P}$ is revealed via a decision-making mechanism $h$ as an observed vector $r \in \mathcal{R}\subset\mathbb{R}^n$. Similarly, let $\widehat{\mathcal{P}}$ denote the latent model preference space. A model preference $\hat{p} \in \widehat{\mathcal{P}}$ manifests through the inference mechanism $f$ into the same response space $\mathcal{R}$. The alignment problem can therefore be abstracted as: $p\xrightarrow{h} r \xrightarrow{f^{-1}} \hat{p}$, where $f^{-1}$ denotes a learning procedure rather than a literal inverse. We use ``revealed'' only in a weak sense: preferences are latent and observable only through the mechanism $h$. The survey responses we study are stated rather than revealed in the economic sense, and no definition or result here invokes rationality or choice-consistency axioms~\citep{revealed_preference_samuelson,microeconomic_theory}.

\begin{restatable}{definition}{defone}
    Individual preference is \textit{revealed-learnable} if $h(\mathcal{P}) \subset \mathrm{conv}(f(\mathcal{\widehat{P}}))$.
\end{restatable}
The convex hull allows for cases where a human preference is not learnable by a single model but can be learned using an ensemble.

\paragraph{Learning population preference via survey data} While the above concerns individual preferences, alignment is population-level. However, for general-purpose LLMs, the relevant population is typically ill-defined: pretraining, supervised finetuning (SFT) and reinforcement learning from human feedback (RLHF) draw on heterogeneous and often undocumented groups. We therefore focus preference alignment on aggregate responses in social surveys, because these surveys are readily available, transparently documented, and designed to represent the views of well-defined target population. Formally, survey responses aggregate individual preference as: $r_\mathrm{pop} = \sum_{i=1}^I w_i h(p_i)$,
where $w_i$ is the population weight of individuals with preference $p_i$ in the population, $I$. This motivates the population analogue of revealed learnability.
\begin{restatable}{definition}{deftwo}
    Population preference is \textit{revealed-learnable} if $\mathrm{conv}\left(h(\mathcal{P})\right) \subset \mathrm{conv}(f(\mathcal{\widehat{P}}))$.\label{def:learnability_pop}
\end{restatable}
By Proposition \ref{prop:iff}, population and individual revealed-learnability are equivalent. Hence, we focus on population preference, while expecting the logic to extend to individual cases.

\paragraph{Three alignment paradigms} 

This formulation distinguishes three paradigms for learning $r_\mathrm{pop}$: The first is a \textit{representative-agent} paradigm: it assumes $\mathrm{conv}(f(\widehat{\mathcal{P}})) = f(\widehat{\mathcal{P}})$ and seeks a single model preference such that $r_\mathrm{pop}=f(\hat{p}_\mathrm{pop})$.\footnote{Appendix~\ref{app:proofs} formalizes this condition and connects it to the representative-agent problem in economics.} The second is \textit{demographic conditioning}~\citep{Argyle_Busby_Fulda_Gubler_Rytting_Wingate_2023, castricato_2025_persona, park2024generativeagentsimulations1000}, which seeks a profile-level map from human preferences $p_i$ to proxy preferences $\hat{p}_i$ using demographic or personal information. The third, which we propose, is \textit{functional-basis reconstruction}, which seeks proxy preferences $\{\hat{p}_j\}_{j=1}^J$ and weights $\{\tilde{w}_j\}_{j=1}^J$ such that $r_\mathrm{pop}= \sum_{j=1}^J\tilde{w}_j f(\hat{p}_j)$. The following observations clarify how these paradigms relate under the this formulation. While simple in form, they show that representative-agent alignment and demographic conditioning impose stronger requirements than functional-basis reconstruction. Short proofs are provided in Appendix~\ref{app:proofs}.

\begin{restatable}{theorem}{obsone}\label{thm:obs1}
    If human population preference can be learned by a representative LLM agent, it can also be learned using a non-trivial ensemble of LLM agents.\footnote{A representative agent is by definition a degenerate ensemble.}
\end{restatable}
\begin{restatable}{theorem}{obstwo}
    If human population preference is revealed-learnable, there could be cases where it can be learned by an ensemble but not by demographic conditioning nor a representative agent.\label{thm:obs2}
\end{restatable}
\begin{restatable}{theorem}{obsthree}
    If human population preference can be learned by demographic conditioning, the ensemble that can learn it is not unique.\label{thm:obs3}
\end{restatable}

\begin{figure}[ht]
    \centering
    \includegraphics[width=\linewidth]{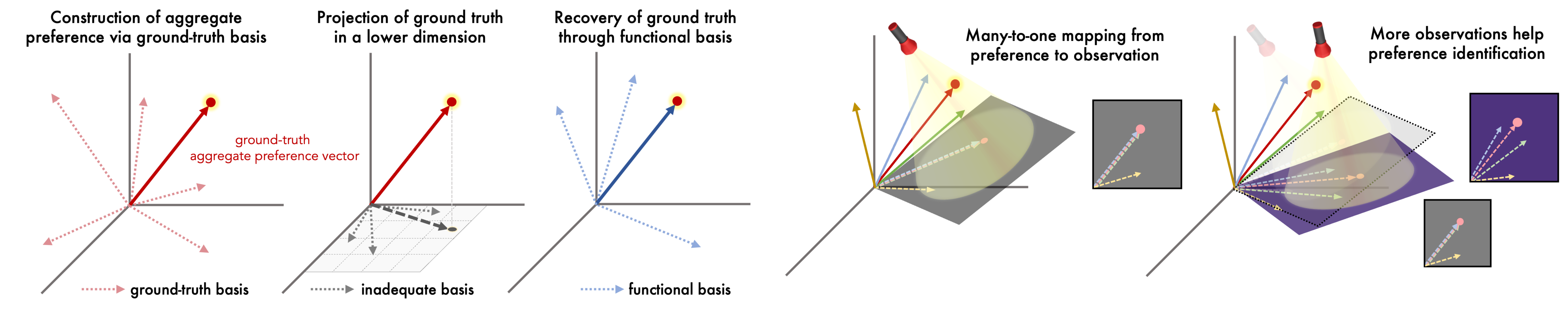}
    \caption{\textbf{Left}: Finding a functional basis to recover latent ground-truth preference. \textbf{Right}: Identification of latent preference through multiple observations.}
    \label{fig:preference_model}
\end{figure}

\paragraph{Preference reconstruction theory} In response to the expressiveness and identification limitations explained in Observations \ref{thm:obs1}-\ref{thm:obs3}, we propose \textit{preference reconstruction theory}: instead of trying to find a one-to-one map between $p_i$ and $\hat{p}_i$, we can construct a response-expressive \textit{functional basis}—a set of proxy LLM agents with distinct revealed response patterns $\{\hat{p}_j\}_{j=1}^J$ spanning the relevant dimensions of human preferences—and then recover the human population preference through appropriate selection and weighting of these agents based on observational data. Figure \ref{fig:preference_model} provides a functional perspective on preference reconstruction theory, together with a geometric interpretation of when the learnability assumption is violated (\textit{inadequate basis}).

\paragraph{Alignment as a two-stage problem}

Let $L$ be a loss function. Preference reconstruction via a functional basis can be formulated as a concrete learning problem:
\begin{equation}
    \min_{\{\hat{p}_j, \tilde{w}_j\}_{j=1}^J} L(r_\mathrm{pop}, \sum_{j=1}^J\tilde{w}_j f(x_0;\hat{p}_j)) \ , \label{eq:pref_recon_obj}
\end{equation}
Directly optimizing (\ref{eq:pref_recon_obj}) is challenging, especially as the dimension $J$ is an optimizer per se.

As a solution strategy, we decompose the problem into two stages with distinct stage goals. The first stage constructs a diverse set of proxy agents, measured by their revealed response variability on training items as the observable signal of preference coverage, \textit{without} access to human-ground-truth distributions. The second stage selects a parsimonious weighted ensemble from the candidate proxy pool, based on the fit between ensemble aggregate responses and human ground truth on training and validation items. Held-out items are used only to assess out-of-sample generalization.

\section{Prompts to Proxies (\algname)}\label{sec:P2P}

\begin{figure*}[!htb]
    \centering
    \includegraphics[width=\textwidth]{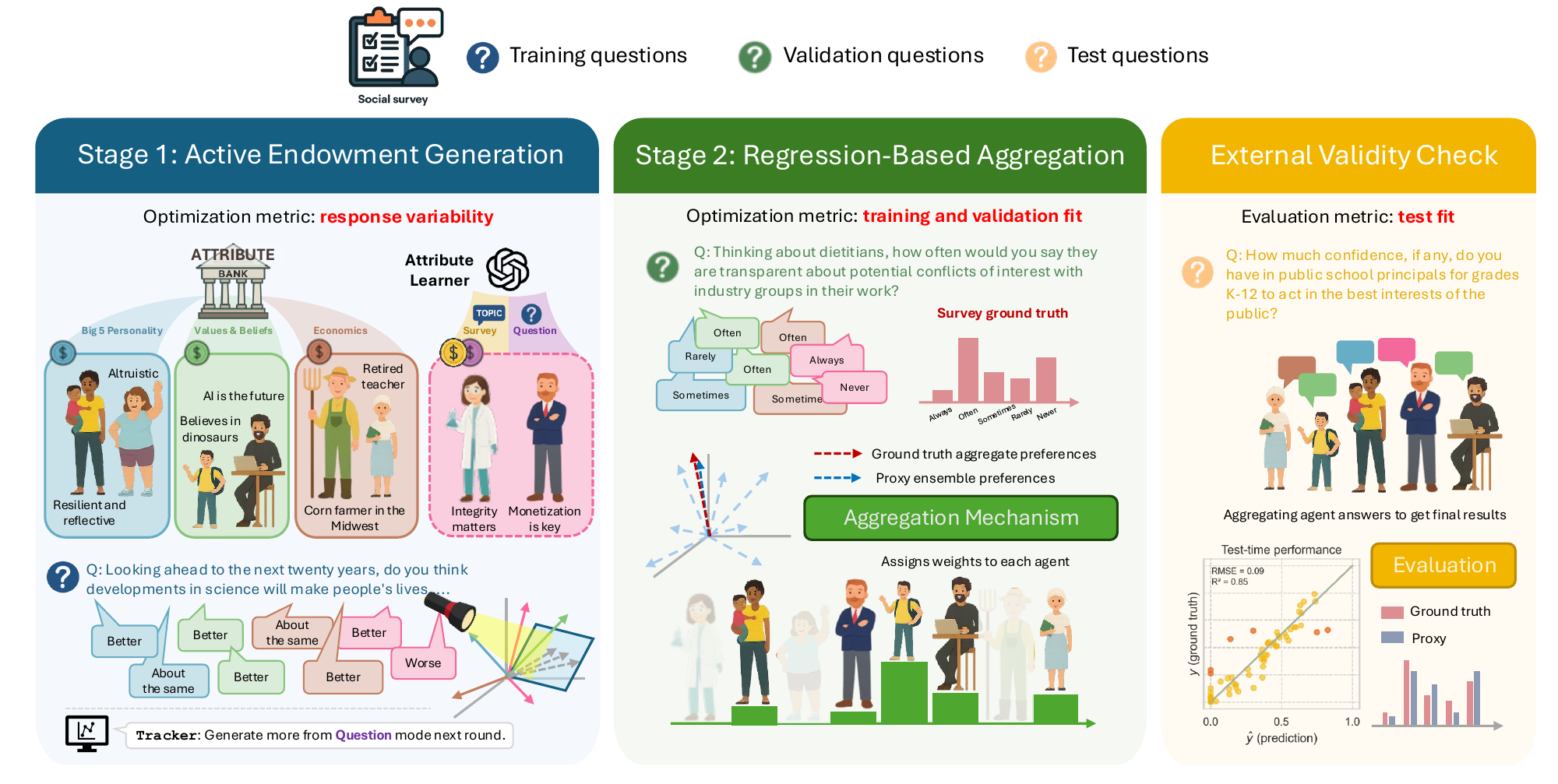}
    \caption{\textbf{System Overview of the Prompts to Proxies (\algname) Framework.} \textbf{Stage 1} constructs a response-expressive proxy pool using a structured attribute bank and entropy-guided adaptive sampling strategies. \textbf{Stage 2} assigns weights to reconstruct population-level preference patterns. Held-out questions afford an external validity check for out-of-sample distributional fidelity.}
    \label{fig:concept illustration}
\end{figure*}

To operationalize preference reconstruction in the survey context, we introduce \textbf{Prompts to Proxies (\algname)}, a modular inference-time alignment system (Figure \ref{fig:concept illustration}). Stage 1, active endowment generation, uses a dynamic attribute bank and entropy-based sampling to construct a response-expressive proxy pool. Stage 2, regression-based aggregation, uses constrained variable selection to recover a compact weighted ensemble matching target-population response distributions.

\subsection{Survey Splitting and Evaluation Protocol}
The survey question set $\mathcal{Q}$ is split into $\mathcal{Q}_\mathrm{train}$, $\mathcal{Q}_\mathrm{valid}$, and $\mathcal{Q}_\mathrm{test}$. Stage 1 constructs the proxy pool using training-question text and proxy responses, without human ground-truth distributions. Stage 2 selects and fits aggregation weights using ground-truth distributions on $\mathcal{Q}_\mathrm{train} \cup \mathcal{Q}_\mathrm{valid}$. $\mathcal{Q}_\mathrm{test}$ is held out to test generalization to unseen questions within the same survey (see Appendix~\ref{sec:generalization} for scoping).

\subsection{Attributes as Control Handles}\label{sec:attributes}
Language models, as next-token predictors, lack intrinsic preferences in the human sense. Their outputs are shaped by statistical associations learned from text corpora and user prompts. Since the prompt space is far too vast to explore in an unstructured manner, we adopt a structured prompting strategy based on \textit{attributes}—interpretable factors, such as ideology, dispositions, or values, whose variation is likely to influence preference expression. \algname treats attributes as \textit{control handles} for steering model behavior.\footnote{This contrasts from the human case, where such attributes are typically viewed as reflections of latent preferences. Rather than inferring attributes from preferences, \algname uses attributes to shape model-expressed preferences.}

\algname constructs endowments, i.e., generated proxy profiles, by drawing attributes from the \textit{attribute bank} and instantiating each attribute with a specific value through a specialized LLM termed \textit{endowment model} (cf. Appendix \ref{sec:endowmentmodel}).\footnote{The use of endowment model ensures the created endowment remains coherent and interpretable to human readers.} The preset attribute bank for \algname lists example attributes taken from different social science disciplines, such as economics and political science. The bank is organized hierarchically: individual attributes are grouped into \textbf{modes}, each representing a coherent subject area (e.g., cognitive biases, political ideology). These modes are further clustered into higher-level \textbf{templates}, which reflect different design purposes or theoretical orientations. Figure \ref{fig:attribute_bank} offers a structured view of the attribute bank and the endowment generation workflow.

While the attribute bank ensures structured coverage, we also enable dynamic discovery through two \textit{freeform templates} that are specific to the survey under analysis. The \textbf{survey} template prompts an LLM \textit{attribute learner} to extract decision-making factors from the overall survey topic. The \textbf{question} template derives attributes for a specific survey question. These attributes are added to the bank only for the survey run. This hybrid approach balances theory- and data-driven methods, enabling human-AI co-discovery of key preference descriptors.

\subsection{Stage 1: Active Endowment Generation}

Section~\ref{sec:theory} requires a response-expressive basis without prescribing how to construct it, and Observation~\ref{thm:obs3} shows that such a basis need not be unique. Prompting can elicit a basis from a single frozen backbone, but freeform prompt engineering is unstructured search, with no parameterization, objective, or update rule. Stage 1 of \algname supplies all three: attributes as the parameterization, response entropy as the objective, and bandit-style adaptive sampling with question patching as the update rule. It begins by invoking an attribute learner to infer survey-specific attributes from the training questions, augmenting the attribute bank. Endowment generation then proceeds via adaptive sampling by mode: After initializing with equal endowments per mode, subsequent steps follow a bandit-style strategy—prioritizing modes that yield high response variability while exploring new modes for under-represented question items (see full Algorithm \ref{alg:P2P_AL} in Appendix \ref{sec:algorithms}).

\paragraph{Tracking response diversity with variability score} To measure preference diversity of agents in a revealed sense, we introduce the notion of \textbf{question entropy}:
\begin{equation}
    H_i(\mathcal{A}) = \frac{-\sum_{k=1}^{K_i} p_{ik} \log_2 p_{ik}}{\log_2 K_i} \ ,
\end{equation}
where $H_i(\mathcal{A})$ denotes the normalized entropy of question $i$ based on responses from a group of agents $\mathcal{A}$, $K_i$ is the number of unique response options $\{1, \dots, K_i\}$ for question $i$, and $p_{ik}$ is the empirical proportion of responses selecting option $k$. The denominator normalizes for scale, ensuring that entropy is comparable across questions with different numbers of choices. A high question entropy indicates greater response diversity for that specific question. By restricting the group $\mathcal{A}$ to agents sampled from a particular mode (subject area), we can compute question entropy at the mode level, termed \textbf{mode-question entropy}, and compare the response variability across modes.

Adaptive sampling references the \textbf{variability score} for each mode. Formally, it is defined as
\begin{equation}
    V(\mathcal{A}_\mathrm{mode}) = \sum_{i \in \mathcal{Q}_\mathrm{train}} \frac{H_i(\mathcal{A}_\mathrm{mode})}{H_i(\mathcal{A}_\mathrm{total})} \ ,
\end{equation}
where $H_i(\mathcal{A}_\mathrm{mode})$ denotes the entropy of responses to question $i$ based on the cumulative set of agents from a given mode up to the current generation step, and $H_i(\mathcal{A}_\mathrm{total})$ is the corresponding question entropy across all agents generated so far.
Computed this way, the variability score favors modes that are capable of generating endowments that elicit diverse responses to questions that other modes typically fail to diversify. In \algname, we task a dedicated class object, \texttt{Tracker}, to compute the variability score for each mode after each generation round (see Algorithm \ref{alg:tracker} in Appendix \ref{sec:algorithms}). The scores are then passed to a softmax function to produce a probability distribution over modes, guiding endowment sampling in the next generation step via a multinomial draw—balancing exploitation of top-performing modes with exploration of under-performing ones.

\paragraph{Patching low-entropy questions} In addition to the existing trade-off between exploitation and exploration, we adopt an additional patching strategy, which we call \textit{question patching}. At the end of each update step, as the \texttt{Tracker} computes the variability scores for each existing mode, it also keeps track of the questions with the lowest entropy values. When question patching is activated, the attribute learner will be called in the next generation step to exclusively infer attributes on each of the lowest-$k$ entropy questions, where $k$ is user-defined. The split of the endowment budget between active generation and patching generation is governed by a user-specified ratio.

To avoid wasting resources on persistently low-entropy questions and to promote cross-pollination between modes, we additionally introduce a \textit{mixed mode} strategy. It is triggered after a question appears $t$ times in the lowest-$k$ entropy list, where $t$ is a user-defined threshold. When in effect, the strategy relocates the assigned endowment budget to a new mixed mode, which combines attributes from the question with those of the current top-performing mode, allowing diversity injection from high-performing modes while maintaining a question-specific focus. Figure \ref{fig:endowments} shows example endowments generated from a real run of \algname, illustrating how different templates and modes yield distinct profiles.

The adaptive sampling cycle continues until the generation budget $N_A$ (an upper bound on the maximum number of agents allowed) is exhausted. By iteratively targeting low-diversity questions, expanding the attribute space and promoting cross-pollination between modes, active endowment generation progressively increases response diversity while preserving coverage over previously explored regions of the preference space (cf. Figure \ref{fig:question_entropy_trajectories} in Appendix \ref{sec:w42_extra_results}).

\subsection{Stage 2: Regression-Based Aggregation}
    With the proxy basis assembled, \algname enters the second stage: regression-based aggregation. The goal at this stage is to estimate aggregation weights to reconstruct the population preference based on the observational data. Depending on the endowment budget $N_A$ and the total number of trainable questions $\mathcal{Q}_\mathrm{train} \cup \mathcal{Q}_\mathrm{valid}$, we could end up in a situation where there are more variables (proxy agents) than observations (trainval questions). Consequently, a variable selection algorithm is vital. Besides, even with $ \vert \mathcal{Q}_\mathrm{train} \cup \mathcal{Q}_\mathrm{valid} \vert \gg N_A$, variable selection allows us to retain a parsimonious agent ensemble, saving inference costs on further simulations. We consider regression methods with an added L1 penalization term and use cross-validation (CV) for hyperparameter tuning. The objective of the learning problem can be stated as
\begin{align}
        &\min_{\mathbf{w}} \sum_{i \in \mathcal{Q}_\mathrm{train}}\sum_{k = 1}^{K_i}\left( r_{ik} - \sum_{j=1}^{N_A} w_j d_{j,ik} \right)^2 + \alpha \Omega(\mathbf{w}) \label{eq:loss} \ , \\
        &\mathrm{s.t.} \quad \sum_{k=1}^{K_i}  \sum_{j=1}^{N_A} w_j d_{j, ik} = 1 \quad \forall i \in \mathcal{Q}_\mathrm{train} \ , \label{eq:constr}
\end{align}
where $\mathbf{w} \in \mathbb{R}^{N_A}$ is the weight vector for the agents, $r_{ik}$ denotes the ground-truth proportion of responses selecting option $k$ for question $i$, $d_{j,ik}$ is a dummy response variable equal to $1$ if agent $j$ selects option $k$ and $0$ otherwise, and $\alpha$ is the hyperparameter for the penalization. When $\Omega$ consists of only an L1 norm, the formulation amounts to constrained lasso. When $\Omega$ subsumes an additional L2 norm, it becomes constrained elastic net. In practice, \algname relaxes the per-question constraint~\eqref{eq:constr} into a quadratic mass penalty with strength $\lambda = 100$, which keeps each question's predicted mass close to one (Appendix~\ref{sec:stage2_config}).

To validate L1-regularized regression as a credible aggregation algorithm, we conduct extensive simulation studies for constrained lasso (Appendix \ref{app:simulation_studies}), where ground truth population preference is constructed using synthetic agents\footnote{This allows us to create a clean environment where alignment result is not affected by the learnability condition (Def. \ref{def:learnability_pop}).}.

Our simulation studies yield three main findings. First, constrained lasso faithfully recovers the ground-truth agents from a large candidate pool under an adequate number of observations, including minority profiles with low weights, and only resorts to reconstruction when observations are extremely scarce (Appendix~\ref{app:simulation_1}). Second, when ground truth agents are removed from the candidate pool, constrained lasso remains effective in selecting a proxy ensemble to recover ground truth even when the number of proxy agents is small (Appendix~\ref{app:simulation_2}). Third, proxy ensembles with insufficient response entropy fail to recover ground-truth preferences, while those with excess entropy show only marginal deterioration, corroborating our choice of response variability as the optimization goal of Stage 1 of \algname (Appendix~\ref{app:simulation_3}). Together, these results provide constructive evidence for \algname's design choices.

\section{Empirical Evaluation}
\label{sec:empiricaleval}

Having validated \algname's design choices through simulation, we now evaluate it on real human survey data to test three core claims: whether active endowment generation improves proxy-basis quality, whether regression-based aggregation recovers target-population response distributions, and whether basis expressiveness can matter more than demographic or locale matching. Our main benchmark is the American Trends Panel (ATP) \citep{pew_ATP}, a nationally representative U.S. survey conducted by Pew Research Center. We obtain the digitized versions of these datasets from the OpinionsQA benchmark \citep{santurkar_opinionsQA_2023}, which repurposed ATP responses to study value pluralism and misalignment in language models. We assess \algname across 14 ATP waves, spanning diverse socio-economic topics. We then test generalizability on the World Values Survey (WVS), analyze cross-locale transfer, conduct ablation studies, and compare against a distribution-calibrated model under topic shift.

\subsection{Experimental Setup} 
For each survey, we split the questions according to a $7:1.5:1.5$ ratio into the training, validation, and test sets. We compare \algname with two ensemble baselines with different endowment generation logic but feed both through the same regression-based aggregation stage as \algname for proxy selection and test set evaluation. This allows the baselines to benefit from the same data-driven selection mechanism, isolating the contribution of \algname's endowment generation strategy (Stage 1). For the \textbf{Vanilla baseline}, endowments are generated using an LLM prompted directly to generate diverse agent profiles conditioned only on the survey topic.\footnote{The endowment generation prompts for \algname and the Vanilla baseline can be found in Appendix~\ref{appendix:prompts}.} For \textbf{PERSONA}, endowments are sampled randomly from the demographically conditioned PERSONA dataset \citep{castricato_2025_persona}, which contains 1000 persona profiles derived from U.S. census data.\footnote{We use random sampling as it preserves the representativeness of the dataset. In Appendix \ref{sec:clustering}, we provide baseline variants using K-means clustering as a pre-processing step. Clustering does not improve performance but increases cost.}

For all methods, we set the endowment generation budget to $300$. For \algname, initial sampling generates $10$ endowments for each mode; active endowment generation is run in 5 update steps with the preset attribute bank, question patching (lowest 3; 75\% endowment budget), mixed-mode generation triggered at $t=2$, and 10 attributes drawn for each endowment generation. Unless otherwise noted, all models use Gemini-2.0-flash as the backend\footnote{See Appendix \ref{sec:backendbaseline} for raw model performance.}, and constrained lasso is used for Stage 2 selection.

\subsection{Main Results: Baseline Comparison on ATP Waves}

\begin{table*}[htp]
\centering
\caption{Performance comparison across 14 ATP waves. Test MSE, entropy, and cost (USD) reported, as means over 3 runs for all methods (sample SDs in Appendix~\ref{sec:addmainresults}). $\Delta$ shows relative test MSE improvement over the best baseline; its average is the mean of the per-wave improvements.}
\label{tab:baseline_comparison_14waves}
\small
\begin{tabular}{llc ccc ccc ccc c}
\toprule
& & & \multicolumn{3}{c}{Vanilla} & \multicolumn{3}{c}{PERSONA} & \multicolumn{3}{c}{\algname} & \\
\cmidrule(lr){4-6} \cmidrule(lr){7-9} \cmidrule(lr){10-12}
Wave & Topic & \# Qs & MSE & Ent. & Cost & MSE & Ent. & Cost\textsuperscript{*} & MSE & Ent. & Cost & $\Delta$ \\
\midrule
W26 & Guns     & 78  & .015 & .50 & \$0.5 & .024 & .37 & \$1.0 & $\mathbf{.010}$ & .55 & \$0.6 & 36\% \\
W27 & Auto     & 96  & .032 & .46 & \$0.6 & .041 & .40 & \$1.3 & $\mathbf{.022}$ & .54 & \$0.7 & 33\% \\
W29 & Gender   & 77  & .026 & .45 & \$0.5 & .030 & .43 & \$1.0 & $\mathbf{.016}$ & .51 & \$0.6 & 41\% \\
W32 & Comm.    & 98  & .019 & .55 & \$0.6 & .025 & .56 & \$1.3 & $\mathbf{.016}$ & .62 & \$0.7 & 17\% \\
W34 & Biomed   & 67  & .021 & .44 & \$0.5 & .018 & .39 & \$0.9 & $\mathbf{.011}$ & .54 & \$0.5 & 39\% \\
W36 & Lead.    & 139 & .033 & .34 & \$0.9 & .028 & .40 & \$1.8 & $\mathbf{.019}$ & .48 & \$1.0 & 31\% \\
W41 & 2050     & 90  & .022 & .44 & \$0.6 & .027 & .48 & \$1.2 & $\mathbf{.013}$ & .55 & \$0.8 & 43\% \\
W42 & Science  & 129 & .029 & .32 & \$0.9 & .036 & .29 & \$1.7 & $\mathbf{.011}$ & .43 & \$1.0 & 61\% \\
W45 & Misinfo  & 95  & .015 & .39 & \$0.6 & .029 & .36 & \$1.2 & $\mathbf{.008}$ & .44 & \$0.7 & 48\% \\
W49 & Privacy  & 98  & .046 & .41 & \$0.6 & .041 & .41 & \$1.3 & $\mathbf{.021}$ & .53 & \$0.7 & 48\% \\
W50 & Family   & 128 & .027 & .36 & \$0.8 & .026 & .36 & \$1.7 & $\mathbf{.013}$ & .44 & \$0.9 & 48\% \\
W54 & Econ     & 116 & .020 & .50 & \$0.8 & .024 & .54 & \$1.5 & $\mathbf{.014}$ & .57 & \$0.9 & 27\% \\
W82 & Global   & 104 & .021 & .50 & \$0.7 & .021 & .49 & \$1.4 & $\mathbf{.008}$ & .57 & \$0.8 & 60\% \\
W92 & Poli.    & 77  & .026 & .58 & \$0.5 & .031 & .57 & \$1.0 & $\mathbf{.017}$ & .61 & \$0.6 & 36\% \\
\midrule
\multicolumn{3}{l}{Average} & .025 & .45 & \$0.7 & .029 & .43 & \$1.3 & $\mathbf{.014}$ & .53 & \$0.8 & 41\% \\
\bottomrule
\end{tabular}

\footnotesize \textsuperscript{*} PERSONA cost only covers the response elicitation stage—higher if persona generation tokens are included.
\end{table*}

Table \ref{tab:baseline_comparison_14waves} shows that \algname consistently improves population-level survey alignment across ATP waves. It achieves an average test MSE of 0.014, compared to 0.025 (Vanilla) and 0.029 (PERSONA)—an average per-wave improvement of 41\% over the best baseline. The improvement is not concentrated in a single topic: 12 of 14 waves achieve mean test MSE below $0.02$, confirming generalization across topics without wave-specific tuning. The comparison also supports the role of Stage 1 as proxy-basis construction. As all methods use the same regression-based aggregation mechanism, the performance gap reflects the quality of the candidate proxy pool. \algname registers higher agent diversity than the baselines, indicated by the higher average entropy (0.53 vs. 0.43-0.45), consistent with the simulation finding that response diversity supports preference reconstruction (Appendix \ref{app:simulation_3}). Notably, this diversity gain is cost-efficient, as \algname runs at approximately \$0.8 per survey, similar to Vanilla (\$0.7) and substantially cheaper than PERSONA (\$1.3, excluding persona generation overhead)\footnote{The PERSONA pipeline \citep{castricato_2025_persona} requires resampling and re-inference, adding further API usage costs.}. Performance comparison is robust under other distributional metrics (1-JSD, EMD) (Appendix~\ref{sec:addmainresults}).

\subsection{Locale Generalization: World Values Survey}
To probe whether \algname generalizes beyond U.S. ATP surveys, we evaluate it on WVS Wave 7 \cite{wvs_round7_v6_2022}, using cleaned data from \citet{cao-etal-2025-specializing}. WVS differs from ATP in survey design, question format, topic coverage, and target populations, making it a useful cross-program test. We focus on three representative regions: the United States, Great Britain, and Hong Kong SAR. Under the same experimental protocol,\footnote{We pass the locale information to the endowment model to ensure regional relevance of the generated endowments.} \algname achieves comparable performance on US and GB (MSE below $0.015$), with weaker but still reasonable results on HK (MSE around $0.024$) (Table \ref{tab:wvs_w7_p2p_performance_concise}).

\begin{table}[ht]
\caption{\algname\ performance on WVS Wave 7 across three locales. Values are mean $\pm$ std over 3 runs. (Full details in Appendix~\ref{sec:addmainresults}.)}
\label{tab:wvs_w7_p2p_performance_concise}
\begin{center}
\small
\begin{tabular}{l | ccc}
\toprule
Metric & US & GB & HK \\
\midrule
Entropy $\uparrow$ & .572 $\pm$ .015 & .571 $\pm$ .004 & .548 $\pm$ .021 \\
Test MSE $\downarrow$    & .015 $\pm$ .001 & .014 $\pm$ .001 & .024 $\pm$ .001 \\
Cost $\downarrow$  & 1.76 $\pm$ .01   & 1.71 $\pm$ .00   & 1.74 $\pm$ .01 \\
\bottomrule
\end{tabular}
\end{center}
\end{table}

\subsection{Basis Expressiveness vs. Locale Matching}\label{sec:erroranalysis}

A central prediction of preference reconstruction is that alignment quality depends on the expressiveness of the proxy basis, not merely on whether proxy profiles are demographically or geographically matched to the target population. The WVS HK setting affords a useful test: HK-native generation is locale-matched, but the US-generated proxy pool has higher entropy. We thus construct HK-transfer by taking US endowments and refitting only the Stage 2 aggregation weights to HK ground-truth responses. We compare HK-transfer with US-native and HK-native.

\begin{figure}[!htb]
    \centering
    \includegraphics[width=0.9\linewidth]{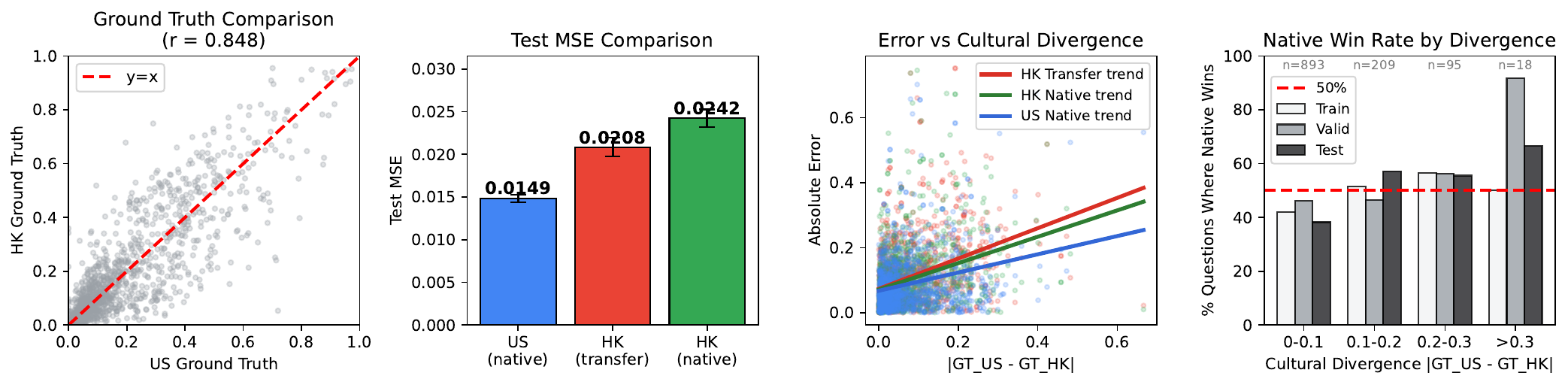}
    \caption{\textbf{Basis Expressiveness vs. Locale Matching on WVS HK.} We compare US-native and HK-native with HK-transfer (US endowments applied to HK via \algname Stage 2). \textbf{(A)} US and HK ground-truth response patterns are correlated but distinct ($r=0.848$). \textbf{(B)} HK-transfer outperforms HK-native in overall performance, corroborating the functional basis perspective. \textbf{(C)} Error gap scales with US-HK ground-truth divergence, suggesting learnability limits (Definition \ref{def:learnability_pop}). \textbf{(D)} Despite higher test MSE, HK-native wins more often on culturally divergent questions, indicating that localization helps in tail cases.}
    \label{fig:wvs_4panel_main}
\end{figure}

Figure \ref{fig:wvs_4panel_main} presents four complementary analyses. We define cultural divergence for each question-option pair as $\vert \mathrm{GT}_\mathrm{US} - \mathrm{GT}_\mathrm{HK} \vert$, the absolute difference in response proportions between populations. Panel (A) shows correlated but distinct US and HK ground-truth response patterns ($r=0.848$), with variation around the diagonal confirming cultural differences. Panel (B) shows that HK-transfer achieves \textit{lower} test MSE than HK-native, constituting a key empirical evidence for the functional basis view—a more diverse agent pool facilitates preference reconstruction, even when generated for a different locale. Panels (C) and (D) qualify this result. Panel (C) shows that reconstruction error scales with cultural divergence: question-option pairs where US and HK responses differ more are generally harder to reconstruct under both transfer and native generations. This makes cultural divergence an observable diagnostic for the learnability limits (Definition \ref{def:learnability_pop}). Moreover, Gemini-based ensembles exhibit a clear bias toward US ground truth, consistent with prior evidence that LLMs better represent Western public opinion~\cite{durmus2024measuringrepresentationsubjectiveglobal}. Panel (D) shows that HK-native often outperforms HK-transfer on culturally divergent questions. Hence, basis expressiveness dominates on average, but locale-specific generation helps in tail items where the target population genuinely differs. This decomposition separates three factors often conflated in demographic-conditioning approaches—basis expressiveness, model-backend cultural skew, and target-population distinctiveness—and demonstrates \algname's potential for model steerability and bias research (cf. Appendix \ref{sec:futurework}).

\subsection{Ablation Studies}
We conduct ablation on ATP Wave 42 to isolate component contributions (details in Appendix \ref{sec:app_abl}). 

\paragraph{Attributes}
Disabling question patching reduces entropy from 0.445 to 0.399 and worsens test MSE from 0.009 to 0.011, while disabling both freeform templates (question \& survey) further degrades MSE to 0.013. This suggests that human-AI co-discovery of preference descriptors yields meaningful performance gains. Varying the per-endowment attribute cap (10 vs. 20) has minimal effect.

\paragraph{Endowment budget} 
Performance improves with budget but plateaus after 3 update steps (170 agents). A no-update baseline (110 agents) underperforms, validating the benefit of adaptive sampling.

\paragraph{Model backend}
We test \algname's performance on six model backends: GPT-4.1-mini, GPT-4.1-nano, Gemini-2.0-flash, Gemini-2.5-flash-lite, DeepSeek-V4-Flash and a locally hosted Qwen. GPT-4.1-mini achieves the best entropy (0.471) and Gemini-2.0-flash the best MSE (0.0087 with lasso) at one-quarter the cost of GPT-4.1-mini. Smaller models (GPT-4.1-nano and Gemini-2.5-flash-lite) perform worse, consistent with our intuition that limited model capacity constrains the learnability (Definition~\ref{def:learnability_pop}).

\paragraph{Regression-based aggregation}

Replacing regression with simple averaging of agent responses increases test MSE from 0.0087 to 0.0332, confirming that regression is essential for learning population preferences, not just ensemble compression. Switching loss weighting from option level to question level has minimal effect on W42 results.

\paragraph{Stage 2 configuration}
We ablate the mass penalty on the full 14-wave ATP panel and probe two anchors that let agents explain deviations from a reference rather than full distributions: a uniform reference respondent and an intercept (Appendix~\ref{sec:stage2_config}). Removing the mass penalty leaves test MSE statistically unchanged but lowers 1-JSD by 0.0075. Both anchors reduce test MSE (from 0.0142 to 0.0130 and 0.0128) and raise 1-JSD, but neither improves EMD.

\subsection{Stress Test: Comparison with an SFT-Aligned Model under Topic Shift}
\algname targets limited-data and limited-compute regimes, so our main baselines focus on methods under similar constraints. Supervised finetuning (SFT), however, is a natural reference when substantially more data and training access are available. We therefore stress-test \algname against \citet{cao-etal-2025-specializing}, who train a distribution-calibrated model on pooled WVS data from 46 countries (approx. 8,400 training entries) to predict answer distributions conditioned on locale.

We follow their topic-shift protocol: training on general attitudinal and religious-ethical items (Q1+Q2) and testing on political-cultural items (Q3). We run \algname separately on the United States, Great Britain, and Hong Kong SAR, using only 172-185 local training entries and no finetuning. Despite using less than 3\% of the training data and no SFT, \algname achieves comparable Jensen-Shannon Divergence (1-JSD) (0.71-0.75 vs. 0.78), with matching or better majority-class prediction accuracy (MCPA) (0.49-0.64 vs. 0.43). We interpret this as evidence that some preference structure transfers across topics though we do not assume such transfer is guaranteed. Studying cross-topic generalization remains an important direction for future work (Appendix~\ref{sec:futurework}). Details on the stress test are presented in Appendix~\ref{sec:stresstest}. We additionally provide a comparison with a second SFT-aligned model (\citet{suh-etal-2025-language}) on the subgroup prediction task in Appendix~\ref{sec:subpop}. A further comparison of \algname with the two SFT-aligned models can be found in Appendix~\ref{sec:related_comparison}.

\section{Use Cases}\label{sec:usecases}
Since \algname generalizes to unseen questions within a survey rather than across surveys (Section~\ref{sec:P2P}; Appendix~\ref{sec:generalization}), a natural question is what practical uses the method serves. We propose two in this section, one diagnostic (zero-demographic benchmark) and one constructive (retrospective imputation of unasked questions). Together, they bound the method's claims from both sides.

\paragraph{Zero-demographic benchmark against overclaiming} Existing preference alignment paradigms center around building ``digital twins'' (demographic conditioning) or a large one-size-fits-all model (SFT/RLHF). \algname instead recovers the target using a proxy basis with no real individual profiles, no demographic marginals, and no fine-tuning. Its held-out score is hence a principled reference, indicating how much of the observed response pattern is recoverable \textit{without} demographic or private information. Methods that add such information—incurring data, privacy, and compute costs—should be expected to beat it to justify the heavier arsenal. In this regard, \algname doubles as a falsifiability check against the assumption that finer conditioning or greater data volume improves emulation. Requiring only API inference, public survey data, and CPU regression, \algname also lowers the barrier to entry for preference alignment research (Appendix~\ref{sec:broader-impacts}).

\paragraph{Retrospective imputation of unasked questions} When a new survey introduces questions absent from earlier waves, shared questions can calibrate a synthetic-agent basis to the historical population, with agents supplying counterfactual response distributions for the previously unasked questions. This supports longitudinal and regression analyses on series that would otherwise begin only at the later wave or suffer from missing data by design. The machinery parallels synthetic-control estimation~\citep{abadie2003economiccosts, Abadie2010syntheticcontrol}, though here the donors are constructed LLM agents rather than pre-existing real units. In a masked missing-question experiment on WVS Waves 6 and 7 (Appendix~\ref{sec:temporal}), the drift refit, which fits the between-wave change, consistently outperforms persistence (W7 marginals carried back unchanged) at every calibration budget. At 72 anchors, the ridge drift refit reduces MSE by 23.3\% on masked-train questions and 12.1\% on test questions.

\section{Conclusion}
We formalize preference alignment as preference reconstruction: constructing a functional basis of proxy agents and recovering aggregation weights through supervised learning. \algname operationalizes this idea via entropy-based adaptive sampling and regression-based aggregation. Across simulation studies, ATP waves, WVS locales, and an SFT stress test, the results show that compact LLM ensembles can recover population-level survey preferences under limited-data and limited-compute constraints. Future work can extend \algname to freeform textual responses, model steerability benchmarks, non-stationary preferences, and semantic or psychometric checks (Appendix~\ref{sec:futurework}).

\begin{ack}
We thank Bruno Abrahao, Jennifer Castle, Zhiliang Chen, David Hendry, Gregory Lau, Xiaoqiang Lin, Bryan Low and S\'{e}verine Toussaert for helpful feedback and discussions, and the anonymous reviewers of this and earlier versions for their constructive comments.

This work was primarily self-funded by the first author, with additional support from Google Cloud free trial credits. The GPT-4 and Qwen experiments reported in the appendices benefited from computational resources provided by the GLOW.AI group.

The views and interpretations expressed in this paper are solely those of the authors. The authors declare no competing interests.
\end{ack}

\bibliographystyle{plainnat}

\bibliography{neurips2026/neurips_2026}

\newpage
\appendix
\section{Modeling Preferences via Functional Bases: Proofs}\label{app:proofs}

\setcounter{theorem}{-1}

\begin{proposition}[Existence of a representative agent]
    Let $\widetilde{\mathcal{P}}$ be a latent preference space (regardless of human or model), with an individual preference denoted by $\tilde{p} \in \widetilde{\mathcal{P}}$. Let $g$ be the map from $\widetilde{\mathcal{P}}$ to the observed response space $\widetilde{\mathcal{R}} \subset \mathbb{R}^n$. Then the following two statements are equivalent:
    \begin{enumerate}
        \item The image $g(\widetilde{\mathcal{P}}) \subset \widetilde{\mathcal{R}}$ is convex.
        \item For any aggregate response $\tilde{r}_\mathrm{pop} = \sum_{i=1}^I w_i \tilde{r}_i$ with $\{w_i\}_{i=1}^I$ convex weights and $\tilde{r}_i = g(\tilde{p}_i)$, there exists $\tilde{p}_\mathrm{pop} \in \widetilde{\mathcal{P}}$ such that $g(\tilde{p}_\mathrm{pop}) = \tilde{r}_\mathrm{pop}$.
    \end{enumerate}
    \label{thm:background}
\end{proposition}
\begin{proof}
    First, we prove $(1) \implies (2)$.

    By definition, $\tilde{r}_i = g(\tilde{p}_i) \in g(\widetilde{\mathcal{P}})$ for all $i \in \{1, \dots, I\}$. Since $\{w_i\}_{i=1}^I$ are convex weights ($\sum_{i=1}^I w_i = 1$ and $w_i \geq 0$ for all $i\in\{1,\dots, I\}$), and $g(\widetilde{\mathcal{P}})$ is convex, it follows that
    \begin{equation*}
        \sum_{i=1}^I w_i \tilde{r}_i \in g(\widetilde{\mathcal{P}}).
    \end{equation*}

    By the definition of image, it follows that there exists $\tilde{p}_\mathrm{pop} \in \widetilde{\mathcal{P}}$ such that
    \begin{equation*}
        g(\tilde{p}_\mathrm{pop}) = \sum_i^I w_i \tilde{r}_i  \triangleq \tilde{r}_\mathrm{pop}.
    \end{equation*}

    Claim: $(2) \implies (1)$.

    Let $\{\tilde{r}_i\}_{i=1}^I$ denote any finite collection of observed responses with $\tilde{r}_i \in g(\widetilde{\mathcal{P}})$, and $\{\lambda_i\}_{i=1}^I$ be any convex weights. Consider $\tilde{r}_\mathrm{agg} = \sum_{i=1}^I\lambda_i \tilde{r}_i$. By (2), there exists $\tilde{p}_\mathrm{agg} \in \widetilde{\mathcal{P}}$ such that
    \begin{equation}
        \tilde{r}_\mathrm{agg} = g(\tilde{p}_\mathrm{agg}) \in g(\widetilde{\mathcal{P}}).
    \end{equation}

    Therefore, $g(\widetilde{\mathcal{P}})$ is convex by definition.
\end{proof}

\begin{remark}[Sufficient condition for representative agent]
If the decision-making mechanism $g$ maps a preference into a convex subset of a probability simplex $\Delta^{n-1}$, then any aggregate of individual responses is realizable by a representative agent. 
\end{remark}
\begin{proof}
    A convex subset of a probability simplex is by definition convex. The result follows from the theorem.
\end{proof}
The implications of this lemma are two-fold:
\begin{enumerate}
    \item In human decision-making, the condition that $g(\widetilde{\mathcal{P}})$ is convex is less likely to be true unless we assume stochastic choice models with stringent functional forms \citep[cf.][]{JEP,JacksonYariv2019}.
    \item In LLM decision-making,  the condition that $g(\widetilde{\mathcal{P}})$ is convex is more likely to hold is more likely to be hold assuming that the model has considerable steerability. (E.g., through finetuning or prompting, we can effectively change the latent model preference embeddings and influence the output probability distributions.)
\end{enumerate}

\defone*
\begin{definition}
    Human preference is \textit{strongly revealed-learnable} if  $h(\mathcal{P}) \subset f(\mathcal{\hat{P}})$.
\end{definition}
\deftwo*
\begin{definition}
    Human population preference is \textit{strongly revealed-learnable} if $\mathrm{conv}\left(h(\mathcal{P})\right) \subset f(\mathcal{\hat{P}})$.
\end{definition}
\begin{proposition}
    Human population preference is revealed-learnable if and only if individual human preference is revealed-learnable. \label{prop:iff}
\end{proposition}
\begin{proof}
    The ``only if'' part follows directly from $h(\mathcal{P}) \subset \mathrm{conv}(h(\mathcal{P}))$. Here we focus on the ``if'' part.

    By definition, a human population preference is defined as
    \begin{equation*}
    r_\mathrm{pop} = \sum_{i=1}^I w_i h(p_i). 
    \end{equation*}
    Since individual human preference is learnable in a revealed sense, i.e., $h(\mathcal{P}) \subset \mathrm{conv}(f(\mathcal{\widehat{P}}))$, for any $p_i$, there exists $\{w_{ik}, \hat{p}_{ik}\}_{k=1}^{K_i}$ such that
    \begin{equation*}
        h(p_i) = \sum_{k=1}^{K_i} w_{ik} f(\hat{p}_{ik}).
    \end{equation*}
    Plug it back into the definition of human population preference and we get
    \begin{equation*}
    r_\mathrm{pop} = \sum_{i=1}^I w_i h(p_i) = \sum_{i=1}^I w_i\sum_{k=1}^{K_i} w_{ik} f(\hat{p}_{ik}) = \sum_{i=1}^I \sum_{k=1}^{K_i} w_i w_{ik} f(\hat{p}_{ik}) := \sum_{\tilde{i}=1}^{\tilde{I}} w_{\tilde{i}} f(\hat{p}_{\tilde{i}}). 
    \end{equation*}
    The last step follows from re-indexing. Therefore, $\mathrm{conv}(h(\mathcal{P})) \subset \mathrm{conv}(f(\mathcal{\widehat{P}}))$.
\end{proof}


\obsone*

\begin{proof}
    By Proposition \ref{thm:background}, if human population preference can be learned by a representative LLM agent, then
    \begin{equation*}
         \mathrm{conv}( h(\mathcal{P})) \subset \mathrm{conv}( f(\mathcal{\widehat{P}})) = f(\mathcal{\widehat{P}}). 
    \end{equation*}
   For any $r_\mathrm{pop} \in \mathrm{conv}\left(h(\mathcal{P})\right)$, it follows from the definition of human population preference that
    \begin{equation*}
        r_\mathrm{pop} = \sum_{i=1}^I w_i h(p_i) := \sum_{i=1}^I w_i r_i.
    \end{equation*}
    Since $r_i \in h(\mathcal{P}) \subset \mathrm{conv}\left(h(\mathcal{P})\right) \subset f(\widehat{\mathcal{P}})$, there exists $\hat{p}_i$ such that $f(\hat{p}_i)=r_i$.

    It follows that
    \begin{equation*}
        r_\mathrm{pop} = \sum_{i=1}^I w_i f(\hat{p}_i) \in \mathrm{span}\left\{f(\hat{p}_i)\right\}_{i=1}^I.
    \end{equation*}
    In words, we can construct an ensemble where each agent represents in preference a ground-truth individual in the human population (\textit{demographic conditioning}). 
    
    NB: Here we implicitly assume that the human population consists of more than one individual with distinct preferences. If the population consists of only 1 individual, then the representative agent and the ensemble coincide.    
\end{proof}


\obstwo*

\begin{proof}
    Based on the proof of Observation \ref{thm:obs1} and Proposition \ref{thm:background}, we see that demographic conditioning hinges on the condition that $h(\mathcal{P}) \subset f(\widehat{\mathcal{P}})$, whereas representative agent relies on an even stronger condition that  $\mathrm{conv}(h(\mathcal{P})) \subset f(\widehat{\mathcal{P}})$. If human population preference is revealed-learnable, we only have the weaker condition\footnote{Equivalently, $h(\mathcal{P}) \subset \mathrm{conv}(f(\widehat{\mathcal{P}}))$ by Proposition \ref{prop:iff}.}:
    \begin{equation}
        \mathrm{conv}(h(\mathcal{P})) \subset \mathrm{conv}(f(\widehat{\mathcal{P}})). \label{cond:learnability}
    \end{equation}

    Consider the case where $h(\mathcal{P}) \not\subset f(\widehat{\mathcal{P}}) $. I.e., there exists $p_i \in \mathcal{P}$ such that for \textit{all} $\hat{p}_j \in\mathcal{\widehat{P}}$, $h(p_i) \neq f(\hat{p}_j)$. Then both demographic conditioning and representative agent fail. However, condition (\ref{cond:learnability}) implies that for any $r_\mathrm{pop} \in \mathrm{conv}\left(h(\mathcal{P})\right)$ there exists $\{\tilde{w}_{j}, \hat{p}_{j}\}_{j=1}^{J}$ such that
    \begin{equation*}
        r_\mathrm{pop} =  \sum_{j=1}^{J} \tilde{w}_{j} f(\hat{p}_{j}).
    \end{equation*}
    In other words, human population preference can be learned using an ensemble.
\end{proof}


\obsthree*

\begin{proof}
    If human population preference can be learned by demographic conditioning, we know from the proof of Observation \ref{thm:obs1} that $h(\mathcal{P}) \subset f(\widehat{\mathcal{P}})$ and that for any $r_\mathrm{pop} \in \mathrm{conv}(h(\mathcal{P}))$ there exist $\{\hat{p}_i\}_{i=1}^I$ such that
    \begin{equation}
          r_\mathrm{pop} = \sum_{i=1}^I w_i r_i= \sum_{i=1}^I w_i f(\hat{p}_i). \label{eq:gt_ensemble}
    \end{equation}
    However, since $r_1 \in h(\mathcal{P}) \subset \mathrm{conv}(h(\mathcal{P}))$, we know there exists a finite set of $\{r^*_k\}_{k=1}^K$ with ${r}_k^* \in h(\mathcal{P})$ and convex weights $\{\lambda_k\}_{k=1}^K$ such that
    \begin{equation*}
        r_1 = \sum_{k=1}^K \lambda_k {r}_k^*.
    \end{equation*}
    Plug this into (\ref{eq:gt_ensemble}) and we get
    \begin{equation*}
        r_\mathrm{pop} = \sum_{k=1}^K w_1\lambda_k {r}_k^* + \sum_{i=2}^I w_i r_i.
    \end{equation*}
     Since ${r}_k^* \in h(\mathcal{P}) \subset f(\widehat{\mathcal{P}})$, there exists $\hat{p}_k^*$ such that $f(\hat{p}_k^*)={r}_k^*$.
     
     In conclusion,
     \begin{equation*}
         r_\mathrm{pop} \in \mathrm{span}\{f(\hat{p}_1^*), \dots, f(\hat{p}_K^*), f(\hat{p}_2), \dots, f(\hat{p}_I)\}.
     \end{equation*}
\end{proof}

\clearpage
\section{Prompts to Proxies (\algname): Implementation Details}\label{app:details}

\subsection{Attribute Bank}\label{sec:att_bank}
\begin{figure*}[!htp]
    \centering
    \includegraphics[width=\textwidth]{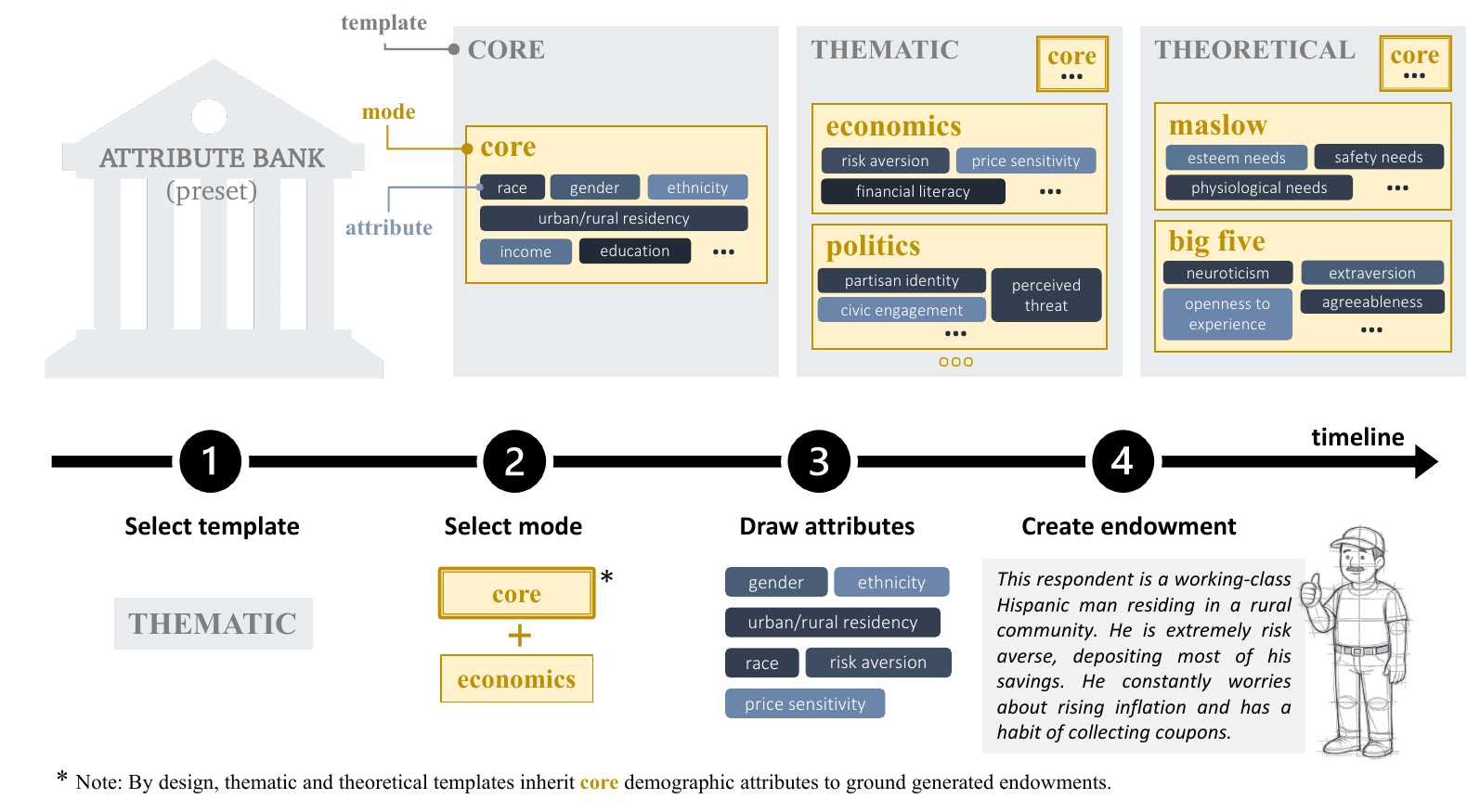}
    \caption{\textbf{Attribute Bank and Endowment Generation Workflow.} The preset attribute bank is organized hierarchically by template (core, thematic, theoretical), mode (e.g., economics, politics, maslow, big five), and attribute (e.g., race, gender, ethnicity, urban/rural residency, risk aversion). Endowment is generated by a specialized LLM (\texttt{EndowmentModel}), which takes drawn attributes as input and outputs a profile with instantiation of these attributes.}
    \label{fig:attribute_bank}
\end{figure*}

For the preset attribute bank (Figure \ref{fig:attribute_bank}), we define three primary templates. The \textit{core} template contains general demographic descriptors and serves as the default baseline. The \textit{thematic} template includes topic-specific modes such as economics, politics, culture and history. The \textit{theoretical} template groups attributes derived from canonical frameworks such as Maslow's hierarchy of needs~\citep{maslow1943motivation} and the Big Five personality traits~\citep{john_big_1999}. These templates and modes are not fixed: they serve as a structured starting point but are fully extensible. Users can define new templates, introduce domain-specific modes, or modify existing attribute sets to suit different experimental contexts. By sampling and varying attributes from this bank, we create prompts that instantiate proxy agents at different locations of the latent model preference space.

\subsection{Model Prompts}~\label{appendix:prompts}

\subsubsection{Attribute Learner (\texttt{AttributeLearner})}

To generate attributes from a single question:
\begin{tcolorbox}[breakable, colback=gray!5, colframe=gray!75, title=Attribute Learner: Learning Attributes from A Single Question, fontupper=\ttfamily]

You are an intelligent research assistant trained to analyze individual survey questions and infer which human attributes might influence how different people respond.
\par\medskip
You are given a single survey question. Your task is to propose a list of relevant human attributes—such as demographics, beliefs, values, personality traits, or ideological leanings—that are likely to shape responses to this question.
\par\medskip
Focus on underlying factors that would cause meaningful variation in answers across different types of people. Avoid generic or overly broad attributes.
\par\medskip
Respond **only** with a Python-style list of double-quoted strings. Do not include any explanation, headers, or prose before or after the list.
\par\medskip
**Example output format:**
["religious affiliation", "political ideology", "trust in government"]
\end{tcolorbox}

To generate attributes from a set of questions:
\begin{tcolorbox}[breakable, colback=gray!5, colframe=gray!75, title=Attribute Learner: Learning Attributes from A Set of Questions, fontupper=\ttfamily]

You are an intelligent research assistant trained to analyze survey questions and infer which human attributes might influence how individuals respond.
\par\medskip
You are given a set of training-only survey questions. Your task is to propose a list of relevant human attributes—such as demographics, beliefs, values, personality traits, or ideological leanings—that are likely to affect responses to these questions.
\par\medskip
Carefully analyze the content and framing of the questions. Identify underlying factors that might shape how different people respond. Focus on attributes that are salient, discriminative, and potentially variable across respondents.
\par\medskip
Respond **only** with a Python-style list of double-quoted strings. Do not include any explanation, headers, or prose before or after the list.
\par\medskip
**Example output format:**
["attribute1", "attribute2", "attribute3", "attribute4"]

\end{tcolorbox}

\subsubsection{Endowment Model (\texttt{EndowmentModel})} \label{sec:endowmentmodel}
System prompt:
\begin{tcolorbox}[breakable, colback=gray!5, colframe=gray!75, title=Endowment Model: System Prompt, fontupper=\ttfamily]

You are an expert assistant trained to generate realistic, diverse, and demographically plausible personas for social science surveys.
\par\medskip
Each persona should include:
\par\medskip
- `eid`: a short, lowercase, variable-safe identifier that encodes key traits (e.g., urban\_liberal\_30s\_female). No punctuation or spaces.
\par\medskip
- `endow\_text`: a natural language description of the persona (1-2 sentences), written as if describing a survey respondent.
\par\medskip
Instructions:
\par\medskip
- Represent a wide range of age, gender, race, education, region, and political ideology.
\par\medskip
- Avoid repetition of phrasing or demographic combinations across personas.
\par\medskip
- Do not include explanations or formatting outside of the JSON array.
\end{tcolorbox}

To generate endowments based on attributes:
\begin{tcolorbox}[breakable, colback=gray!5, colframe=gray!75, title=Endowment Model: Endowment Generation Based on Attributes, fontupper=\ttfamily]

Generate \{n\} diverse persona(s) that vary meaningfully along the following attributes: \{attr\_string\}.
\par\medskip
Each persona should reflect a distinct combination or value expression of these traits.
\par\medskip
Return a JSON array of dictionaries, each with:
\par\medskip
- `eid`: a short, lowercase, variable-safe identifier
\par\medskip
- `endow\_text`: a brief natural-language description of the persona
\end{tcolorbox}

To generate endowments based on survey topics (also used for Vanilla baseline):
\begin{tcolorbox}[breakable, colback=gray!5, colframe=gray!75, title=Endowment Model: Endowment Generation Based on A Survey Topic, fontupper=\ttfamily]

Generate \{n\} diverse personas for a survey experiment.
\par\medskip
Each persona must include:
\par\medskip
- eid: short, lowercase identifier
\par\medskip
- endow\_text: a short natural-language description
\par\medskip
The survey topic is: \{topic\}
\par\medskip
Ensure diversity across demographics.
\par\medskip
Return the result as a JSON array.

\end{tcolorbox}

\subsubsection{Survey Conductor (\texttt{SurveyConductor})}

To use an agent to answer a survey question:
\begin{tcolorbox}[breakable, colback=gray!5, colframe=gray!75, title=Survey Conductor: Answering A Survey Question, fontupper=\ttfamily]
You are completing a survey. 
\par\medskip
Your answer should reflect the person described in the profile above, using their preferneces, beliefs and experiences.
\par\medskip
Respond with only the final answer string, not the code or label in brackets.
\par\medskip
Do not include any reasoning, explanation, or commentary.
\par\medskip
Do not preface your answer with phrases like 'I would choose'.
\par\medskip
Just return the answer text exactly as it appears in the options.
\end{tcolorbox}

\subsection{Active Endowment Generation}\label{sec:algorithms}

\begin{algorithm}[H]
\caption{Active Endowment Generation}
\label{alg:P2P_AL}
\begin{algorithmic}[1]
\STATE Initialize attribute bank $\mathcal{B}$ with core, thematic, and theoretical templates
\STATE Derive \texttt{survey}-specific attributes from $\mathcal{Q}_\mathrm{train}$ via \texttt{AttributeLearner}; append to $\mathcal{B}$

\vspace{0.5em}
\noindent \textbf{Initial Sampling Stage}
\STATE Sample initial endowments $\mathcal{E}_0$ via equal-mode sampling from $\mathcal{B}$
\STATE Instantiate initial agents $\mathcal{A}_0$ from $\mathcal{E}_0$ and elicit responses $\mathcal{R}_0$ on $\mathcal{Q}_\mathrm{train}$
\STATE Initialize \texttt{ThemeVariabilityTracker} (henceforce, \texttt{Tracker}) with $\mathcal{A}_0$, $\mathcal{R}_0$, and $\mathcal{Q}_\mathrm{train}$.

\vspace{0.5em}
\noindent \textbf{Expansion Stage}
\WHILE{agent budget $N_A$ not yet reached}
    \STATE Compute variability scores and mode sampling probabilities via \texttt{Tracker}
    \STATE Allocate endowment budget across modes according to probabilities

    \IF{question patching is enabled}
        \STATE Identify lowest-entropy questions $\mathcal{Q}_\mathrm{low}$ via \texttt{Tracker}
        \FOR{each $q \in \mathcal{Q}_\mathrm{low}$}
            \IF{$q$ is queried for the first time}
                \STATE Extract attributes for $q$ via \texttt{AttributeLearner} and append to $\mathcal{B}$
            \ELSIF{$q$ has appeared more than $n$ times}
                \STATE Enable mixed-mode strategy using $q$ and top-performing mode
            \ENDIF
        \ENDFOR
    \ENDIF

    \STATE Generate new endowments $\mathcal{E}_\mathrm{new}$ using allocated budget over sampled modes, patched question modes (if any) and mixed modes (if any).
    \STATE Instantiate agents $\mathcal{A}_\mathrm{new}$ and elicit responses $\mathcal{R}_\mathrm{new}$ on $\mathcal{Q}_\mathrm{train}$
    \STATE Update: $\mathcal{A} \gets \mathcal{A} \cup \mathcal{A}_\mathrm{new}$,\quad $\mathcal{R}_{\mathcal{A}} \gets \mathcal{R}_{\mathcal{A}} \cup \mathcal{R}_\mathrm{new}$
    \STATE Update \texttt{Tracker} with $\mathcal{A}$, $\mathcal{R}_{\mathcal{A}}$, and $\mathcal{Q}_\mathrm{train}$
\ENDWHILE
\STATE \textbf{Return:} Final agent pool $\mathcal{A}$ and full response set $\mathcal{R}_{\mathcal{A}}$
\end{algorithmic}
\end{algorithm}

\begin{algorithm}[H]
\caption{Tracker Update Procedure}\label{alg:tracker}
\begin{algorithmic}[1]
\STATE Input: Current agent pool $\mathcal{A}$, current agent responses $\mathcal{R}_{\mathcal{A}}$, and training questions $\mathcal{Q}_\mathrm{train}$
\FOR{each question $i \in \mathcal{Q}_\mathrm{train}$}
    \STATE Compute question entropy $H_i(\mathcal{A})$
\ENDFOR
\FOR{each mode}
    \STATE Compute variability score $V(\mathcal{A}_\mathrm{mode})$
\ENDFOR
\STATE Compute softmax probabilities from variability scores
\STATE Output: Variability scores, question entropies, and sampling probabilities.
\end{algorithmic}
\end{algorithm}

\newpage
\subsection{Demonstration of \algname: An Example Run on Wave 42}\label{sec:w42_extra_results}

To illustrate how \algname operates in practice, we present a detailed walkthrough of a single run on ATP Wave 42 (Trust in Science, 129 questions), tracing the system from the active endowment generation through final ensemble selection.

\subsubsection{Experimental Setup}
Questions are split according to a $7:1.5:1.5$ ratio into the training, validation, and test sets. We set the endowment budget to be $300$, with $10$ endowments generated for each mode during initial sampling. Active endowment generation is run in $10$ update steps with the preset attribute bank, question patching (lowest 3; 75\% endowment budget), and 10 attributes drawn for each endowment generation. Gemini-2.0-flash is used as the model backend.

\subsubsection{Active Endowment Generation} 
\begin{figure*}[!htb]
    \centering
    \includegraphics[width=\linewidth]{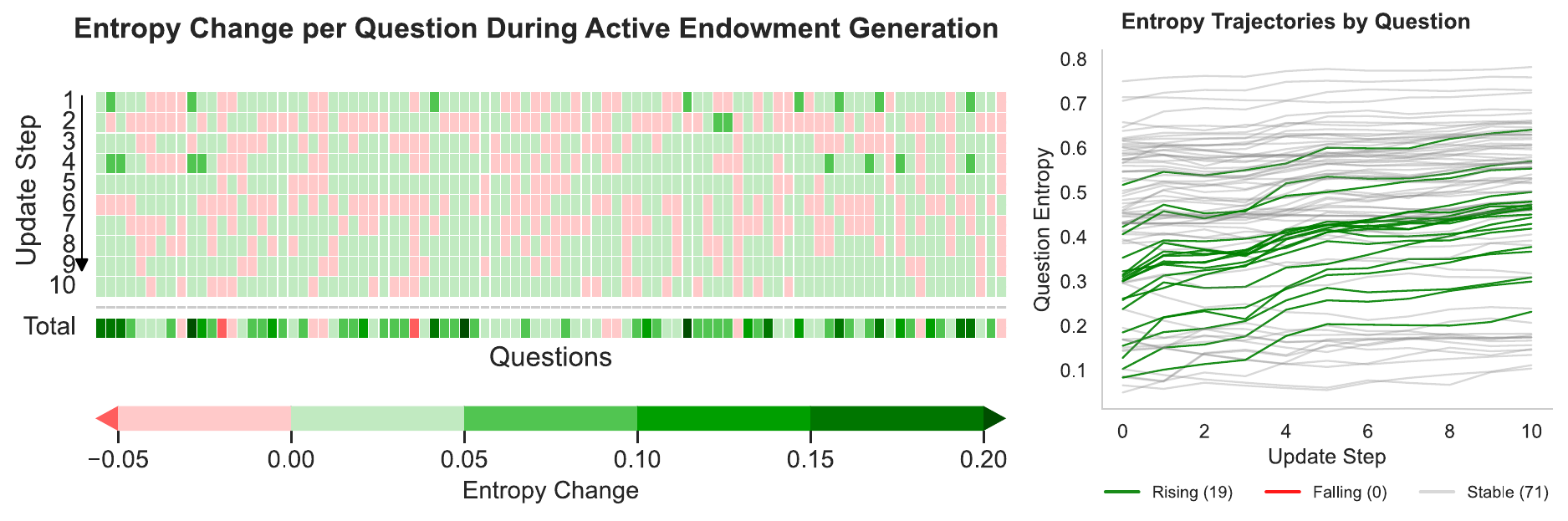}
    \caption{\textbf{Left:} Entropy change across update steps for each question during active endowment generation for ATP W42. Each column represents a question. Green indicates an increase in entropy from the previous step, while red indicates a decline. \textbf{Right:} Entropy trajectories by question (line chart). Active endowment generation progressively increases response diversity. Trends are classified based on slope and volatility of the entropy trajectory: “rising” if the slope exceeds 0.1 and standard deviation is above 0.02; “falling” if the slope is below –0.1 with sufficient volatility; otherwise labeled “stable”.}
    \label{fig:question_entropy_trajectories}
\end{figure*}

\begin{figure}[!htp]
    \centering
    \includegraphics[width=\linewidth]{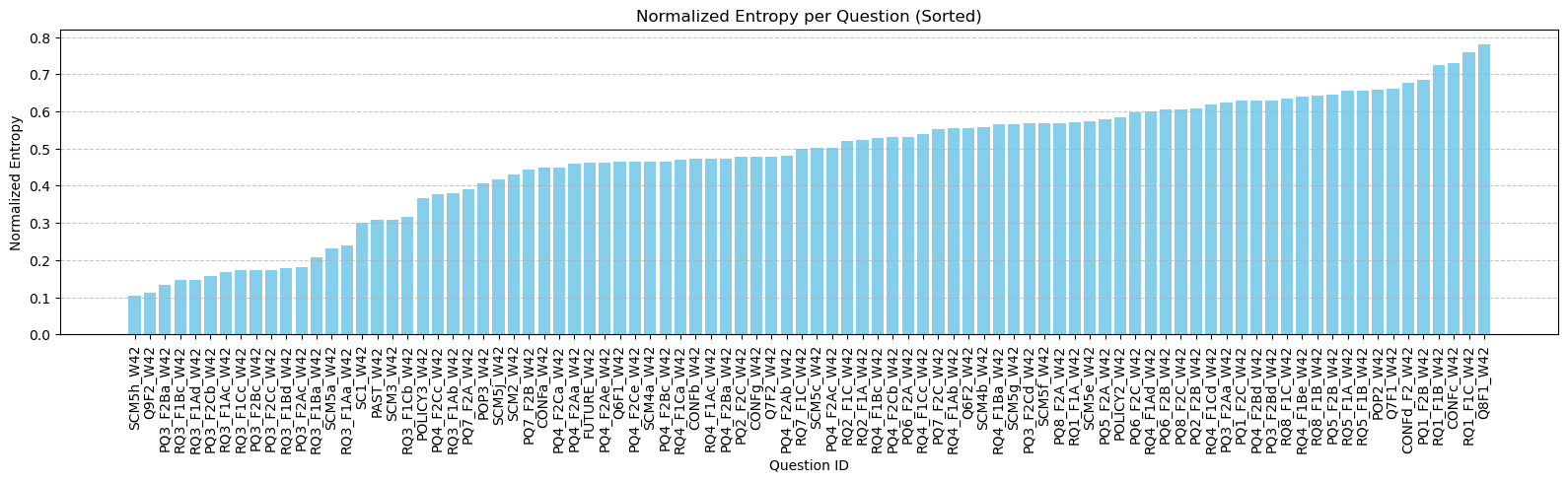}
    \caption{Question entropies at the end of active endowment generation.}
    \label{fig:entropy_bar_chart}
\end{figure}

Figure \ref{fig:question_entropy_trajectories} displays the entropy change for each question across the update steps. Evidently, a significant number of questions experience a noticeable rise in entropy at the end of the first update step. From the second step onward, mixed mode enters the generation loop. Marginal gains in entropy are observed for the majority of questions, whereas some experience mild entropy drops. Figure \ref{fig:entropy_bar_chart} displays the question entropies at the end of generation.

\newpage
\subsubsection{Regression-based Aggregation}
\begin{figure}[!htp]
    \centering
    \includegraphics[width=0.5\linewidth]{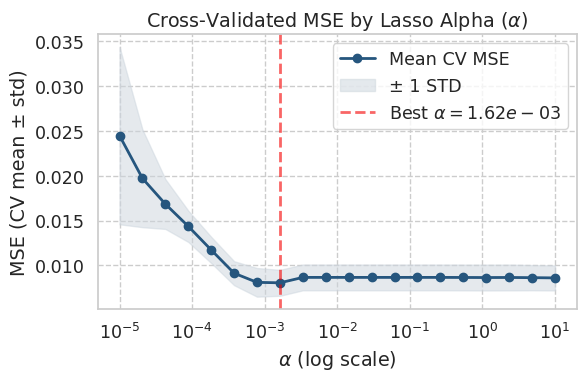}
    \caption{Cross-validated MSE by alpha ($\alpha$), the penalization parameter, for constrained lasso. The best alpha selected by CV is $1.62\times 10^{-3}$.}
    \label{fig:diagnostics_lasso}
\end{figure}

 For regression analysis, categorical questions are binarized into question-option pairs. For the selection of the agent ensemble, we fit both a constrained lasso and a constrained elastic net, using CV to select the penalization parameters. Consequently, lasso selects 58 out of the 300 synthetic agents to form the agent ensemble, while elastic net selects 107. Figure~\ref{fig:diagnostics_lasso} shows the hyperparameter selected for lasso using cross-validation.

\begin{figure}[!htb]
    \centering
    \includegraphics[width=0.9\linewidth]{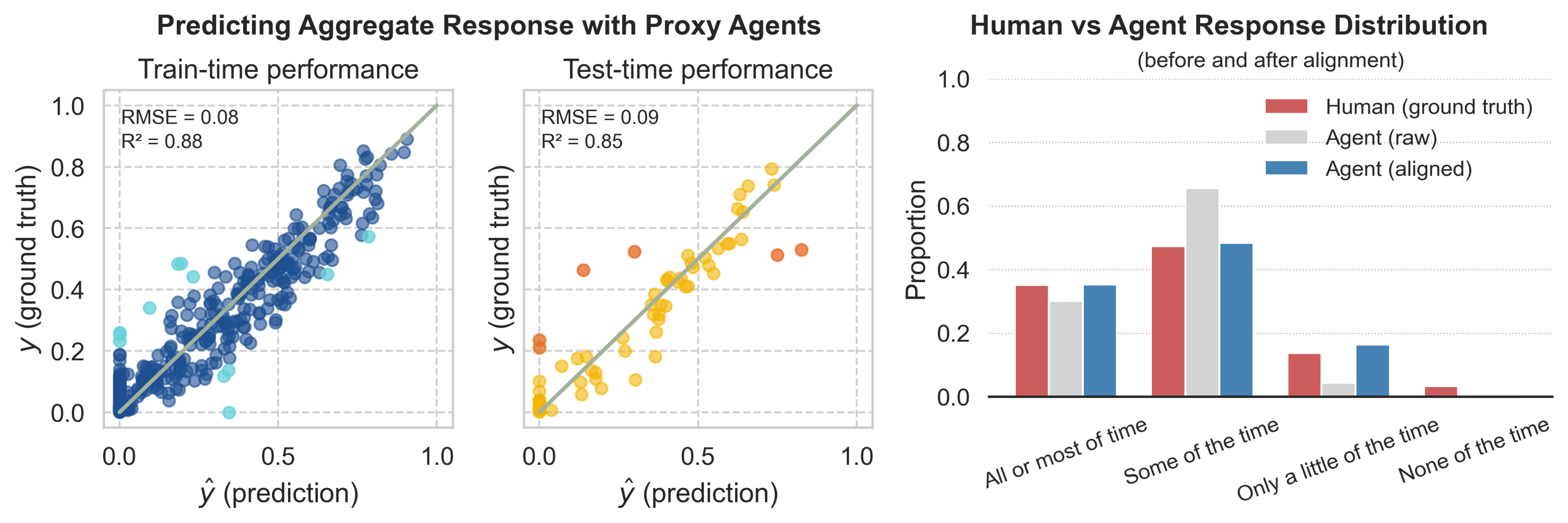}
    \caption{Prompts to Proxies: Emulation Results for ATP W42. \textbf{Left}: Train (left) and test (right) prediction performances for the binarized responses using constrained lasso. \textbf{Right}: Human and agent (pre- and post- alignment) aggregated responses for the sample test question \texttt{RQ4\_F1Ae\_W42}.}
    \label{fig:prediction_results}
\end{figure}

\begin{figure}[!htb]
    \centering
    \includegraphics[width=0.6\linewidth]{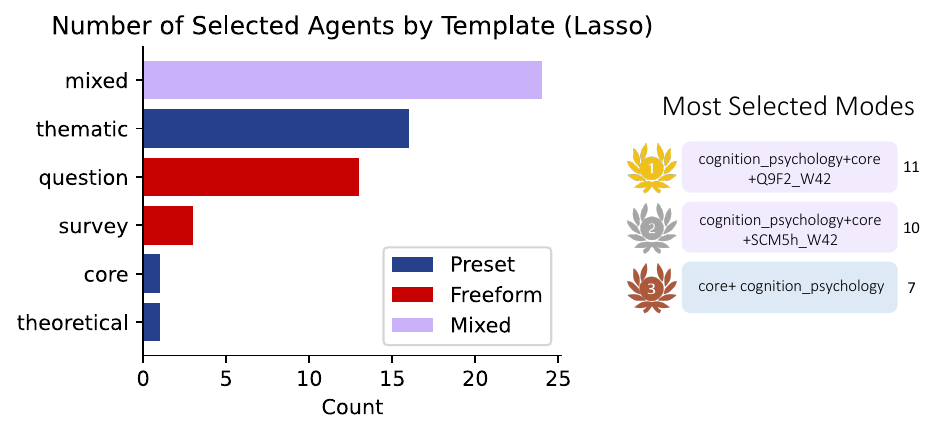}
    \caption{Breakdown of Selected Agents (58 in total) for W42 by Template. Preset, freeform, and mixed templates all contribute non-trivially to the final ensemble.}
    \label{fig:agent_breakdown}
\end{figure}

\begin{figure}[htb]
    \centering
\includegraphics[width=\linewidth]{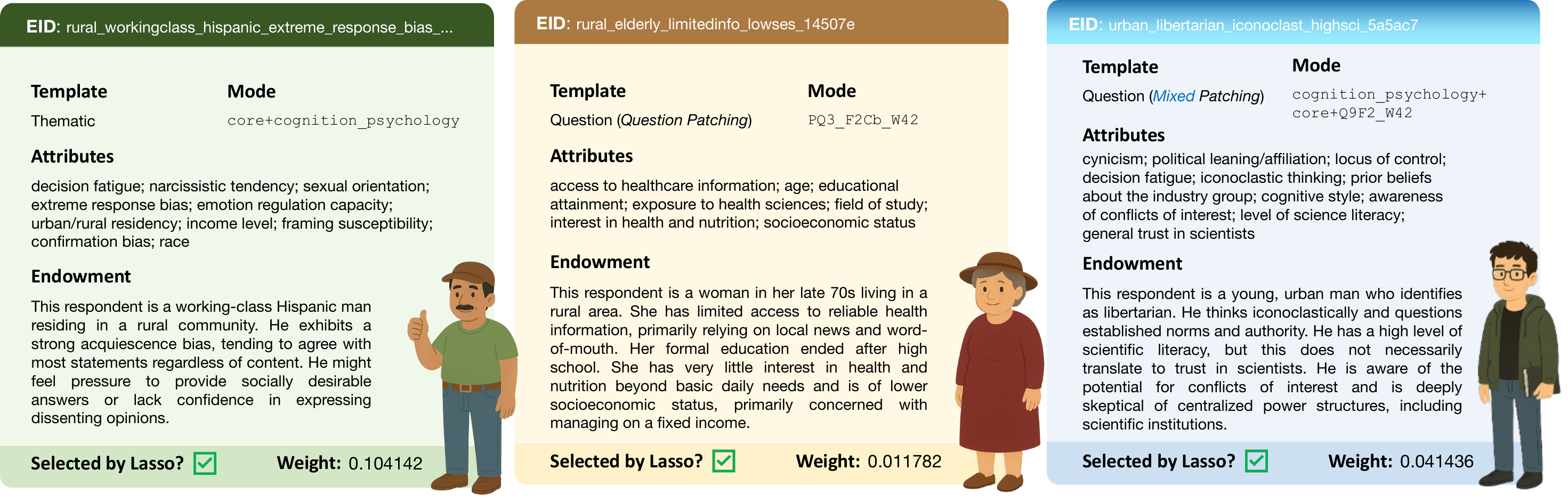}
    \caption{Examples of endowments generated by \algname for a run on ATP W42.}
    \label{fig:endowments}
\end{figure}

\begin{figure}[!htb]
    \centering
    \includegraphics[width=\linewidth]{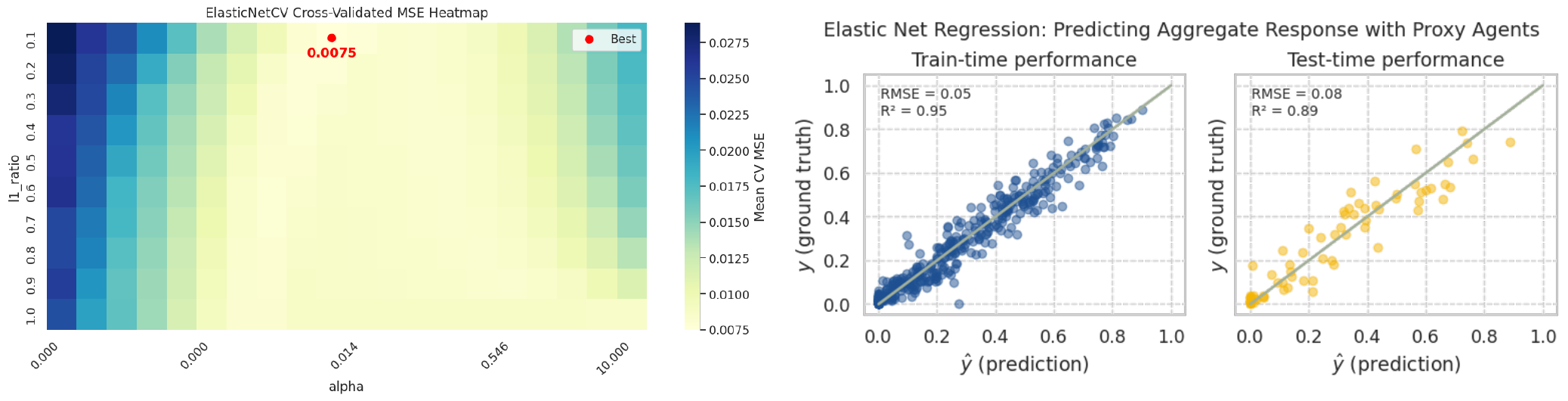}
    \caption{\textbf{Left:} Cross-validated MSE by alpha, the overall penalization parameter, and L1 ratio, the penalization weight for the L1 term, for constrained elastic net. The lowest cross-validated MSE is achieved at $\alpha = 6.95 \times 10^{-3}$, with l1 ratio set at $0.10$. \textbf{Right:} Prediction performance of the agent ensemble (107 selected agents) using constrained elastic net. Results are similar to those using constrained lasso.}
    \label{fig:diagnostics_elasticnet}
\end{figure}

Figure \ref{fig:prediction_results} shows the prediction performances of the agent ensemble selected with constrained lasso. The train-time root mean squared error (RMSE) is 0.06, with test RMSE slightly higher at 0.08, indicating adequate generalization ability of the functional basis. The right panel of Figure \ref{fig:prediction_results} presents a snapshot for the aggregated responses to a test set question, comparing the human ground truth with agentic emulations before and after regression-based aggregation. Overall, the results are encouraging, especially considering that \algname constructs the endowments based solely on preset and learned attributes and is agnostic on the ground-truth demographic data. Figure \ref{fig:agent_breakdown} provides a breakdown of the agent ensemble selected by lasso. Notably, preset, freeform, and mixed templates all contribute non-trivially to the final ensemble. Figure \ref{fig:endowments} provides three examples of endowments generated and selected  by \algname for the run on ATP W42, tracing back to their attributes, modes, and templates. Additionally, results of constrained elastic net are shown in Figures \ref{fig:diagnostics_elasticnet}. Overall, the prediction results are similar to those using constrained lasso.

\clearpage
\section{Design Validation: Simulation Studies} \label{app:simulation_studies}
In order to gauge the validity of the proposed preference model and the factors affecting alignment performance, we conduct three simulation studies where we have the privilege of ground-truth knowledge.
\paragraph{Simulation Design}
For the simulation studies, we use the American Trends Panel (ATP) Wave 42 (W42) as the survey source, covering the topic of public trust in science and scientists. We run the active endowment generation on the survey to get 300 persona endowments, and partition the survey questions into training, validation and test sets by the 0.7:0.15:0.15 ratio. We subsequently simulate the responses to the survey questions using endowed GPT-4o agents as the synthetic respondents. The ATP W42 dataset has in total 129 questions, which after binarization leads to 517 observations—85\% of these observations form the train-and-validation (trainval) set and the rest 15\% is used to test the predictive accuracy of the selected agent ensemble. 

\paragraph{Construction of the Ground Truth Data}
To construct the ground truth data, we sample a fraction of the endowments as the ground truth and randomly assign weights to these endowments, which represent the proportions of the associated personas in the ground truth population. The ground truth data—percentages of individuals choosing difference options for a given question—are computed using weighted averages of responses provided by the ground-truth agents. The unselected endowments are used to construct the pool of proxy agents.

\paragraph{Regression Method}
For the simulation studies, we choose constrained lasso as our main regression alignment method for investigation. As a secondary choice, constrained elastic net is an extension of constrained lasso that trades off between the L1 and L2 penalty terms. It is expected to have more stable performance when the features (agent responses) are highly correlated. Our initial simulations indicate that constrained elastic net demonstrates near identical performance to constrained Lasso, partly because the latter is a special case of the former and we are using cross-validation to select the best penalization parameters. Therefore, for the sake of building intuitions, we focus on constrained lasso in the simulations and expect the results to be largely indicative of constrained elastic net's performance.

\subsection{Recovering Ground Truth Agents with Constrained Lasso}
\label{app:simulation_1}

As a first test of constrained lasso as a variable selection algorithm, we aim to evaluate its ability to recover ground-truth agents from a broader candidate pool. To render the selection task nontrivial, we set the prevalence of the ground-truth agents to 0.3—out of 10 selectable agents, only 3 contribute to the construction of the ground-truth signal. As the proxy agents dominate the agent pool, an adequate number of observations becomes essential for Lasso to reliably identify the true contributors. As a key metric, we define \textbf{observation-to-agent ratio (OAR)} to be the ratio of the number of training and validation observations to the number of selectable agents. By default, the OAR for our simulation baseline is 1.203.

To develop a panoramic view of constrained lasso's ability in ground truth recovery, we conduct multiple simulation rounds under varying OARs. We generate different OAR variations using two complementary subsampling strategies:
\begin{itemize}
    \item \textbf{Subsampling endowments}. To increase the OAR, we adopt a subsampling strategy on endowments, where we subsample from the pool of selectable endowments, varying the total number of candidate agents while keeping the proportion of ground-truth agents fixed at 30\%. The ground-truth weights are renormalized to sum to 1. We vary the subsample fraction from 0.5 to 1 in 20 equally spaced steps.
    \item \textbf{Subsampling questions}. To decrease the OAR, we subsample survey questions while keeping the original train-validation-test split and ratio unchanged. This mimics a setting with reduced behavioral signal for inference. As in the endowment-based approach, we use 20 subsample fractions ranging from 0.5 to 1.
\end{itemize}

\begin{figure*}[!htb]
    \centering
    \includegraphics[width=0.8\textwidth]{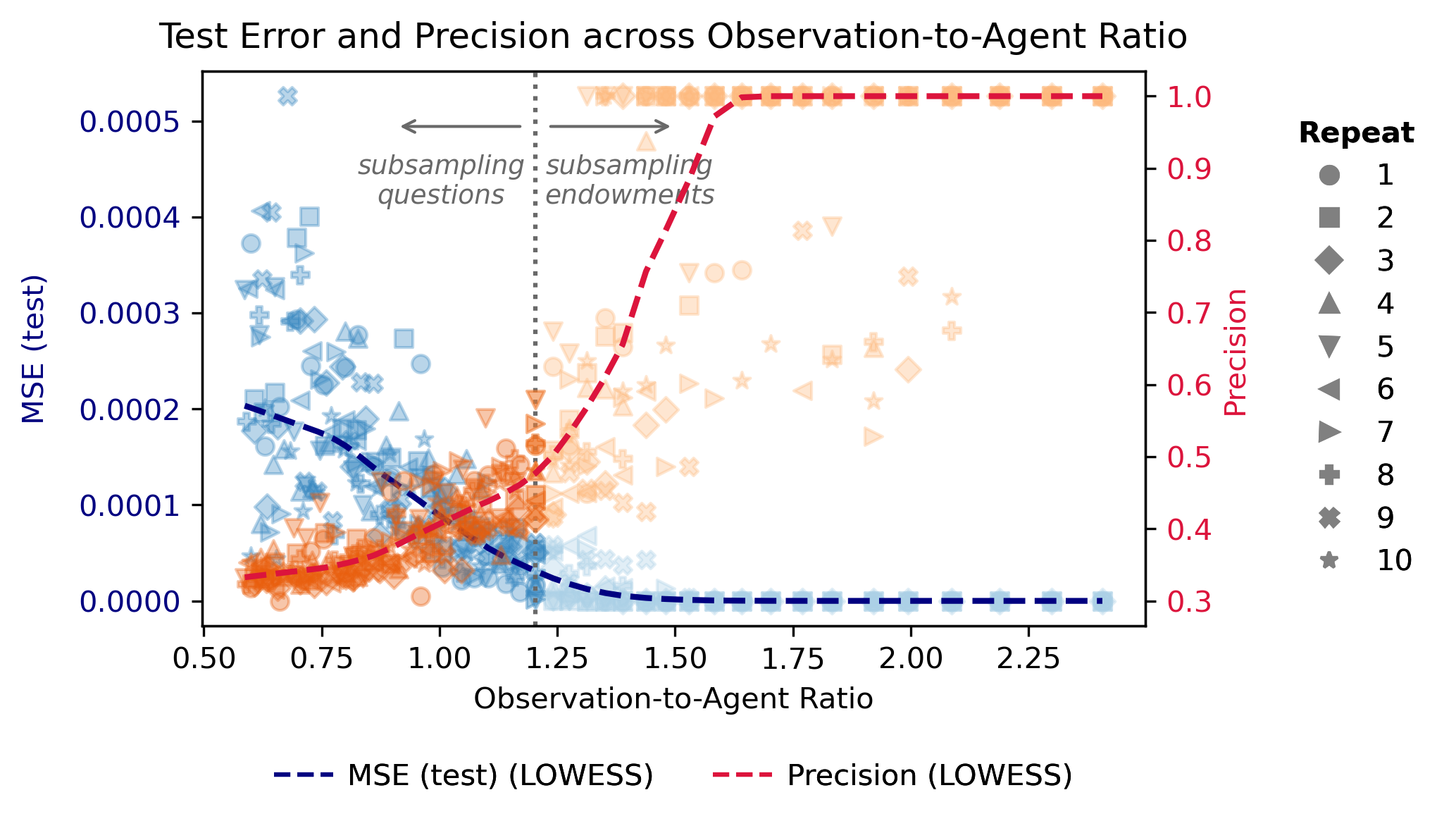}
    \caption{Mean Squared Error (Test Set) and Precision by Different Observation-to-Agent Ratio (OAR) for Constrained Lasso. Each simulation is repeated 10 times with different random seeds. Subsampling of endowments is done with fixed ground-truth-to-proxy ratio. Subsampling of responses is done with fixed train-valid-test split and ratio.}
    \label{fig:ground_truth_recovery}
\end{figure*}

\begin{figure*}[!htb]
    \centering
    \includegraphics[width=0.8\textwidth]{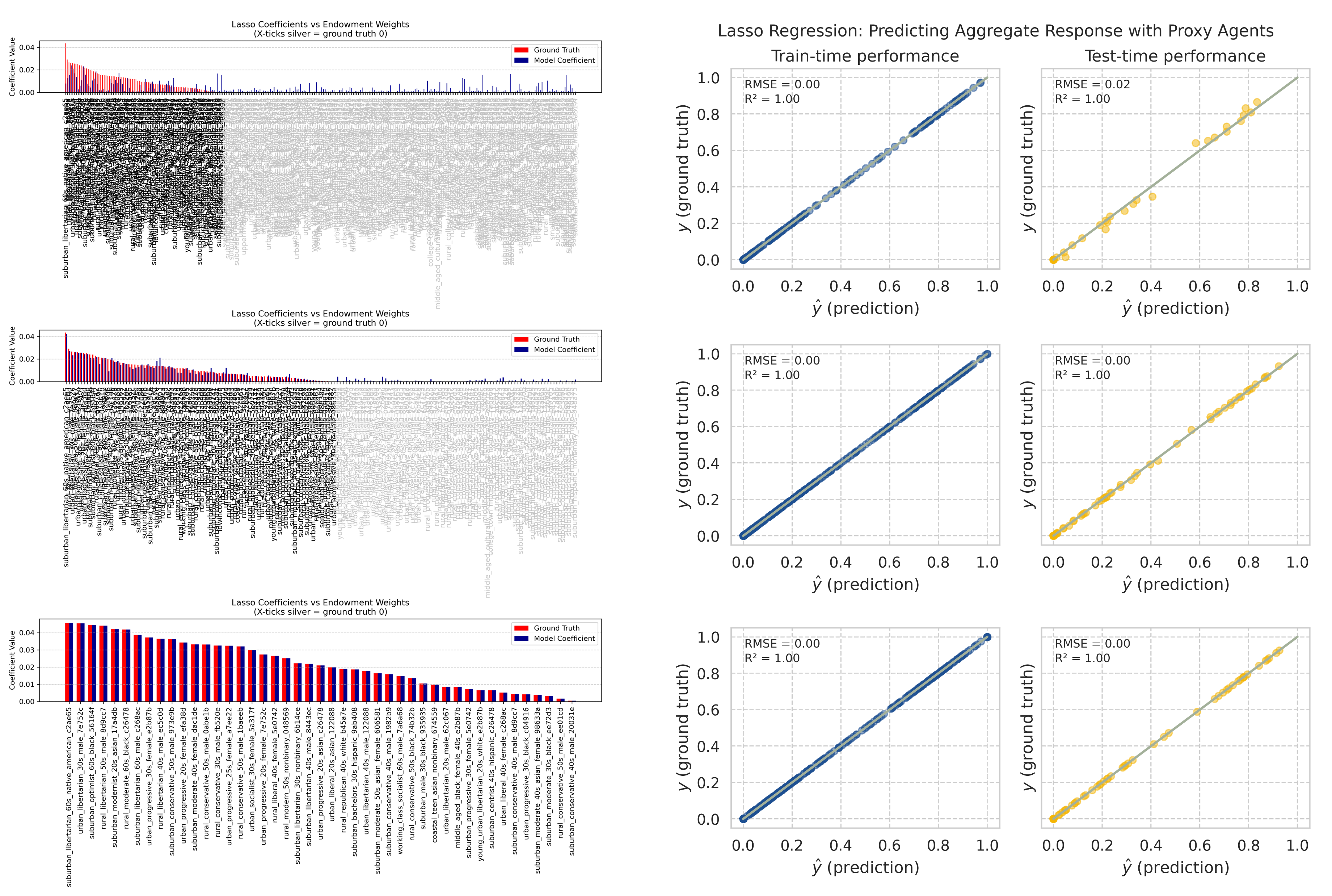}
    \caption{Snapshots of Weight-vs-Coefficient Comparison and Prediction Performance. Each row represents a simulation run. The red bars on the left panel represent ground truth agent weights used to generate the aggregate data and the blue bars Lasso coefficients. The observation-to-agent ratios for the top, middle and bottom panels are: 0.617, 1.203, 2.407. Lasso's ability to recover the ground truth agents  from the agent pool degrades with lower observation-to-agent ratio, but its predictive accuracy only drops mildly.}
    \label{fig:ss1_checkpoints}
\end{figure*}

We repeat each subsampling procedure 10 times to construct the final dataset used in our simulation study. Figure \ref{fig:ground_truth_recovery} shows the trends of the mean squared error of the test set and the precision—that is, the proportion of selected agents who possess ground-truth endowments—across different OARs. To further illustrate the dynamics of selection, Figure \ref{fig:ss1_checkpoints} presents three representative snapshots of simulation runs, corresponding to low, medium, and high OAR settings.

As the figures indicate, constrained lasso's ability to recover ground-truth agents from the candidate pool improves with a higher observation-to-agent ratio. This is intuitive as more observations offer a richer signal of both individual-level idiosyncrasies and aggregate preferences. With abundant behavioral cues, the selection algorithm is better equipped to distinguish true contributors from correlated siblings. In contrast, when observations are sparse, Lasso tends to relinquish its sharpness as a selection tool and instead defaults to identifying a functional basis that best explains the limited data. The mild decline in predictive accuracy under low OAR regimes suggests that this functional basis still approximates the aggregate preference reasonably well, albeit imperfectly.

\subsection{Emulating Ground Truth Responses with Proxy Agents}
\label{app:simulation_2}
\begin{figure*}[!htb]
    \centering
    \includegraphics[width=1\textwidth]{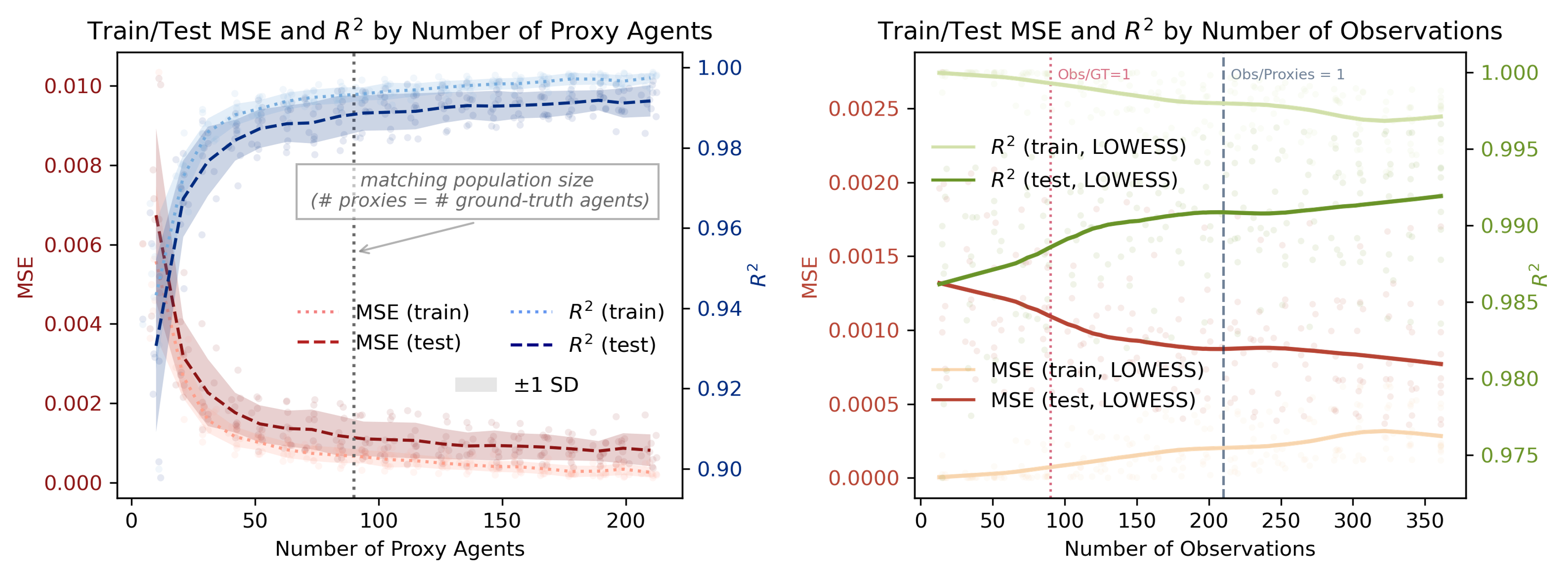}
    \caption{Left: Mean Squared Errors and Coefficients of Determination ($R^2$) by Number of Proxy Agents. Right: Mean Squared Errors and Coefficients of Determination ($R^2$) by Number of Observation. Each simulation is repeated 10 times with different random seeds.}
    \label{fig:proxy_emulation}
\end{figure*}

While the first simulation exercise has readily allured to constrained lasso's ability to find a functional basis to explain the observed data, in this second simulation study we extend the investigation by removing the ground-truth agents from the selectable agent pool. Under this setting, constrained lasso needs to fully rely on proxy agents to recover the ground-truth patterns. We analyze two distinct scenarios:
\begin{itemize}
    \item \textbf{Subsampling proxies.} We fix the total number of observations and reduce the proxy pool by retaining only a fraction as selectable agents. We sweep across the fractions from 0.05 to 1 in 20 equal spaced steps.
    \item \textbf{Subsampling training and validation questions.} We fix the proxy pool but retain a fraction of training and validation observations. Similar to subsampling proxies, we construct 20 simulation rounds gradually raising the retained fraction from 0.05 to 1.
\end{itemize}
Like in Simulation Study 1, we repeat each simulation 10 times using different random seeds to form the final results. In Figure \ref{fig:proxy_emulation}, the left panel displays the mean squared errors (MSE) and coefficients of determination ($R^2$) for the subsampling proxies scenario, while the right panel showcases the associated metrics for the subsampling training and validation questions scenario.

\begin{figure*}[!htb]
    \centering
    \includegraphics[width=1\textwidth]{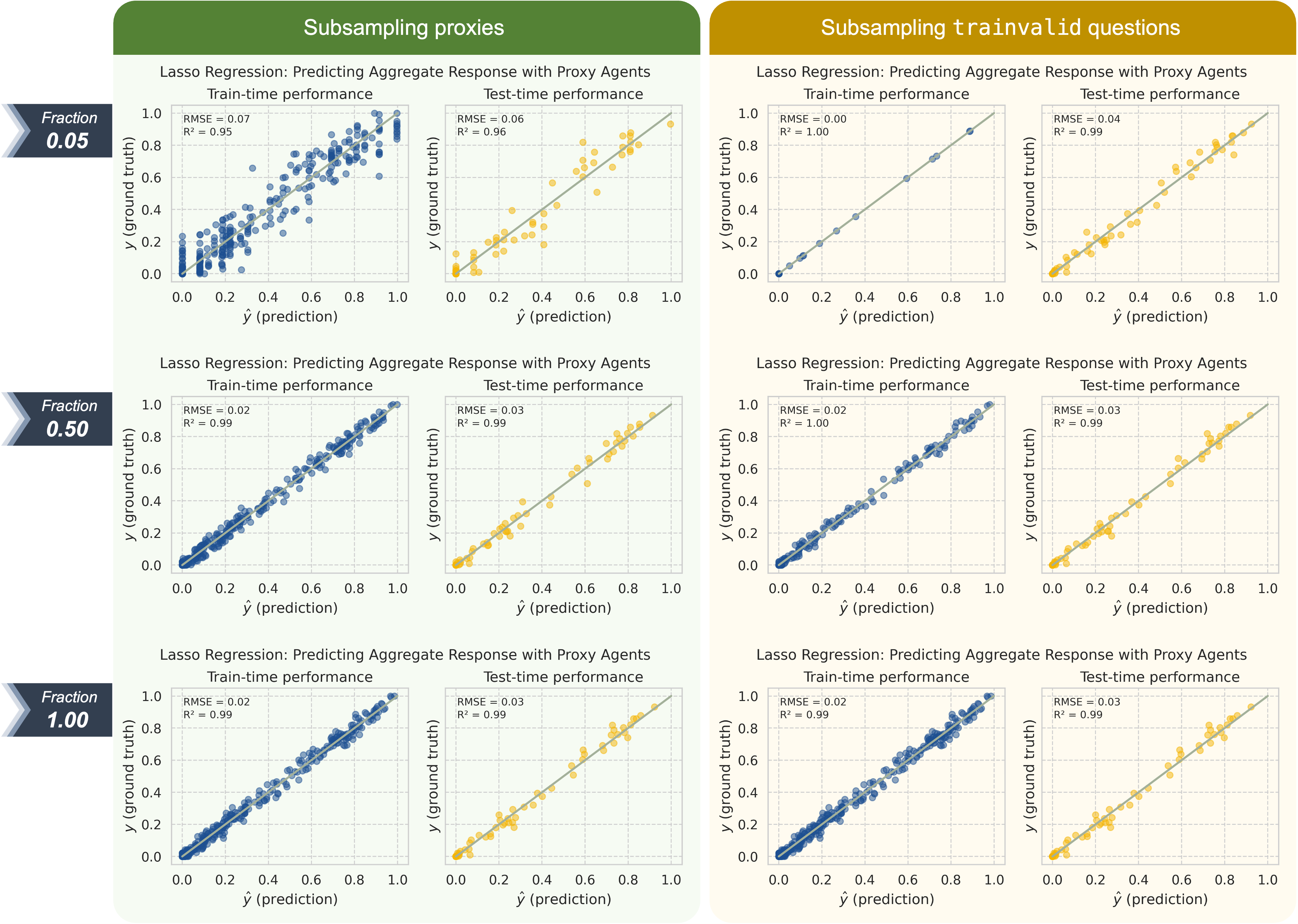}
    \caption{Snapshots of Prediction Performance for Simulation Study 2.}
    \label{fig:ss2_checkpoints}
\end{figure*}

Notably, constrained lasso exhibits robust performance across all simulated conditions. Even with only 10 selectable proxy agents, it achieves a mean test MSE of $0.0067$, yielding an $R^2$ of $0.93$. Predictive accuracy improves as we increase the number of selectable proxy agents, plateauing at $0.99$. In comparison, decreasing the number of observations has a lesser effect on predictive performance: in the recorded run with the lowest number of observations at 13, constrained lasso attains a test MSE of $0.0019$ and an $R^2$ of $0.98$. Figure \ref{fig:ss2_checkpoints} displays representative snapshots of the simulation runs.

The high predictive accuracy of the simulation runs indicates that constrained lasso can efficiently find a functional basis from a select few agents and determine the agent weights to emulate the ground-truth data patterns based on limited observations. This is in fact not surprising because in the initial assignment of the endowments into the ground truth and proxy groups we have used a random partition. As informed by our proposed behavioral preference model, this causes the ground truth vectors—used in the simulation study to form the aggregate preference—and proxy vectors to span the same preference space. As the preference space is not directly accessible, we can derive a partial gauge through the observed agent responses. If the responses of a group of agents have higher variability, it is an indication that the preference space they span is large. Conversely, if agents tend to agree on their responses to different questions, it is a signal that their associated vectors in the preference space might overlap in sub-dimensions, leading to a pool generalization ability.

To effectively measure response variability, in the paper we introduce the notion of \textbf{question entropy}:
\begin{equation}
    H_i(\mathcal{A}) = \frac{-\sum_{k=1}^{K_i} p_{ik} \log_2 p_{ik}}{\log_2 K_i}
\end{equation}
where $H_i(\mathcal{A})$ is the normalized entropy of question $i$ based on responses from a group of agents $\mathcal{A}$, $K_i$ denotes the number of unique response options $\{1, \dots, K_i\}$ for question $i$, and $p_{ik}$ represents the empirical proportion of responses selecting option $k$.
The normalization factor ensures comparison across questions. A high question entropy indicates that the responses are highly varied for this question.

For the simulation studies, we define \textbf{group entropy} as the average question entropy computed using responses from that group:
\begin{equation}
    {H}(\mathcal{A}) = \frac{1}{N}\sum_{i=1}^{N} H_i(\mathcal{A}).
\end{equation}
The group entropy is calculated using responses to all survey questions, regardless of the split. Based on it, we introduce the notion of \textbf{Entropy Coverage Ratio (ECR)} defined as the ratio between the group entropy of proxy agents and that of the ground-truth agents:
\begin{equation}
    ECR = \frac{{H}(\mathcal{A}_\text{proxy})}{{H}(\mathcal{A}_\text{gt})}.
\end{equation}
An ECR lower than 1 indicates that the proxy agents have less response variability than the ground truth, while an ECR greater than 1 indicates the opposite.

\begin{figure*}[!htb]
    \centering
    \includegraphics[width= 0.8\textwidth]{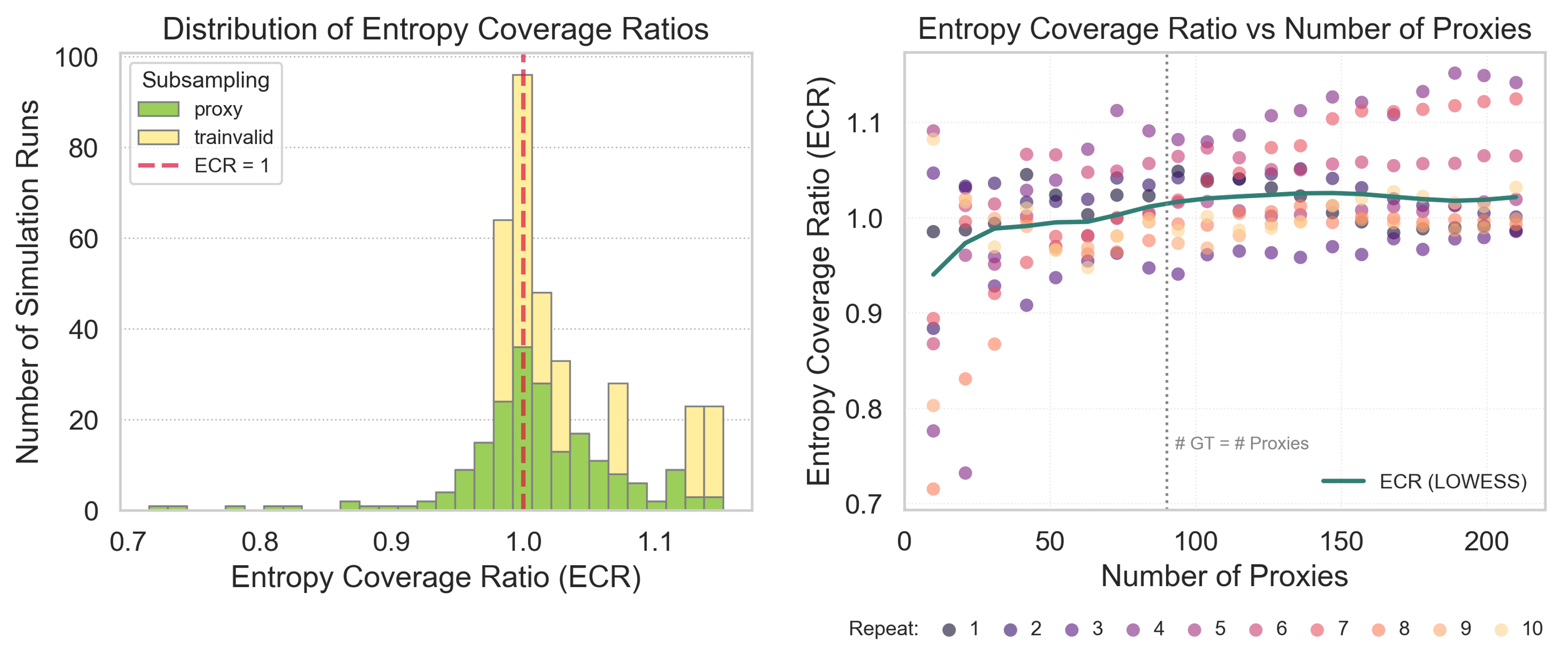}
    \caption{Entropy Coverage Ratio Diagnostics for Simulation Study 2.}
    \label{fig:SS2_ECR}
\end{figure*}

The left panel of Figure \ref{fig:SS2_ECR} shows the distribution of entropy coverage ratios (ECRs) for the two analyzed scenarios. The mass is concentrated around $1.0$, corroborating the earlier observation that random partitioning tends to preserve entropy structure. The right panel plots ECR against the number of proxies for the proxy-subsampling scenario. The relationship between ECR and the number of proxies is subtly positive: when the number of proxies is small, the likelihood of drawing an unrepresentative sample increases. As the sample size grows, group-level entropy converges toward the population-level entropy—an empirical manifestation of the law of large numbers.

\subsection{Response Variability Matters}
\label{app:simulation_3}
In this final simulation study, we analyze the effect of the entropy coverage ratio on predictive accuracy for constrained lasso.

\begin{figure*}[!htb]
    \centering
    \includegraphics[width=0.8\textwidth]{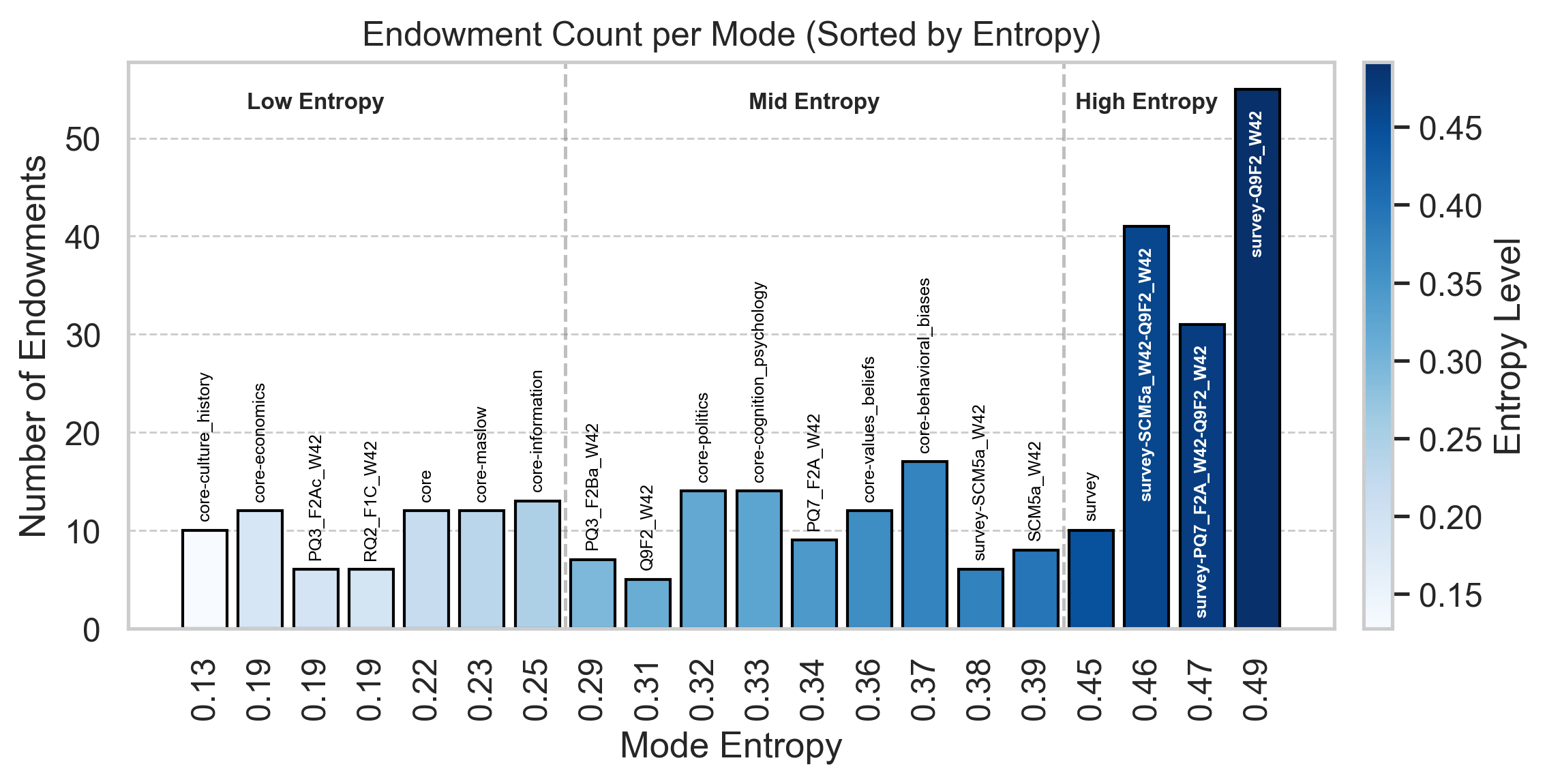}
    \caption{Endowment Count per Mode (Sorted by Entropy).}
    \label{fig:mode_entropy_distribution}
\end{figure*}

Figure \ref{fig:mode_entropy_distribution} displays the group entropies for THE W42 endowments organized by modes—we call them \textbf{mode entropies} in the endowment generation logic. We subsume them further into three different tiers according to the group entropy value: in total, we have 71 endowments in the low entropy tier, 92 in the mid entropy tier, and 137 in the high entropy tier.

To cover a wide range of ECR ratios, we design the following simulation strategy:
\begin{enumerate}
    \item \textbf{Ground Truth Construction.} For each simulation round, pick one of the three tiers as the ground-truth tier and sample 30 endowments from the tier to construct the ground-truth agents and assign ground-truth weights. The unselected endowments from the tier are joined with the endowments from the other two tiers to form the pool of proxy candidates.
    \item \textbf{Proxy Agents Construction.} For each simulation run, we select 30 proxy agents to form the proxy pool. We begin by selecting agents from the lowest entropy modes (ensuring that the cumulative candidate endowments surpass 30), and in each consecutive run, include the next mode with higher entropy into the selectable modes. We sweep across the modes by increasing mode entropy.
    \item \textbf{Proxy Agents Selection and Response Emulation.} In each simulation run, after the proxy pool has been constructed, we use constrained lasso to select proxy agents to emulate the  observed aggregate response data.
\end{enumerate}

\begin{figure*}[!htb]
    \centering
    \includegraphics[width= 0.8\textwidth]{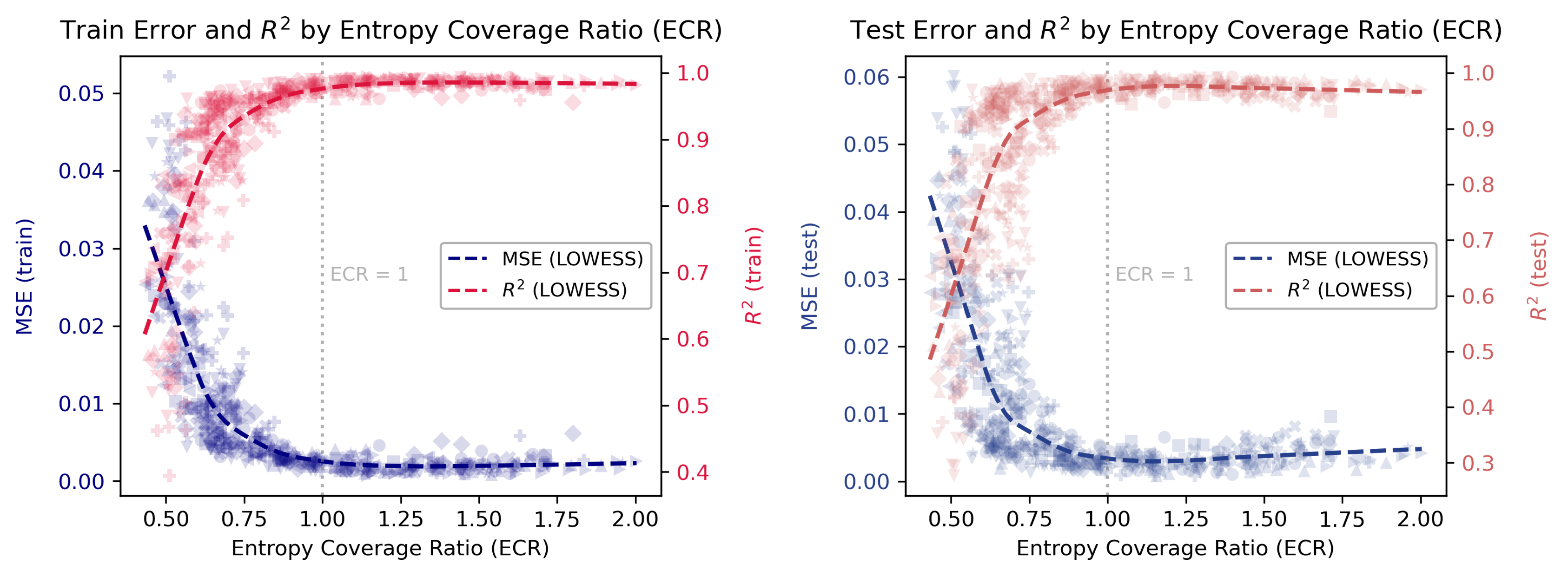}
    \caption{Mean Squared Error and Coefficient of Determination ($R^2$) by Entropy Coverage Ratio (ECR) for the trainval set (left) and test test (right). Entropy Coverage Ratio is defined as the ratio of the average response entropies of the proxies and ground-truth agents. LOWESS curves are fit using 10 repetitions.}
    \label{fig:error_r2_by_ecr}
\end{figure*}

We repeat each simulation 10 times to form the final results. Figure \ref{fig:error_r2_by_ecr} visualizes the train and test performance of the constrained lasso by the entropy coverage ratio. As the ECR increases, the predictive accuracy improves at both the training and the test times, lending further credence to the analysis based on the preference model in the second simulation study.

\begin{figure*}[!htb]
    \centering
    \includegraphics[width=1\textwidth]{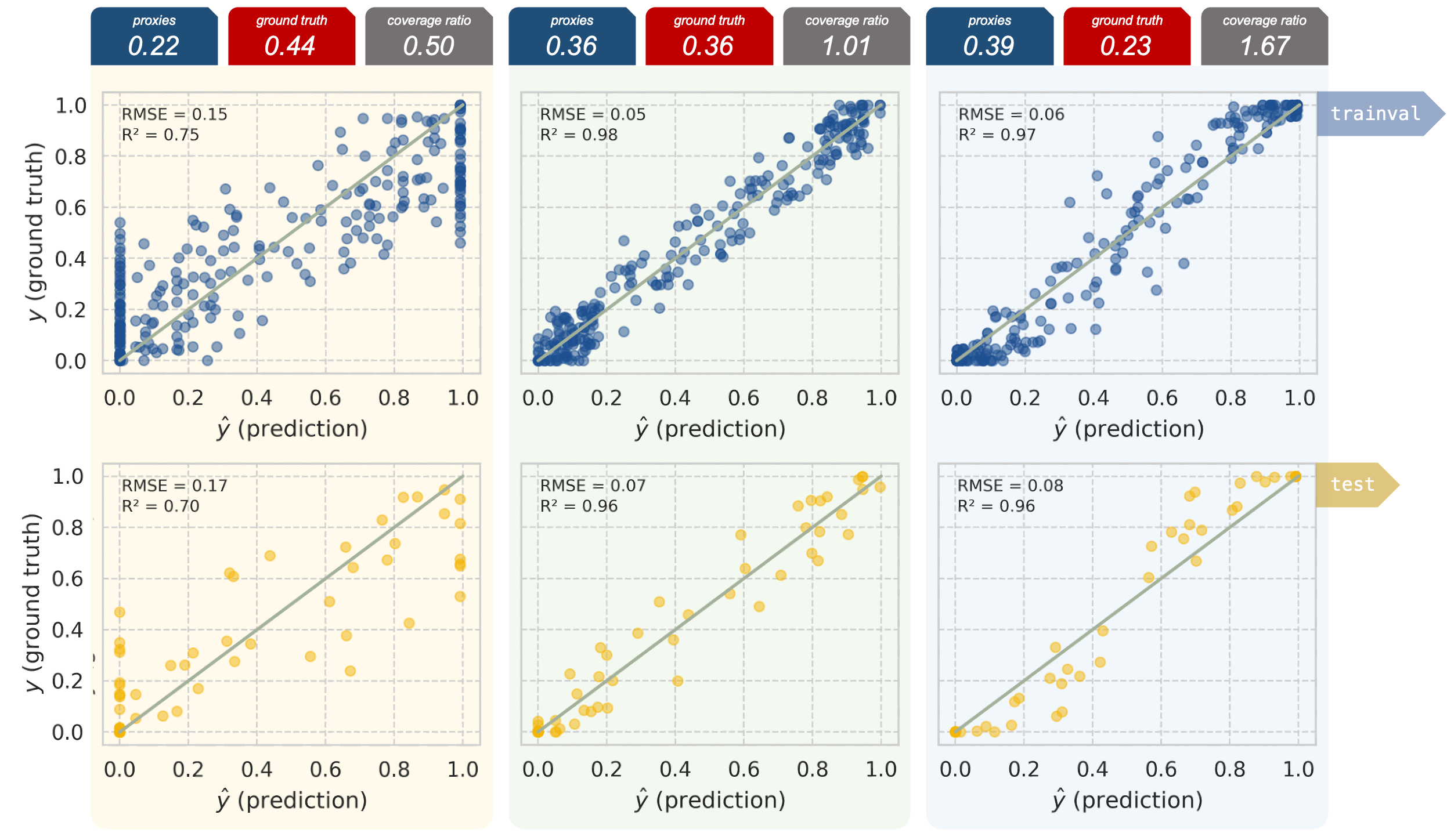}
    \caption{Snapshots of Prediction Performance for Simulation Study 3.}
    \label{fig:ss3_checkpoints}
\end{figure*}

Figure \ref{fig:ss3_checkpoints} offers three snapshots of the simulation runs corresponding to ECRs equaling $0.50$, $1.01$ and $1.67$. When proxy entropy is low compared with ground truth, the proxy agents fail to provide a sufficiently expressive basis to span the aggregate ground-truth preference. As a result, constrained Lasso can only approximate a lower-dimensional projection of the true preference signal. This results in poor prediction performance, as illustrated in the leftmost snapshot. Conversely, when proxy entropy exceeds ground-truth entropy, prediction error begins to increase—albeit at an infinitesimal scale. The subtle deterioration of performance when ECR surpasses 1 could be due to various reasons. A plausible cause is from the data perspective: when the proxy pool becomes too diverse, it may introduce spurious variability that does not align well with the true underlying structure. Put differently, an observed response might amalgamate signals from different latent preference factors, some of which are irrelevant to the aggregate preference in focus. While greater diversity expands the representational capacity of the proxy pool, it may also dilute the signal and induce overfitting, thereby reducing generalization accuracy. In practice, we posit that over-diversity is less likely than under-diversity, as existing studies using demographic conditioning \cite{Argyle_Busby_Fulda_Gubler_Rytting_Wingate_2023, park2024generativeagentsimulations1000} indicate LLM agents do not match closely with human ground truth. This justifies optimizing response variability in Stage 1 of \algname. 

The key takeaway from the simulation study is that absent the knowledge of ground truth, it is crucial to ensure an adequate group entropy among the proxy agents to form a functional basis for ground-truth emulation. However, one should also refrain from solely relying on entropy as a measure of preference diversity lest spurious correlations beguile us under limited data regimes. While our alignment method is intentionally designed to function with minimal data—requiring only aggregate response data—in practice, if individual human responses are available, a practitioner may use the group entropy computed from these responses as an anchor for endowment generation.

\clearpage
\section{Empirical Experiments: Supplementary Results}\label{sec:addmainresults}

\subsection{Main Results: Baseline Comparison on ATP Waves}
In our setup, MSE is computed on the option-level probabilities after binarizing each multiple-choice item, so MSE penalizes misallocation of probability mass across the full categorical distribution. To facilitate comparisons across works and corroborate the use of MSE as a valid distribution metric, we follow previous studies \citep{cao-etal-2025-specializing,feng-etal-2024-modular,durmus2024measuringrepresentationsubjectiveglobal} and additionally report Jensen-Shannon Divergence (1-JSD) and Earth Mover Distance (EMD) for the key results in this paper.

Table~\ref{tab:full_metrics_14waves} presents the full distribution metrics comparison of baselines and \algname across 14 ATP waves. \algname consistently outperforms Vanilla and PERSONA across all three metrics, registering lower MSE (0.014 vs. 0.025--0.029), higher 1-JSD (0.830 vs. 0.759--0.771), and lower EMD (0.066 vs. 0.085--0.091). The 23--27\% reduction in EMD indicates that \algname predictions are structurally closer to ground-truth distributions.

\begin{table*}[htp]
\centering
\newcommand{\sd}[1]{{\footnotesize(#1)}}
\caption{Full distribution metrics comparison across 14 ATP waves. MSE measures mean squared error of predicted response probabilities. 1-JSD (1 minus Jensen-Shannon Divergence) measures distributional similarity (higher is better). EMD (Earth Mover's Distance) measures distributional distance (lower is better). Each cell reports the mean over three runs, with the sample SD in parentheses below.}
\label{tab:full_metrics_14waves}
\small
\begin{tabular}{l ccc ccc ccc}
\toprule
& \multicolumn{3}{c}{Vanilla} & \multicolumn{3}{c}{PERSONA} & \multicolumn{3}{c}{\algname} \\
\cmidrule(lr){2-4} \cmidrule(lr){5-7} \cmidrule(lr){8-10}
Wave & MSE & 1-JSD & EMD & MSE & 1-JSD & EMD & MSE & 1-JSD & EMD \\
\midrule
W26 & .015 & .820 & .072 & .024 & .772 & .082 & .010 & .868 & .056 \\
    & \sd{.002} & \sd{.011} & \sd{.004} & \sd{.003} & \sd{.009} & \sd{.004} & \sd{.000} & \sd{.001} & \sd{.001} \\ \addlinespace[2pt]
W27 & .032 & .795 & .086 & .041 & .728 & .104 & .022 & .823 & .071 \\
    & \sd{.004} & \sd{.008} & \sd{.007} & \sd{.001} & \sd{.004} & \sd{.001} & \sd{.001} & \sd{.006} & \sd{.003} \\ \addlinespace[2pt]
W29 & .026 & .767 & .095 & .030 & .748 & .106 & .016 & .815 & .083 \\
    & \sd{.003} & \sd{.009} & \sd{.002} & \sd{.002} & \sd{.008} & \sd{.005} & \sd{.005} & \sd{.026} & \sd{.013} \\ \addlinespace[2pt]
W32 & .019 & .804 & .067 & .025 & .772 & .082 & .016 & .832 & .065 \\
    & \sd{.000} & \sd{.006} & \sd{.001} & \sd{.000} & \sd{.004} & \sd{.001} & \sd{.002} & \sd{.007} & \sd{.003} \\ \addlinespace[2pt]
W34 & .021 & .777 & .084 & .018 & .790 & .080 & .011 & .822 & .070 \\
    & \sd{.004} & \sd{.015} & \sd{.009} & \sd{.001} & \sd{.010} & \sd{.006} & \sd{.001} & \sd{.010} & \sd{.007} \\ \addlinespace[2pt]
W36 & .033 & .747 & .085 & .028 & .787 & .074 & .019 & .821 & .061 \\
    & \sd{.001} & \sd{.011} & \sd{.005} & \sd{.007} & \sd{.011} & \sd{.002} & \sd{.004} & \sd{.016} & \sd{.008} \\ \addlinespace[2pt]
W41 & .022 & .761 & .101 & .027 & .756 & .102 & .013 & .828 & .069 \\
    & \sd{.001} & \sd{.003} & \sd{.003} & \sd{.003} & \sd{.018} & \sd{.009} & \sd{.001} & \sd{.002} & \sd{.006} \\ \addlinespace[2pt]
W42 & .029 & .767 & .081 & .036 & .753 & .090 & .011 & .853 & .053 \\
    & \sd{.005} & \sd{.010} & \sd{.004} & \sd{.001} & \sd{.003} & \sd{.001} & \sd{.000} & \sd{.003} & \sd{.001} \\ \addlinespace[2pt]
W45 & .015 & .805 & .072 & .029 & .773 & .085 & .008 & .848 & .054 \\
    & \sd{.003} & \sd{.011} & \sd{.007} & \sd{.001} & \sd{.006} & \sd{.001} & \sd{.002} & \sd{.013} & \sd{.010} \\ \addlinespace[2pt]
W49 & .046 & .734 & .093 & .041 & .746 & .089 & .021 & .814 & .065 \\
    & \sd{.001} & \sd{.013} & \sd{.004} & \sd{.006} & \sd{.013} & \sd{.001} & \sd{.004} & \sd{.017} & \sd{.004} \\ \addlinespace[2pt]
W50 & .027 & .710 & .100 & .026 & .699 & .112 & .013 & .805 & .075 \\
    & \sd{.003} & \sd{.019} & \sd{.006} & \sd{.002} & \sd{.011} & \sd{.001} & \sd{.004} & \sd{.015} & \sd{.008} \\ \addlinespace[2pt]
W54 & .020 & .753 & .085 & .024 & .748 & .080 & .014 & .801 & .068 \\
    & \sd{.003} & \sd{.021} & \sd{.009} & \sd{.003} & \sd{.009} & \sd{.002} & \sd{.002} & \sd{.019} & \sd{.007} \\ \addlinespace[2pt]
W82 & .021 & .761 & .088 & .021 & .772 & .081 & .008 & .855 & .052 \\
    & \sd{.004} & \sd{.016} & \sd{.007} & \sd{.002} & \sd{.006} & \sd{.002} & \sd{.002} & \sd{.012} & \sd{.005} \\ \addlinespace[2pt]
W92 & .026 & .800 & .089 & .031 & .780 & .101 & .017 & .834 & .076 \\
    & \sd{.002} & \sd{.006} & \sd{.004} & \sd{.001} & \sd{.003} & \sd{.002} & \sd{.003} & \sd{.009} & \sd{.005} \\
\midrule
\textbf{Avg} & .025 & .771 & .085 & .029 & .759 & .091 & \textbf{.014} & \textbf{.830} & \textbf{.066} \\
\bottomrule
\end{tabular}
\end{table*}

Across the three runs, sample SDs are small relative to the differences among the methods: at most .007 for MSE, .026 for 1-JSD, and .013 for EMD (Table~\ref{tab:full_metrics_14waves}). For the entropy and cost in Table~\ref{tab:baseline_comparison_14waves}, entropy SDs are at most .015 (Vanilla), .007 (PERSONA), and .025 (\algname), and cost SDs at most \$0.007 for both baselines and \$0.26 for \algname. The latter comes from W41, where one run generated 496 instead of 300 agents after repeated connection retries during generation.

Table~\ref{tab:lasso_vs_enet} compares lasso and elastic net as Stage 2 regression choices for \algname. Performance is nearly identical across all waves (average $\Delta$ MSE $\approx$ 0.0001, with 9 of 14 waves showing no difference). As lasso typically selects a more compact ensemble than elastic net, we prioritize its use as the default regression method.

\paragraph{Runtime}
Runtime is dominated by API inference. By design, \algname parallelizes both prompt generation and survey response elicitation, but the exact runtime varies by \texttt{update\_steps}, \texttt{max\_workers}, API rate limits and connection stability. For reference, a typical run on a single ATP wave (20 workers, 5 update steps, Gemini-2.0-Flash) takes about 26 mins, with regression under 2 minutes.

\begin{table}[htp]
\centering
\caption{Comparison of lasso and elastic net regression for \algname across 14 ATP waves. Values are mean $\pm$ sample SD over three runs. Both methods yield nearly identical performance, demonstrating robustness to regularization choice.}
\label{tab:lasso_vs_enet}
\small
\begin{tabular}{lcccc}
\toprule
Wave & Lasso MSE & ENet MSE & $\Delta$ & Winner \\
\midrule
W26 & .0098 $\pm$ .0002 & .0098 $\pm$ .0002 & .0000 & Tie \\
W27 & .0218 $\pm$ .0010 & .0238 $\pm$ .0027 & .0020 & Lasso \\
W29 & .0156 $\pm$ .0049 & .0156 $\pm$ .0049 & .0000 & Tie \\
W32 & .0160 $\pm$ .0019 & .0160 $\pm$ .0019 & .0000 & Tie \\
W34 & .0111 $\pm$ .0011 & .0111 $\pm$ .0011 & .0000 & Tie \\
W36 & .0191 $\pm$ .0042 & .0187 $\pm$ .0037 & .0004 & ENet \\
W41 & .0126 $\pm$ .0009 & .0126 $\pm$ .0009 & .0000 & Tie \\
W42 & .0110 $\pm$ .0003 & .0101 $\pm$ .0006 & .0009 & ENet \\
W45 & .0079 $\pm$ .0023 & .0079 $\pm$ .0023 & .0000 & Tie \\
W49 & .0211 $\pm$ .0037 & .0211 $\pm$ .0037 & .0000 & Tie \\
W50 & .0133 $\pm$ .0039 & .0133 $\pm$ .0039 & .0000 & Tie \\
W54 & .0144 $\pm$ .0018 & .0146 $\pm$ .0018 & .0002 & Lasso \\
W82 & .0083 $\pm$ .0016 & .0083 $\pm$ .0016 & .0000 & Tie \\
W92 & .0168 $\pm$ .0029 & .0173 $\pm$ .0023 & .0005 & Lasso \\
\midrule
\textbf{Avg} & \textbf{.0142} & \textbf{.0143} & .0001 & -- \\
\bottomrule
\end{tabular}
\end{table}

\subsubsection{Clustering}\label{sec:clustering}
Based on Vanilla and PERSONA, we design two variants using K-means clustering as a pre-processing step. Specifically, we first expand the initial pool of personas for Vanilla and PERSONA to 1000, computing their embeddings (via OpenAI’s text-embedding-3-large) and cluster them (k=300). We pick the personas (medoids) closest to the centroids to construct the candidate pool and then run our regression stage on the resulting pool. Table \ref{tab:clustering_baseline_results} shows results on ATP W42.

\begin{table}[htp]
\caption{Comparison of Vanilla, PERSONA and their clustering-based
variants where 1000 agents are reduced to 300 via $k$-means clustering.}
\label{tab:clustering_baseline_results}

\begin{center}
\begin{tabular}{l | cccccc}
\toprule
\textbf{Metric} 
  & \textbf{Vanilla} 
  & \textbf{Vanilla} 
  & \textbf{PERSONA} 
  & \textbf{PERSONA} 
   \\

  &  
  & \textbf{(1000$\to$300)} 
  & 
  & \textbf{(1000$\to$300)} 
   \\
\midrule
Avg.\ Entropy $\uparrow$ 
  & 0.3172 $\pm$ 0.0069 
  & 0.3161 
  & 0.2871 $\pm$ 0.00026 
  & 0.2844 
  \\

Test MSE (Lasso) $\downarrow$
  & 0.0285 $\pm$ 0.00490
  & 0.0266
  & 0.0362 $\pm$ 0.00116
  & 0.0395
   \\

Test MSE (Enet) $\downarrow$
  & 0.0286 $\pm$ 0.00476
  & 0.0266
  & 0.0339 $\pm$ 0.00124
  & 0.0387
   \\

Cost (USD) $\downarrow$
  & \textbf{0.8649 $\pm$ 0.0070}
  & 0.8833\textsuperscript{\textdagger} 
  & 1.7129 $\pm$ 0.000016\textsuperscript{*}
  & 1.7203\textsuperscript{*\textdagger}
\\
\bottomrule
\end{tabular}
\end{center}

\footnotesize
\textsuperscript{*} Only considering survey elicitation cost on 300 personas—actual cost is higher. \\
\textsuperscript{\textdagger} Excluding embedding cost.
\end{table}

Clustering yields a test MSE of 0.0266 for Vanilla that lies well within the variability band of the original Vanilla baseline (0.0285 $\pm$ 0.0049) and slightly worse MSE for PERSONA (0.0395 vs. 0.0362 $\pm$ 0.0011), with no conspicuous change on entropy and a slight increase in cost (for Vanilla due to extra generation). Hence, clustering does not yield a stronger baseline.

This matches our intuition for two reasons. First, when we apply clustering on top of the existing baselines (starting from the same 300 endowments/personas), it can only discard information that the regression step could have used. In other words, it weakens the candidate pool without giving L1 regression any new capability. Second, even in the more lenient setting where we give clustering a head start by generating 1000 candidates and then compressing them to 300, the mechanism still does not produce a stronger baseline. As clustering operates in the embedding space and tends to group personas by textual or stylistic similarity, it does not necessarily track preference-relevant variation in survey responses. As a result, minority or low-density preference types are prone to being averaged away in embedding clustering, whereas our regression layer selects proxies directly based on their contribution to matching the observed response distributions. This corroborates the design of our main baselines.

\subsubsection{Raw Model Performance}\label{sec:backendbaseline}
To better contextualize the gain of the ensemble methods (Vanilla, PERSONA, and \algname), we provide the raw model backend performance as a reference baseline. To construct the baseline, we directly query the model backend (gemini-2.0-flash) for answers to the survey questions, using the following instruction:
\begin{quote}
    \noindent\texttt{You are completing a public opinion survey. Answer each question as an average adult respondent in the US population, based only on the question text and options. Respond with only the final answer string, not the code or label in brackets. Do not include any reasoning, explanation, or commentary. Do not preface your answer with phrases like 'I would choose'. Just return the answer text exactly as it appears in the options.}
\end{quote}
We draw 300 Monte-Carlo samples per question with temperature set to 0.7. This results in an empirical answer distribution per question. On ATP W42, this baseline attains test MSE 0.0918 and average entropy 0.0161, conspicuously worse than the results for the ensemble baselines, confirming the gains via ensemble methods (Vanilla, PERSONA, and \algname).

\subsection{Locale Generalization: World Values Survey}\label{app:localegen}

The main paper results focus on the American Trends Panel (ATP), which represents a single survey program targeting the US population. While our 14-wave study demonstrates that \algname generalizes across multiple topics and different cross sections, it remains intriguing how it performs at other locales and on different survey programs. To address this question, we evaluate \algname on the World Values Survey (WVS) Wave 7 \cite{wvs_round7_v6_2022}, using cleaned data from \citet{cao-etal-2025-specializing}. We select three locales representing different cultural and linguistic contexts: \textbf{United States}, as it provides a bridge to our ATP findings, enabling direct comparison within the same population under a different survey program; \textbf{Great Britain}, which represents Western Europe, sharing linguistic and value overlap with the US but differing in political and cultural contexts; and \textbf{Hong Kong SAR}, a multicultural hub in East Asia, allowing us to test on populations less represented in typical LLM training data.

\subsubsection{Experiment Setup}
For the WVS experiment, we use a similar setup to the ATP study: survey questions are split randomly according to a $7:1.5:1.5$ ratio into the training, validation, and test sets; endowment budget is set at 300; initial sampling generates $10$ endowments for each mode; active endowment generation is run in 5 update steps with the preset attribute bank, question patching (lowest 3; 75\% endowment budget); 10 attributes drawn for each endowment generation; Gemini-2.0-flash is used as the backend. Different from the ATP study, we additionally pass the locale information (US, GB, HK) to the endowment model to ensure regional relevance of generated personas.

Notably, our random split protocol differs from \citet{cao-etal-2025-specializing}'s Q1$+$Q2$\rightarrow$Q3 topic shift protocol (Appendix~\ref{sec:stresstest}). This is intentional, as it isolates the effect of locale from confounding topic shift effects, providing a cleaner assessment of cross-locale generalizability.

Table \ref{tab:wvs_w7_p2p_performance} reports results across the three locales. Each experiment is repeated 3 times.

\subsubsection{Results and Discussion}

\begin{table}[htp]
\caption{\algname performance on WVS Wave 7 (US, Great Britain, Hong Kong SAR) under a 7:1.5:1.5 question split, using gemini-2.0-flash as the backend. Reported are mean $\pm$ sample SD over three repeated runs.}
\label{tab:wvs_w7_p2p_performance}

\begin{center}
\begin{tabular}{l | ccc}
\toprule
\textbf{Metric} 
  & \textbf{WVS W7 US}
  & \textbf{WVS W7 GB}
  & \textbf{WVS W7 HK} \\
\midrule
Avg.\ Entropy $\uparrow$
  & 0.5720 $\pm$ 0.0149
  & 0.5710 $\pm$ 0.0044
  & 0.5478 $\pm$ 0.0211 \\

Cost (USD) $\downarrow$
  & 1.7551 $\pm$ 0.0097
  & 1.7095 $\pm$ 0.0031
  & 1.7365 $\pm$ 0.0080 \\
\midrule

Test MSE (Lasso) $\downarrow$
  & 0.0149 $\pm$ 0.0005
  & 0.0141 $\pm$ 0.0014
  & 0.0242 $\pm$ 0.0013 \\

Test 1-JSD (Lasso) $\uparrow$
  & 0.8021 $\pm$ 0.0042
  & 0.7892 $\pm$ 0.0052
  & 0.7532 $\pm$ 0.0056 \\

Test MCPA (Lasso) $\uparrow$
  & 0.6667 $\pm$ 0.0152
  & 0.6481 $\pm$ 0.0321
  & 0.5789 $\pm$ 0.0263 \\

Test EMD (Lasso) $\downarrow$
  & 0.1073 $\pm$ 0.0032
  & 0.1032 $\pm$ 0.0072
  & 0.1423 $\pm$ 0.0035 \\
\midrule

Test MSE (Elastic Net) $\downarrow$
  & 0.0147 $\pm$ 0.0005
  & 0.0137 $\pm$ 0.0010
  & 0.0244 $\pm$ 0.0005 \\

Test 1-JSD (Elastic Net) $\uparrow$
  & 0.8056 $\pm$ 0.0023
  & 0.7950 $\pm$ 0.0073
  & 0.7555 $\pm$ 0.0051 \\

Test MCPA (Elastic Net) $\uparrow$
  & 0.6930 $\pm$ 0.0152
  & 0.6296 $\pm$ 0.0321
  & 0.5877 $\pm$ 0.0304 \\

Test EMD (Elastic Net) $\downarrow$
  & 0.1064 $\pm$ 0.0022
  & 0.1004 $\pm$ 0.0054
  & 0.1431 $\pm$ 0.0044 \\
\bottomrule
\end{tabular}
\end{center}
\end{table}

\begin{table}[htp]
\caption{Comparison of P2P performance on ATP (US) and WVS (US). ATP values are mean $\pm$ sample SD over 14 waves; WVS values are mean $\pm$ sample SD over 3 runs.}
\label{tab:wvs_atp_comparison}

\begin{center}
\begin{tabular}{l | cc}
\toprule
\textbf{Metric} 
  & \textbf{ATP (14 waves)}
  & \textbf{WVS US} \\
\midrule
Test MSE (Lasso) $\downarrow$
  & 0.014 $\pm$ 0.004
  & 0.015 $\pm$ 0.001 \\

Avg.\ Entropy $\uparrow$
  & 0.53 $\pm$ 0.06
  & 0.57 $\pm$ 0.01 \\

Cost (USD) $\downarrow$
  & 0.8 $\pm$ 0.2
  & 1.76 $\pm$ 0.01 \\
\bottomrule
\end{tabular}
\end{center}
\end{table}

As shown in Table \ref{tab:wvs_w7_p2p_performance}, performance is higher for US (MSE $\approx$ 0.015) and GB (MSE $\approx$ 0.014) and somewhat lower for HK (MSE $\approx$ 0.024), aligning with broader findings that large language models trained primarily on English and Western-centric data tend to better match Western public opinion than non-Western populations \cite{durmus2024measuringrepresentationsubjectiveglobal}. The HK performance gap motivates the error analysis presented in Section \ref{sec:erroranalysis} of the main paper. Despite the gap, the HK results lie just above the performance range (0.008--0.022) observed across our 14-wave ATP study presented in the main paper (cf. Table \ref{tab:baseline_comparison_14waves}), suggesting that while locale affects difficulty, \algname remains broadly applicable.  Compared with ATP, WVS results register a higher per-survey cost at approximately \$1.73, due to the higher number of survey questions (244 for US and HK, 232 for GB). Notably, the US results are nearly identical across survey programs (ATP: 0.014; WVS: 0.015), providing evidence against program-specific overfitting (Table~\ref{tab:wvs_atp_comparison}).

\subsection{Ablation Studies}\label{sec:app_abl}

\subsubsection{Attributes}\label{sec:attribute}
As discussed in Section \ref{sec:attributes}, attributes play the role of control handles in \algname's generation logic, and \textit{freeform} templates allow \algname to derive attributes from a specific survey or question, complementing the preset attribute bank with data-driven insights on endowment generation. To evaluate the potency of the freeform templates, we conduct an ablation in study in which we sequentially turn off question patching, survey patching and both. We additionally test the impact of per-endowment attribute cap through a setting where the default max attribute=10 is changed to 20. 

\begin{table}[htp]
\caption{Attribute ablation. Reported are mean $\pm$ sample SD over three repeated runs.}
\label{tab:ablation_patches}

\begin{center}
\begin{tabular}{l | cccc}
\toprule
\textbf{Metric} 
  & \textbf{Avg.\ Entropy $\uparrow$} 
  & \textbf{Test MSE $\downarrow$}
  & \textbf{Test MSE $\downarrow$}
  & \textbf{Cost (USD) $\downarrow$} \\

  &  
  & \textbf{(Lasso)}
  & \textbf{(Enet)}
  &  \\
\midrule
Original 
  & \textbf{0.4451} $\pm$ .0256
  & \textbf{0.0087} $\pm$ .0026
  & \textbf{0.0082} $\pm$ .0025
  & 0.9708 $\pm$ .0036 \\

Survey patch off
  & 0.4378 $\pm$ .0134
  & 0.0104 $\pm$ .0024
  & 0.0102 $\pm$ .0017
  & 0.9785 $\pm$ .0122 \\

Question patch off
  & 0.3989 $\pm$ .0203
  & 0.0114 $\pm$ .0016
  & 0.0109 $\pm$ .0015
  & 0.9698 $\pm$ .0011 \\

Freeform off
  & 0.3953 $\pm$ .0097
  & 0.0128 $\pm$ .0008
  & 0.0123 $\pm$ .0012
  & \textbf{0.9688} $\pm$ .0015 \\

max attributes = 20
  & 0.4363 $\pm$ .0061
  & 0.0124 $\pm$ .0046
  & 0.0115 $\pm$ .0041
  & 0.9880 $\pm$ .0167 \\
\bottomrule
\end{tabular}
\end{center}
\end{table}
Results in Table \ref{tab:ablation_patches} suggest that performance of \algname is robust to the attribute cap: entropy is essentially unchanged, and the difference in test MSE falls within run-to-run variability. Turning off freeform templates (especially question patching) degrades both average entropy and test MSE without meaningful cost gain. Therefore, from a performance vantage, it is recommended to keep the freeform templates for the extra benefits of data-driven insights.

\subsubsection{Endowment Budget}

We run \algname for 9 different endowment budgets, scaled linearly from 130 to 450. We increase the update steps of active endowment generation accordingly to ensure that at each step the number of new endowments generated is fixed at 20. We also include a budget of 110 with no updates as an ablation of adaptive sampling.  Table \ref{tab:endowment_budget} shows that on W42, a higher budget generally improves test performance but gains plateau after about three updates, indicating a moderate budget suffices.

\begin{table}[htp]
\centering
\caption{Performance of \algname with varying endowment budget. The average question entropy, test MSE (Lasso and ElasticNet), and generation cost are reported. All values are mean $\pm$ sample SD over three repeated runs.}
\label{tab:endowment_budget}
\begin{subtable}{0.80\textwidth}
\centering
\caption{Budgets 110--210}
\resizebox{\linewidth}{!}{%
\begin{tabular}{l|cccc}
\toprule
Endowment Budget           & 110                 & 130                 & 170                 & 210                 \\
\midrule
Avg.\ Entropy $\uparrow$   & 0.3777 $\pm$ 0.0191 & 0.3697 $\pm$ 0.0057 & 0.4180 $\pm$ 0.0080 & 0.4341 $\pm$ 0.0043 \\
Test MSE (Lasso) $\downarrow$ & 0.0144 $\pm$ 0.0018 & 0.0167 $\pm$ 0.0041 & 0.0100 $\pm$ 0.0025 & 0.0111 $\pm$ 0.0006 \\
Test MSE (Enet) $\downarrow$  & 0.0149 $\pm$ 0.0014 & 0.0156 $\pm$ 0.0028 & 0.0103 $\pm$ 0.0026 & 0.0104 $\pm$ 0.0006 \\
Cost (USD) $\downarrow$    & \textbf{0.3506} $\pm$ 0.0033 & 0.4154 $\pm$ 0.0054 & 0.5484 $\pm$ 0.0058 & 0.6752 $\pm$ 0.0098 \\
\bottomrule
\end{tabular}}
\end{subtable}

\vspace{0.5em}

\begin{subtable}{0.99\textwidth}
\centering
\caption{Budgets 250--330}
{%
\begin{tabular}{l|ccc}
\toprule
Endowment Budget           & 250                 & 290                 & 330                 \\
\midrule
Avg.\ Entropy $\uparrow$   & 0.4179 $\pm$ 0.0165 & 0.4369 $\pm$ 0.0168 & 0.4308 $\pm$ 0.0208 \\
Test MSE (Lasso) $\downarrow$ & 0.0124 $\pm$ 0.0043 & 0.0090 $\pm$ 0.0013 & 0.0133 $\pm$ 0.0030 \\
Test MSE (Enet) $\downarrow$  & 0.0123 $\pm$ 0.0045 & 0.0097 $\pm$ 0.0017 & 0.0121 $\pm$ 0.0031 \\
Cost (USD) $\downarrow$    & 0.8043 $\pm$ 0.0032 & 0.9365 $\pm$ 0.0027 & 1.0688 $\pm$ 0.0119 \\
\bottomrule
\end{tabular}}
\end{subtable}

\vspace{0.5em}

\begin{subtable}{0.99\textwidth}
\centering
\caption{Budgets 370--450}
{%
\begin{tabular}{l|ccc}
\toprule
Endowment Budget           & 370                 & 410                 & 450                 \\
\midrule
Avg.\ Entropy $\uparrow$   & 0.4216 $\pm$ 0.0179 & \textbf{0.4540} $\pm$ 0.0142 & 0.4433 $\pm$ 0.0092 \\
Test MSE (Lasso) $\downarrow$ & 0.0140 $\pm$ 0.0037 & \textbf{0.0080} $\pm$ 0.0006 & 0.0099 $\pm$ 0.0022 \\
Test MSE (Enet) $\downarrow$  & 0.0136 $\pm$ 0.0055 & \textbf{0.0084} $\pm$ 0.0004 & 0.0096 $\pm$ 0.0019 \\
Cost (USD) $\downarrow$    & 1.2050 $\pm$ 0.0069 & 1.3295 $\pm$ 0.0034 & 1.4619 $\pm$ 0.0137 \\
\bottomrule
\end{tabular}}
\end{subtable}
\end{table}

\subsubsection{Model Backend}

We test \algname's performance on six model backends: GPT-4.1-mini, GPT-4.1-nano, Gemini-2.0-flash, Gemini-2.5-flash-lite, DeepSeek-V4-Flash and a locally hosted Qwen. Performance varies substantially, with the best average question entropy (0.471) achieved by GPT-4.1-mini (Table~\ref{tab:ablation_model}). Gemini-2.0-flash achieves the lowest test MSE (0.0087 with lasso) at about one quarter of the cost of GPT-4.1-mini. DeepSeek-V4-Flash is the cheapest backend and ranks second on entropy, but with an intermediate test MSE (0.0165 with lasso). Smaller models, like GPT-4.1-nano and Gemini-2.5-flash-lite, perform worse, consistent with our theoretical intuition that limited model capacity constrains the learnability condition (Definition~\ref{def:learnability_pop}).

\begin{table}[htp]
\caption{Comparison of \algname performance on W42 using different model backends. Reported are average question entropy, test MSE (Lasso and ElasticNet), and generation cost. For each model, we fix the endowment budget at 300. Values are means over three repeated runs, with sample SDs in parentheses.}
\label{tab:ablation_model}

\begin{center}
\begin{tabular}{lcccc}
\toprule
Model & Avg.\ Entropy & Test MSE (Lasso) & Test MSE (Enet) & Cost (USD) \\
\midrule
GPT-4.1-mini & \textbf{0.4708} & \underline{0.0090} & \underline{0.0088} & 3.8342 \\
             & (0.0168) & (0.0014) & (0.0014) & (0.0449) \\
GPT-4.1-nano  & 0.3996 & 0.0226 & 0.0213 & \underline{0.9265} \\
              & (0.0154) & (0.0042) & (0.0048) & (0.0043) \\
Qwen         & 0.1932 & 0.0209 & 0.0201 & -- \\
             & (0.0403) & (0.0035) & (0.0029) &  \\
Gemini-2.0-flash & 0.4451 & \textbf{0.0087} & \textbf{0.0082} & 0.9708 \\
             & (0.0256) & (0.0026) & (0.0025) & (0.0036) \\
Gemini-2.5-flash-lite & 0.3608 & 0.0149 & 0.0151 & 1.0225 \\
             & (0.0084) & (0.0009) & (0.0013) & (0.0100) \\
DeepSeek-V4-Flash & \underline{0.4463} & 0.0165 & 0.0157 & \textbf{0.7234} \\
             & (0.0073) & (0.0035) & (0.0051) & (0.0153) \\
\bottomrule
\end{tabular}
\end{center}
\footnotesize \textit{Note.} Qwen was run on a local server, thereby incurring no API cost.
\end{table}

\subsubsection{Regression-based aggregation}

To gauge the contribution of the regression-based aggregation module, we ablate it by replacing regression with simple averaging of agent responses, which increases the test MSE from 0.0087 to 0.0332 (Table~\ref{tab:regression}). In essence, regression not only distills the agent collection into a more compact ensemble, but it also plays a crucial role in ensuring the population preference reflected in the survey data are actually learned by the ensemble. 

\begin{table}[htp]
\caption{Comparison of \algname performance with and without the regression-based aggregation stage on ATP W42. In the ablation, agent responses from active endowment generation are combined by simple averaging, while the regression setting uses lasso. Reported are mean $\pm$ sample SD over three repeated runs.}
\label{tab:regression}

\begin{center}
\begin{tabular}{l | cc}
\toprule
Metric & Without regression & With regression \\
\midrule
Test MSE $\downarrow$ & 0.0332 $\pm$ 0.0024 & \textbf{0.0087} $\pm$ 0.0026 \\
\bottomrule
\end{tabular}
\end{center}

\end{table}

We also compare the default option-level weighting against block weighting, where each question contributes equally to the loss. Performance remains comparable under both schemes for W42 (Table~\ref{tab:block_weighting_results}).

\begin{table}[htp]
\caption{Sensitivity check of weighting schemes.
Reported are mean $\pm$ sample SD over 3 runs.}
\label{tab:block_weighting_results}

\begin{center}
\begin{tabular}{l | cccc}
\toprule
\textbf{Weighting}
  & \textbf{Train MSE $\downarrow$}
  & \textbf{Test MSE $\downarrow$}
  & \textbf{Train MSE $\downarrow$}
  & \textbf{Test MSE $\downarrow$} \\
\textbf{Scheme}
  & \textbf{(Lasso)}
  & \textbf{(Lasso)}
  & \textbf{(Enet)}
  & \textbf{(Enet)} \\
\midrule
Default
  & 0.00497 $\pm$ .00156
  & 0.00870 $\pm$ .00256
  & \textbf{0.00403 $\pm$ .00144}
  & 0.00821 $\pm$ .00249 \\

Block
  & \textbf{0.00475 $\pm$ .00232}
  & \textbf{0.00868 $\pm$ .00257}
  & 0.00442 $\pm$ .00236
  & \textbf{0.00806 $\pm$ .00278} \\
\bottomrule
\end{tabular}
\end{center}
\end{table}

\subsubsection{Stage~2 Configuration} \label{sec:stage2_config}
\algname's Stage~2 uses regression algorithms to select the agent ensemble. Users have considerable latitude in configuring the regression algorithm, including, inter alia, the choice of constraints and the inclusion of additional anchor features. Before computing the distributional metrics (1-JSD, EMD, MCPA), renormalization is performed by design to ensure that the final answers constitute valid probability distributions. To illustrate the implications of these design choices, we present an analysis and ablation study examining the effects of the mass penalty constraints and additional anchor features.

\paragraph{Mass penalty and renormalization} Recall that $j$ indexes agents. Write $\hat r_{ik}(\mathbf{w}) = \sum_{j=1}^{N_A} w_j d_{j,ik}$ for the predicted share of option $k$ of question $i$, and $m_i(\mathbf{w}) = \sum_{k=1}^{K_i}\hat r_{ik}(\mathbf{w})$ for the predicted \textit{mass} of question $i$. The constraint in Eq.~\eqref{eq:constr} requires every question's mass to equal one:
\begin{equation}
\min_{\mathbf{w}}\; L(\mathbf{w}) + \alpha \|\mathbf{w}\|_1 \quad \mathrm{s.t.} \quad m_i(\mathbf{w}) = 1 \quad \forall i \in \mathcal{Q}_\mathrm{train}, \label{eq:form_hard}
\end{equation}
where $L(\mathbf{w}) = \frac{1}{2n}\sum_{i}\sum_{k}(r_{ik} - \hat r_{ik}(\mathbf{w}))^2$ is the squared loss over the $n$ training option rows. In most cases this is equivalent to a single sum-to-one constraint on the weights $\mathbf{w}$. When every agent answers every question exactly once, $\sum_k d_{j,ik} = 1$ for all $i$ and $j$, so $m_i(\mathbf{w}) = \sum_j w_j$ for every question and the $|\mathcal{Q}_\mathrm{train}|$ constraints reduce to
\begin{equation}
\min_{\mathbf{w}}\; L(\mathbf{w}) + \alpha \|\mathbf{w}\|_1 \quad \mathrm{s.t.} \quad \textstyle\sum_{j=1}^{N_A} w_j = 1 . \label{eq:form_sum}
\end{equation}
The two constraints differ in their treatment of edge cases: e.g., (i) when an agent leaves a question unanswered while remaining in the basis, so that its answers no longer sum to one within that question, and (ii) when the fit includes an intercept (below), which contributes an option-count-dependent term to each question's mass. The default pipeline of \algname adopts column-drop cleaning, which removes an agent with a missing or invalid answer from the candidate pool before the selection step, thereby ruling out case (i). Other cleaning schemes, for example ones that retain partially answering agents, would fall into case (i). By imposing a constraint per question, \eqref{eq:form_hard} handles these edge cases more flexibly than \eqref{eq:form_sum}, but simultaneously it engenders a larger set of equality constraints that is difficult to optimize. Rather than imposing \eqref{eq:form_hard} as hard constraints, \algname relaxes it into a quadratic \textit{mass penalty}:
\begin{equation}
\min_{\mathbf{w}}\; L(\mathbf{w}) + \alpha \|\mathbf{w}\|_1 + \lambda \sum_{i \in \mathcal{Q}_\mathrm{train}} \big(m_i(\mathbf{w}) - 1\big)^2 , \qquad \lambda = 100. \label{eq:form_soft}
\end{equation}
On the main panel (Table \ref{tab:baseline_comparison_14waves}), it keeps $m_i$ between 0.911 and 1.000. Therefore, predictions are approximately but not exactly valid distributions. For the distributional metrics (1-JSD, EMD, MCPA), each question's predicted shares are clipped at zero and renormalized to sum to one before scoring, so validity does not depend on the implementation of the mass penalty. Removing the penalty altogether (Table~\ref{tab:stage2_configurations}) leaves test MSE statistically unchanged (95\% CI [$-$0.0004, $+$0.0014]) but lowers 1-JSD by 0.0075 (95\% CI [$-$0.0118, $-$0.0036]). Hence, we retain the mass penalty in the main paper, as it keeps raw predictions, on which MSE is computed, close to valid distributions, and it improves distributional fit after renormalization.

\begin{table}[htp]
\caption{Stage 2 configuration ablation on the main panel (14 ATP waves, three repeated runs each; lasso; test split; Gemini-2.0-flash). Reported are means over the 42 runs. Agents is the mean number of agents with non-zero weight.}
\label{tab:stage2_configurations}

\begin{center}
\begin{tabular}{l | ccccc}
\toprule
\textbf{Configuration}
  & \textbf{MSE $\downarrow$}
  & \textbf{1-JSD $\uparrow$}
  & \textbf{EMD $\downarrow$}
  & \textbf{MCPA $\uparrow$}
  & \textbf{Agents} \\
\midrule
\multicolumn{6}{l}{\textit{With mass penalty}} \\
\quad No uniform component (default) & 0.0142 & 0.8300 & \textbf{0.0657} & 0.6764 & 105.7 \\
\quad + Reference respondent & 0.0130 & 0.8365 & 0.0699 & 0.6790 & 51.9 \\
\quad + Intercept & \textbf{0.0128} & \textbf{0.8403} & 0.0679 & \textbf{0.6928} & 60.4 \\
\midrule
\multicolumn{6}{l}{\textit{Without mass penalty}} \\
\quad No uniform component & 0.0146 & 0.8225 & 0.0672 & 0.6749 & 43.7 \\
\quad + Reference respondent & 0.0133 & 0.8381 & 0.0686 & 0.6782 & 44.3 \\
\quad + Intercept & 0.0132 & 0.8223 & 0.0771 & 0.6814 & 40.5 \\
\bottomrule
\end{tabular}
\end{center}
\end{table}

\paragraph{Anchors: reference respondent and intercept}

The default Stage~2 uses proxy agents to reconstruct observed response patterns of the target population. The design, albeit intuitive, requires proxy agents to account for the entire target distribution—every unit of predicted mass must come from some agent, including components that no agent expresses well, such as the uniform response mass of indifferent or inattentive respondents. Including anchors may aid Stage~2 selection, for they allow the proxy agents to explain deviations from a reference distribution rather than levels, the same logic as the drift refit in Appendix~\ref{sec:temporal}. We probe two \textit{anchors} that are inspired by the uniform distribution. A \textit{reference respondent} is an extra column with $d_{0,ik} = 1/K_i$, weighted and selected like any agent. Intuitively, it represents an indifferent respondent who answers a survey question by randomly picking an option. An \textit{intercept}, in contrast, adds an unpenalized constant $b$ to every option row, $\hat r_{ik} = \sum_j w_j d_{j,ik} + b$, so that $m_i(\mathbf{w}, b) = \sum_j w_j + K_i b$. Table~\ref{tab:stage2_configurations} shows that on the main panel both anchors lower test MSE and raise 1-JSD with the mass penalty. The reference respondent also halves the selected ensemble, from 105.7 to 51.9 agents. However, neither anchor improves EMD. As no anchored configuration dominates across metrics, we keep the configuration without an anchor as the default.

\subsection{Stress Test: Comparison with an SFT-Aligned Model under Topic Shift}\label{sec:stresstest}

\algname is designed as a plug-and-play preference alignment method that can be deployed on new populations with limited-data and limited-compute regimes. In the main paper, the baseline comparisons thus focus on prompting-based ensemble methods under the same constraints. As discussed in Appendix \ref{sec:lit}, supervised finetuning (SFT) for a distribution-calibrated model offers an alternative approach under more lenient data and compute environments. An exemplar is \citet{cao-etal-2025-specializing}, where SFT is used to train a predictive model based on pooled World Values Survey (WVS) data from 46 countries, and conditioned on locale to directly predict answer distribution for a given survey question. They further test the SFT-aligned model on a distinct unseen set of WVS questions along with other cross-country and cross-dataset validation tasks.

In this regard, our work differs from \citet{cao-etal-2025-specializing} not just in terms of training regimes, but also in key methodological premises. Specifically, we regard each survey panel as a benchmark for \textit{a specific population preference on a given topic at a given moment in time}, where such preference is \textit{gaugeable} through survey responses (cf. Figure \ref{fig:preference_model}, right panel). We do \textit{not} assume that preferences on a topic domain can be reliably inferred if the training questions never probe that domain. For example, inferring political attitudes solely from non-political or religious items is a strong assumption for both humans and synthetic agents. Likewise, we do \textit{not} assume preferences can carry over from one survey program to another, particularly when the surveys are conducted at different points in time and on different subpopulations of the same locale. Studying how aligned preferences generalize across topics, survey programs and over time is itself a key direction for future work, and will require both theoretical and empirical advances.

Nevertheless, \citet{cao-etal-2025-specializing}'s setup constitutes an interesting stress test for \algname outside our studied scenarios. We present in this section a head-to-head comparison with their method on their WVS arena. We follow their protocol: Q1 and Q2 for train/validation and Q3 as test, which induces a topic shift from general attitudinal \& religious/ethical items (Q1+Q2) to political interest and culture items (Q3). In our language, this asks: if we align to a population on broad attitudinal dimensions (Q1+Q2), how well does the aligned ensemble generalize to a new unseen dimension (Q3: political-cultural questions)? As in Section \ref{app:localegen}, we focus on the three representative locales: US (overlaps ATP), Great Britain (Western Europe) and Hong Kong SAR (East Asia). For each, we filter \citet{cao-etal-2025-specializing}’s data to that locale, train P2P (backend: gemini-2.0-flash; 300 endowments) on local Q1+Q2 (train+val), and evaluate on Q3. Table \ref{tab:topic_shift_dataset_sizes} shows the training schemes for \algname and \citet{cao-etal-2025-specializing}. 

\begin{table}[htp]
\caption{Different training regimes for \algname and \cite{cao-etal-2025-specializing}.}
\label{tab:topic_shift_dataset_sizes}

\begin{center}
\begin{tabular}{l | ccc}
\toprule
\textbf{Setup} 
  & \textbf{Train+Val Entries}
  & \textbf{Test Entries}
  & \textbf{Model backend} \\ 
  & \textbf{(Q1 + Q2)}
  & \textbf{(Q3)}
  & \\
\midrule
\algname (US)
  & 185 (150 + 35)
  & 59
  & gemini-2.0-flash \\

\algname (GB)
  & 172 (140 + 32)
  & 60
  & gemini-2.0-flash \\

\algname (HK)
  & 185 (150 + 35)
  & 59
  & gemini-2.0-flash \\

Cao et al.\ (C1: 46 countries)
  & 8427 (6841 + 1586)
  & 2719
  & Llama3-8B-Instruct \\
\bottomrule
\end{tabular}
\end{center}
\end{table}

\begin{table}[htp]
\caption{\algname vs.\ \citet{cao-etal-2025-specializing}\ on WVS (Q3). Reported are mean $\pm$ sample SD over 3 repeated runs.}
\label{tab:p2p_vs_cao_wvs_q3}

\begin{center}
\begin{tabular}{l | l | cccc}
\toprule
\textbf{Setup} 
  & \textbf{Method}
  & \textbf{Test MSE $\downarrow$}
  & \textbf{Test 1-JSD $\uparrow$}
  & \textbf{Test MCPA $\uparrow$}
  & \textbf{Test EMD $\downarrow$} \\
\midrule
\algname (US) 
  & Lasso 
  & 0.0217 $\pm$ .0014 
  & 0.7473 $\pm$ .0031 
  & 0.6271 $\pm$ .0169 
  & 0.1174 $\pm$ .0051 \\

\algname (US) 
  & Enet 
  & 0.0218 $\pm$ .0019 
  & 0.7488 $\pm$ .0044 
  & 0.6384 $\pm$ .0098 
  & 0.1177 $\pm$ .0072 \\

\algname (GB) 
  & Lasso 
  & 0.0249 $\pm$ .0009 
  & 0.7416 $\pm$ .0055 
  & 0.4889 $\pm$ .0347 
  & 0.1296 $\pm$ .0045 \\

\algname (GB) 
  & Enet 
  & 0.0249 $\pm$ .0009 
  & 0.7463 $\pm$ .0067 
  & 0.5111 $\pm$ .0096 
  & 0.1294 $\pm$ .0048 \\

\algname (HK) 
  & Lasso 
  & 0.0284 $\pm$ .0023 
  & 0.7115 $\pm$ .0125 
  & 0.4972 $\pm$ .0353 
  & 0.1507 $\pm$ .0122 \\

\algname (HK) 
  & Enet 
  & 0.0290 $\pm$ .0023 
  & 0.7132 $\pm$ .0150 
  & 0.4859 $\pm$ .0427 
  & 0.1528 $\pm$ .0105 \\

Cao et al. 
  & SFT
  & -- 
  & 0.777 
  & 0.43 
  & --\textsuperscript{*} \\
\bottomrule
\end{tabular}
\end{center}

\footnotesize\textsuperscript{*} EMD not reported due to incompatible definition in \citet{cao-etal-2025-specializing}.
\end{table}

Table \ref{tab:p2p_vs_cao_wvs_q3} presents the results on the performance comparison. In summary, \algname achieves 1-JSD on Q3 comparable to \citet{cao-etal-2025-specializing}’s SFT model and exceeds their majority-class prediction accuracy (MCPA), despite using far fewer training entries per locale and no fine-tuning (Table \ref{tab:topic_shift_dataset_sizes}). Following our earlier discussion, we interpret performance on Q3 as evidence that some preference structure transfers across topics, rather than as a requirement that a preference-alignment method must perfectly extrapolate to arbitrary unseen domains.

\subsection{Subgroup Prediction: Comparison with SubPOP-FT} \label{sec:subpop}
Section~\ref{sec:erroranalysis} demonstrates that a proxy basis generated for one population can be refit to another by refitting only its Stage~2 weights. The same principle applies to subgroups within a population. An expressive basis should represent each subgroup through different aggregation weights, without generating new agents or conditioning on demographics. We compare this approach with SubPOP-FT~\citep{suh-etal-2025-language}, a model fine-tuned specifically to predict subgroup response distributions. The methods differ in supervision: SubPOP-FT is fine-tuned on subgroup responses from other survey waves and predicts the target wave without seeing its responses, whereas \algname fits its aggregation weights on the training-question subgroup marginals of the target wave (see Table~\ref{tab:sft_comparison} for a broader comparison). Therefore, they serve complementary use cases, making SubPOP-FT an external reference rather than a competitor. We evaluate both methods on ATP and WVS.

\paragraph{ATP setup} We use the OpinionQA subgroup benchmark of \citet{suh-etal-2025-language}, rebuilding its subgroup targets from the official OpinionQA respondent release, with numerical agreement to within $5 \times 10^{-9}$ across all 9,741 cells. Using the question splits of Table~\ref{tab:baseline_comparison_14waves}, we evaluate the 13 waves included in the benchmark, 22 subgroups and 208 held-out test questions. Following \citet{suh-etal-2025-language}, refusals enter as zero entries on that question without removing the corresponding agents. We compute Wasserstein distance (WD), the benchmark's metric, alongside our three distributional metrics, with 95\% intervals obtained by bootstrapping test questions (10,000 replicates). We report two agent panels with different levels of expressiveness: the gemini-2.0-flash bases of Table~\ref{tab:baseline_comparison_14waves}, and a second panel generated with GPT-4o answering the survey and learning attributes and GPT-4o-mini generating endowments.

\begin{table}[htp]
\caption{Subgroup prediction on the OpinionQA benchmark of \citet{suh-etal-2025-language} (13 ATP waves, 22 subgroups, 208 test questions, 4,572 question--subgroup cells). \algname reports mean $\pm$ sample SD over three agent bases (lasso); SubPOP-FT uses the released adaptor. All four metrics, including MSE, are computed on \algname's clipped and renormalized predictions. WD is the normalized ordinal Wasserstein distance used by \citet{suh-etal-2025-language}.}
\label{tab:subpop_opinionqa}
\begin{center}
\small
\begin{tabular}{l | cccc}
\toprule
\textbf{Configuration} & \textbf{WD $\downarrow$} & \textbf{MSE $\downarrow$} & \textbf{1-JSD $\uparrow$} & \textbf{MCPA $\uparrow$} \\
\midrule
\multicolumn{5}{l}{\textit{\algname, gemini-2.0-flash panel (Table~\ref{tab:baseline_comparison_14waves} bases)}} \\
\quad Mass penalty (default) & 0.1231 $\pm$ .0014 & 0.0253 $\pm$ .0007 & 0.8257 $\pm$ .0028 & 0.6714 $\pm$ .0010 \\
\quad + Reference respondent & 0.1127 $\pm$ .0013 & 0.0192 $\pm$ .0004 & 0.8504 $\pm$ .0016 & 0.6817 $\pm$ .0070 \\
\quad + Intercept & 0.1155 $\pm$ .0013 & 0.0204 $\pm$ .0003 & 0.8477 $\pm$ .0013 & 0.6690 $\pm$ .0024 \\
\quad No mass penalty + intercept & 0.1148 $\pm$ .0009 & 0.0187 $\pm$ .0003 & 0.8500 $\pm$ .0010 & 0.6897 $\pm$ .0054 \\
\midrule
\multicolumn{5}{l}{\textit{\algname, GPT-4o/GPT-4o-mini panel}} \\
\quad Mass penalty (default) & 0.1213 $\pm$ .0016 & 0.0269 $\pm$ .0014 & 0.8156 $\pm$ .0027 & 0.6820 $\pm$ .0137 \\
\quad + Reference respondent & 0.1100 $\pm$ .0033 & 0.0184 $\pm$ .0009 & 0.8529 $\pm$ .0034 & 0.7003 $\pm$ .0053 \\
\quad + Intercept & 0.1157 $\pm$ .0025 & 0.0217 $\pm$ .0009 & 0.8445 $\pm$ .0018 & 0.6852 $\pm$ .0141 \\
\quad No mass penalty + intercept & 0.1069 $\pm$ .0019 & 0.0169 $\pm$ .0006 & 0.8584 $\pm$ .0027 & 0.7104 $\pm$ .0056 \\
\midrule
SubPOP-FT \citep{suh-etal-2025-language} & 0.1066 & 0.0177 & 0.8663 & 0.6667 \\
\bottomrule
\end{tabular}
\end{center}
\end{table}

Table~\ref{tab:subpop_opinionqa} indicates that the default configuration yields the weakest performance on both panels, significantly underperforming SubPOP-FT on WD, MSE and 1-JSD. Adding an anchor closes most of the gap. With the reference respondent, \algname is statistically indistinguishable from SubPOP-FT on WD and MSE on both panels. On the GPT-4o/GPT-4o-mini panel, removing the mass penalty and adding an intercept matches SubPOP-FT on WD (95\% CI [$-$0.0092, $+$0.0095]), MSE and 1-JSD, and exceeds it on MCPA (95\% CI [$+$0.002, $+$0.086]).

\paragraph{WVS setup} We use the WVS US subgroups common to Waves 6 and 7, with the three gemini-2.0-flash bases of Table~\ref{tab:wvs_w7_p2p_performance}, which were generated from W7 questions. We evaluate on 16 held-out W6 questions under three calibration regimes: (i) weights refit on W6 subgroup marginals; (ii) weights fit on W7 subgroup marginals and applied to W6; and (iii) weights fit on W7 population marginals and applied to every W6 subgroup. The third regime serves as a population-level reference, using no subgroup information. SubPOP-FT is applied zero-shot.

Table~\ref{tab:subpop_wvs} shows that all \algname configurations outperform SubPOP-FT by a wide margin (test MSE 0.0088--0.0129 against 0.0249), and the other metrics agree: for the W6 refit without the mass penalty but with an intercept, 1-JSD is 0.841, EMD 0.098 and MCPA 0.644, against 0.737, 0.168 and 0.438. Notably, fitting on W7 subgroup marginals, without observing any W6 responses, retains nearly all of the performance of a direct W6 refit (0.0090 against 0.0088 with the reference respondent). Subgroup-specific weights improve on population weights in every anchored configuration, though only modestly with the reference respondent (0.0088 against 0.0092). We read this as evidence of portability within WVS rather than of unconditional superiority: SubPOP-FT is trained on ATP, and extending it to WVS may require WVS-specific fine-tuning.

\begin{table}[htp]
\caption{Subgroup prediction on WVS (US; 16 held-out W6 questions, 10 subgroups common to W6 and W7). Reported is test MSE, mean $\pm$ sample SD over the three gemini-2.0-flash bases of Table~\ref{tab:wvs_w7_p2p_performance} (lasso). SubPOP-FT is applied zero-shot.}
\label{tab:subpop_wvs}
\begin{center}
\small
\begin{tabular}{l | ccc}
\toprule
\textbf{Configuration}
  & \textbf{\begin{tabular}[c]{@{}c@{}}W6 subgroup\\ refit\end{tabular}}
  & \textbf{\begin{tabular}[c]{@{}c@{}}W7 subgroup\\ fit $\to$ W6\end{tabular}}
  & \textbf{\begin{tabular}[c]{@{}c@{}}W7 population\\ fit $\to$ W6\end{tabular}} \\
\midrule
Mass penalty (default) & 0.0127 $\pm$ .0008 & 0.0129 $\pm$ .0001 & 0.0123 $\pm$ .0008 \\
+ Reference respondent & 0.0088 $\pm$ .0010 & 0.0090 $\pm$ .0002 & 0.0092 $\pm$ .0003 \\
+ Intercept & 0.0108 $\pm$ .0006 & 0.0112 $\pm$ .0001 & 0.0116 $\pm$ .0007 \\
No mass penalty + intercept & 0.0093 $\pm$ .0010 & 0.0096 $\pm$ .0002 & 0.0101 $\pm$ .0001 \\
\midrule
SubPOP-FT (zero-shot) & \multicolumn{3}{c}{0.0249} \\
\bottomrule
\end{tabular}
\end{center}
\end{table}

\paragraph{Connection to other sections of the appendices} Three observations connect to our other ablation studies. First, subgroup prediction benefits from anchors far more than population prediction (cf.\ Appendix~\ref{sec:stage2_config}). The reference respondent has a mean weight of 0.37--0.41 on OpinionQA/ATP and 0.35--0.46 on WVS, consistent with anchors letting the agents explain each subgroup's deviation from a reference distribution. Second, treating refusals as zero entries keeps agents with missing answers in the basis, qualifying for the first edge case of Appendix~\ref{sec:stage2_config}: a question's mass then sums only over its answering agents—no longer equivalent to a single sum-to-one constraint on the weights—and the configuration without it performs best on the GPT-4o/GPT-4o-mini panel. Third, the model backend matters, echoing the backend ablation (Appendix~\ref{sec:app_abl}): with the same configuration, the GPT-4o/GPT-4o-mini panel reaches WD 0.1069 against 0.1148 for gemini-2.0-flash, though the difference is uneven across waves. Taken together, the comparison highlights two ways of exploiting survey data. When extensive subgroup-labelled data are available, supervised fine-tuning can amortize them into a specialized predictor. When target-population observations are limited but informative, \algname offers a portable alternative that reuses a response-expressive basis and updates only its aggregation weights. \citet{cao-etal-2025-specializing} remains the closer comparison for our setting, as it holds the survey program and target populations fixed and varies only the topic (Appendix~\ref{sec:stresstest}). We provide an extended comparison of the methods in Appendix~\ref{sec:related_comparison}.

\subsection{Use Case: Retrospective Imputation via Temporal Re-anchoring} \label{sec:temporal}
When a new survey introduces questions absent from earlier waves, shared questions can calibrate a synthetic-agent basis to the historical population. The calibrated agent ensemble can then supply counterfactual response distributions for the previously unasked questions, extending series that would otherwise begin only at the later wave. This enables longitudinal and regression analyses that would otherwise suffer from missing data by design. We demonstrate this use case of \algname with Waves 6 and 7 of the WVS. The proxy agent pool is constructed using the more recent WVS W7 training questions (Appendix~\ref{app:localegen}), and re-anchored to the historical WVS W6 population using a limited number of shared questions. To make this otherwise unverifiable counterfactual testable, we adopt a masked missing-question design: we conceal the observed marginals of some questions that were actually asked, treat them as historically unasked, and reveal them only for evaluation.

\paragraph{Re-anchoring schemes} We analyze two schemes requiring different levels of information: 
\begin{itemize}
    \item \textbf{Direct refit} uses only WVS W6 training marginals and predicts levels. Stage~2 is refitted to W6 as in the main pipeline.
    \item \textbf{Drift refit} uses both WVS W6 and W7 training marginals and predicts deltas. In practice, the W7 marginals enter every prediction with weight one, $\hat r^{\,\mathrm{W6}}_{ik} = r^{\mathrm{W7}}_{ik} + \sum_j w_j d_{j,ik}$, and the agent weights sum to zero, so that each prediction still sums to one.
\end{itemize}
We regard drift refit as more principled: predicting levels pressures the ensemble to reconstruct the full preference profile, including stable preferences it may fail to capture under limited data and expressiveness (where learnability, Definition~\ref{def:learnability_pop}, binds), whereas differencing frees the basis to focus on genuine context-dependent and drifting preferences. We fit the direct refit under three Stage~2 configurations (Appendix~\ref{sec:stage2_config}) and the drift refit with ridge, lasso, and elastic net penalties. As a reference, we additionally introduce a \textit{persistence} baseline, where the W7 marginals are carried back unchanged.

\paragraph{Evaluation} We further report two errors capturing distinct reconstruction settings:
\begin{itemize}
    \item \textbf{Masked-train MSE} evaluates questions whose W6 marginals are withheld from refitting but whose wording has contributed to constructing the W7 agents—the topic is covered by agent construction, but the historical marginals are unavailable.
    \item \textbf{Test MSE} evaluates the fixed W7-test questions, which contributed neither to agent construction nor regression fitting or tuning—agents constructed from an existing contemporary survey are used to reconstruct historical responses for newly introduced questions.
\end{itemize}

\paragraph{Protocol} We use the original W7 split among 125 verified shared questions (90 train, 19 validation, and 16 test). Across 10 random splits and the three cached Table~\ref{tab:wvs_w7_p2p_performance_concise} bases, we conceal 18 train questions, re-anchor using nested subsets of the remaining 72 (5--72 questions), use validation only for hyperparameter selection, and reserve test questions for final evaluation. Results are shown in Table~\ref{tab:temporal}.

\paragraph{Results} We make three observations based on Table~\ref{tab:temporal}:
\begin{itemize}
    \item \textbf{Both schemes improve overall as more W6 calibration questions are supplied.} As the calibration budget increases from 5 to 72, MSE falls by 5.4--20.6\% on test questions and by 9.9--18.9\% on masked questions across configurations, though not always monotonically.
    \item \textbf{Direct refit test MSE falls with budget}, from 0.0122 to 0.00969 (for the configuration without mass penalty and uniform component), moving towards the original W7 error on the same shared test questions (0.00868). Masked-train MSE similarly drops from 0.0175 to 0.0142. Notably, both W6 refit schemes use far fewer training questions as anchors (capped at 72) than the W7 baseline (170).
    \item \textbf{Drift refit consistently outperforms persistence at every budget.} At 72 anchors, the ridge drift refit reduces MSE by 23.3\% on masked-train questions, and 12.1\% on test questions.
\end{itemize}
These results support using a synthetic agent ensemble for historical counterfactuals, especially via drift prediction, and offer an initial assessment of question-budget scaling.

\begin{table}[htp]
\caption{W6 reconstruction via the W7 proxy basis (WVS, US; three bases $\times$ ten random splits). Reported is mean MSE (in units of $10^{-3}$) over the 30 basis--split cells, with SD in parentheses (for persistence, across the ten splits). Direct refits predict the W6 marginals with elastic net. Drift refits predict the W6$-$W7 change using W7 as the reference, with agent weights summing to zero. All fits use option-level weighting, with penalties chosen on 19 fixed validation questions. Persistence uses the W7 marginals unchanged. Bold indicates the lowest MSE per column and panel.}
\label{tab:temporal}

\begin{center}
\begin{tabular}{l | cccccc}
\toprule
 & \multicolumn{6}{c}{\textbf{Calibration anchors}} \\
\textbf{Configuration} & \textbf{5} & \textbf{10} & \textbf{20} & \textbf{40} & \textbf{60} & \textbf{72} \\
\midrule
\multicolumn{7}{c}{\textbf{Test questions (16)}} \\
\midrule
Persistence (W7 marginals) & \multicolumn{6}{c}{3.64} \\
\multicolumn{1}{l|}{\textit{Direct refit, with mass penalty}} & & & & & & \\
\quad No uniform component (default) & 11.64 & 11.56 & 11.28 & 10.93 & 11.19 & 11.01 \\[-2pt]
 & {\scriptsize (0.71)} & {\scriptsize (0.72)} & {\scriptsize (0.92)} & {\scriptsize (0.80)} & {\scriptsize (0.65)} & {\scriptsize (0.42)} \\ \addlinespace[3pt]
\quad + Reference respondent & 10.69 & 11.06 & 10.63 & 10.07 & 9.92 & 9.67 \\[-2pt]
 & {\scriptsize (1.24)} & {\scriptsize (0.78)} & {\scriptsize (1.05)} & {\scriptsize (1.17)} & {\scriptsize (1.04)} & {\scriptsize (1.15)} \\ \addlinespace[3pt]
\multicolumn{1}{l|}{\textit{Direct refit, without mass penalty}} & & & & & & \\
\quad No uniform component & 12.20 & 11.58 & 11.05 & 10.31 & 10.04 & 9.69 \\[-2pt]
 & {\scriptsize (1.66)} & {\scriptsize (1.66)} & {\scriptsize (1.10)} & {\scriptsize (0.80)} & {\scriptsize (0.75)} & {\scriptsize (0.60)} \\ \addlinespace[3pt]
\multicolumn{1}{l|}{\textit{Drift refit, W7 reference}} & & & & & & \\
\quad Ridge & \textbf{3.42} & \textbf{3.42} & \textbf{3.39} & \textbf{3.24} & \textbf{3.23} & 3.20 \\[-2pt]
 & {\scriptsize (0.21)} & {\scriptsize (0.23)} & {\scriptsize (0.15)} & {\scriptsize (0.13)} & {\scriptsize (0.12)} & {\scriptsize (0.09)} \\ \addlinespace[3pt]
\quad Lasso & 3.47 & 3.57 & 3.47 & 3.37 & 3.26 & 3.22 \\[-2pt]
 & {\scriptsize (0.18)} & {\scriptsize (0.20)} & {\scriptsize (0.27)} & {\scriptsize (0.21)} & {\scriptsize (0.15)} & {\scriptsize (0.14)} \\ \addlinespace[3pt]
\quad Elastic net & 3.44 & 3.55 & 3.43 & 3.34 & 3.26 & \textbf{3.18} \\[-2pt]
 & {\scriptsize (0.18)} & {\scriptsize (0.20)} & {\scriptsize (0.24)} & {\scriptsize (0.20)} & {\scriptsize (0.14)} & {\scriptsize (0.15)} \\
\midrule
\multicolumn{7}{c}{\textbf{Masked train questions (18)}} \\
\midrule
Persistence (W7 marginals) & \multicolumn{6}{c}{3.01 {\scriptsize (1.38)}} \\
\multicolumn{1}{l|}{\textit{Direct refit, with mass penalty}} & & & & & & \\
\quad No uniform component (default) & 16.98 & 16.84 & 15.86 & 15.48 & 15.52 & 15.31 \\[-2pt]
 & {\scriptsize (2.83)} & {\scriptsize (2.91)} & {\scriptsize (2.74)} & {\scriptsize (2.41)} & {\scriptsize (2.40)} & {\scriptsize (2.72)} \\ \addlinespace[3pt]
\quad + Reference respondent & 16.31 & 16.15 & 15.08 & 14.73 & 14.30 & 14.23 \\[-2pt]
 & {\scriptsize (2.66)} & {\scriptsize (2.59)} & {\scriptsize (2.59)} & {\scriptsize (2.52)} & {\scriptsize (2.53)} & {\scriptsize (2.80)} \\ \addlinespace[3pt]
\multicolumn{1}{l|}{\textit{Direct refit, without mass penalty}} & & & & & & \\
\quad No uniform component & 17.46 & 16.96 & 15.52 & 14.90 & 14.35 & 14.16 \\[-2pt]
 & {\scriptsize (4.03)} & {\scriptsize (3.80)} & {\scriptsize (2.52)} & {\scriptsize (2.37)} & {\scriptsize (2.33)} & {\scriptsize (2.48)} \\ \addlinespace[3pt]
\multicolumn{1}{l|}{\textit{Drift refit, W7 reference}} & & & & & & \\
\quad Ridge & 2.74 & \textbf{2.56} & \textbf{2.51} & \textbf{2.46} & \textbf{2.36} & \textbf{2.31} \\[-2pt]
 & {\scriptsize (1.14)} & {\scriptsize (1.18)} & {\scriptsize (1.19)} & {\scriptsize (1.23)} & {\scriptsize (1.20)} & {\scriptsize (1.16)} \\ \addlinespace[3pt]
\quad Lasso & 2.75 & 2.77 & 2.62 & 2.59 & 2.47 & 2.44 \\[-2pt]
 & {\scriptsize (1.14)} & {\scriptsize (1.22)} & {\scriptsize (1.24)} & {\scriptsize (1.29)} & {\scriptsize (1.24)} & {\scriptsize (1.24)} \\ \addlinespace[3pt]
\quad Elastic net & \textbf{2.69} & 2.74 & 2.61 & 2.56 & 2.45 & 2.39 \\[-2pt]
 & {\scriptsize (1.11)} & {\scriptsize (1.22)} & {\scriptsize (1.28)} & {\scriptsize (1.27)} & {\scriptsize (1.25)} & {\scriptsize (1.19)} \\
\bottomrule
\end{tabular}
\end{center}
\end{table}

\clearpage
\section{Extended Literature Review} \label{sec:lit}
\subsection{Pluralistic Alignment}

AI alignment refers to the process of ensuring that an AI system operates in accordance with human intentions and values, whether at an individual or aggregate level~\citep{ai_alignment_2023,leike2018scalableagentalignmentreward}. As AI systems are adopted by increasingly diverse users, they must be designed to recognize and address a wider range of needs. This necessitates pluralistic systems capable of capturing and representing the diversity of human values and perspectives~\citep{Sorensen_pluralistic2024}. Pluralistic alignment therefore seeks to align models not with a single gold standard, but with a diverse range of user preferences across various attributes. 

Most works on pluralistic alignment touch on three complementary ways, namely overton, steerable, and distributional pluralism, in which a single AI model or system can support diversity of views. In overton pluralism, the model aims to output a whole spectrum of reasonable responses~\citep{sorensen2024value}. Methods to achieve this include looking at alignment shifts to transform distributions toward overton coverage~\citep{lake2024distributionaltooverton} or few-shot prompting from community-specialized LLMs~\cite{feng-etal-2024-modular}. In steerable pluralism, models are typically steered to adopt or favor particular perspectives or value attributes, often through few-shot examples~\cite{feng-etal-2024-modular,adams2025steerable}. By contrast, our approach steers purely through attributes, with the goal of constructing diverse agents that embody different perspectives. In distributional pluralism, the distribution of the model over possible answers is intended to match that of some target population~\citep{Sorensen_pluralistic2024}, often achieved through model fine-tuning on human responses~\citep{cao-etal-2025-specializing, suh-etal-2025-language}.

Another way to categorize pluralistic alignment work is by the amount of data they rely on. While most approaches in pluralistic alignment involve post-training, they range from resource-intensive approaches that fine-tune on large annotated corpora to lightweight methods that operate with minimal or no additional data.
With sufficient data, specialized community models can be fine-tuned and subsequently combined using model merging techniques~\citep{yuan2024cultural}, few-shot prompting from community model outputs~\citep{feng-etal-2024-modular}, or federated averaging~\citep{Srewa_pluralLLM_2025} to scale with user diversity. If limited data are available, an alternative line of post-training approaches leverage in-context learning, where similarity-based retrieval~\citep{adams2025steerable} or group-informed retrieval~\citep{chen2024spica} is performed to select few-shot examples that guide models toward pluralistic alignment. In the absence of detailed data from individual pluralistic users, a third post-training approach is inference-time conditioning. Inference-time conditioning uses structured prompts that encode sociodemographic traits or behavioral dispositions to condition an LLM to emulate the responses of a specific agent. These agents are also known as \textit{endowments}, personas~\citep{castricato_2025_persona} or Silicon Samples~\citep{horton_homosilicus_2023,Argyle_Busby_Fulda_Gubler_Rytting_Wingate_2023} and are lightweight to deploy and data-efficient. As inference-time conditioning is cheap and fast, they enable mixture or vector-based formulations of pluralistic alignment, which combine agents via multi-objective rewards to avoid averaging out minority perspectives~\citep{chen2025pal,feng2025policyprototypingllmspluralistic}. Therefore, \algname follows this line of work for pluralistic alignment. In the first step, \algname performs inference-time conditioning to generate a sufficient set of agents that adequately span the preference space. Thereafter, in the second step, it combines agents to achieve a compact yet pluralistic set of preferences that represent the target survey population. 

Other related works on pluralistic alignment relate to the curation of datasets, which support research on pluralistic alignment~\citep{sorensen2024value, zhou2025fair}.

\subsubsection{Attributes} \label{sec:attributes_defn}
Pluralistic alignment works that use inference-time conditioning often steer agents toward specific profiles using \textit{attributes} that include moral and value dimensions~\citep{adams2025steerable}, or broad morals, values, characteristics, and perspectives~\citep{sorensen2024value}. Attributes may also refer to demographic features, that are more tangible. Examples of these attributes include age, sex, education, income, and religion~\citep{castricato_2025_persona}. \algname uses the term \textit{attributes} more broadly to encompass both moral and value dimensions and demographic features. Bundling these together allows \algname to define and condition LLM agents in a way that reflects both who they are and how they evaluate options.

\subsection{Large Language Models to Emulate Human Preferences}

Leveraging the ability of LLMs to emulate human or human-group responses with considerable fidelity, LLM alignment or personalization~\cite{tseng_persona_two_2024} has been largely focused on emulating human responses using various techniques. These include learning from feedback or reward models~\cite{bai2022traininghelpfulharmlessassistant,deepRLHF_christiano2017}, under distribution shifts~\cite{leike2018scalableagentalignmentreward,krueger2021hidden}, finetuning~\cite{tan-etal-2024-democratizing}, retrieval augmented generation~\cite{Prahlad_2025}, representation learning~\cite{ren_2024_representationlearning}, and using human-generated~\cite{horton_homosilicus_2023,castricato_2025_persona,Argyle_Busby_Fulda_Gubler_Rytting_Wingate_2023} prompts. However, when diverse responses need to be generated from many users or user groups instead of individuals, the data and time resources required for alignment grow linearly with the number of LLMs that need to be aligned. To reduce alignment costs, some works instead rely on AI-generated~\cite{Schuller_2024,Salminen_2024,ge2025scalingsyntheticdatacreation,simmons-2023-moral} prompts, or alignment to multiple user groups~\cite{mondal2025grouppreferencealignmentcustomized, zhao2024group}, even entire populations~\cite{cao-etal-2025-specializing}. However, even the best models struggle with this task~\cite{cao-etal-2025-specializing}. Furthermore, responses from LLMs have been found to have less variation than responses from real surveys~\cite{Bisbee_Clinton_Dorff_Kenkel_Larson_2024}, and LLMs have a tendency to respond in the middle category~\cite{wang_2024_answer_c}. Consequently, it is imperative to address these limitations and devise data- and time-efficient approaches that enable LLMs to more accurately and reliably serve as proxies for survey populations. Our work contributes to this goal by introducing an active learning–inspired alignment framework that dynamically constructs a diverse agent basis under resource constraints and reconstructs population-level preferences via regression-based aggregation, without relying on ground-truth demographic profiles.

\subsection{Comparison with Related Methods}\label{sec:related_comparison}
Table~\ref{tab:method_comparison} sets out how \algname differs from the methods most relevant to it. The key distinction is where the heterogeneity comes from and how it is adapted to a target population. Existing approaches either rely on externally supplied representations, such as demographic personas, retrieved perspectives or specialized community models, or learn implicit representations inside model or adapter parameters (GPO \citep{zhao2024group}, \citet{cao-etal-2025-specializing, suh-etal-2025-language}). \algname instead actively constructs the heterogeneity required by the task at hand. Stage~1 searches for a response-expressive basis that is survey-specific (optimizing endowments based on survey question entropy) but population-independent (without observing target-population response patterns). Target-population calibration responses are only used in Stage~2 to estimate the target-specific basis composition. This separation is what enables a single basis to be reused across locales (Section~\ref{sec:erroranalysis}).

The same comparison can be read as a question of where and how scarce human survey data should enter the alignment pipeline. \citet{cao-etal-2025-specializing} pool cross-country data to fine-tune a locale-conditioned predictor, and \citet{suh-etal-2025-language} use U.S. data to train a demographic specialist (SubPOP-FT). \citet{krsteski-etal-2026-valid} divide survey data between response synthesis and post-hoc correction. \algname explores a further point in this design space: it leaves the backbone model unchanged and adapts only the inference procedure. Table~\ref{tab:sft_comparison} details the two fine-tuning approaches vis-à-vis \algname. Of these, \citet{cao-etal-2025-specializing} is the closer comparison, since it holds the survey program and target populations fixed and varies only the topic. We therefore use it for the stress test in Appendix~\ref{sec:stresstest}. Additional comparisons with \citet{suh-etal-2025-language} can be found in Appendix~\ref{sec:subpop}.

\begin{table}[htp]
\caption{How \algname differs from related methods. Existing methods take heterogeneity from external representations or learn it in model parameters; \algname actively constructs it from the task at hand.}
\label{tab:method_comparison}
\centering\small
\begin{tabular}{@{}>{\raggedright\arraybackslash}p{2.3cm}>{\raggedright\arraybackslash}p{3.6cm}>{\raggedright\arraybackslash}p{3.6cm}>{\raggedright\arraybackslash}p{2.8cm}@{}}
\toprule
Method & Representation of heterogeneity & Supervision required & What is adapted \\
\midrule
GPO~\citep{zhao2024group} & In-context group preference examples & Cross-group preference data (meta-training) and target-group context examples & Preference module (meta-trained offline) \\ \addlinespace[3pt]
Modular Pluralism~\citep{feng-etal-2024-modular} & Specialized community models & Community-specific corpora & Community LMs (offline) \\ \addlinespace[3pt]
SPICA~\citep{chen2024spica} & Retrieved group perspectives/scenarios & Group-labelled scenario resources & None (in-context) \\ \addlinespace[3pt]
PERSONA~\citep{castricato_2025_persona} & Synthetic demographic personas with LLM-enriched features & Census-derived demographic information & None (in-context) \\ \addlinespace[3pt]
\citet{cao-etal-2025-specializing} & Locale-conditioned predictor & ${\sim}$8.4K pooled response entries from 46 countries & Fine-tuned model weights \\ \addlinespace[3pt]
SubPOP-FT~\citep{suh-etal-2025-language} & Subgroup-conditioned predictor & 70K subgroup--response pairs from 53 ATP waves & LoRA weights \\ 
\midrule 
\algname & Response-expressive proxy basis & Target-population marginals on calibration questions & Proxy basis (from question text) and aggregation weights \\
\bottomrule
\end{tabular}
\end{table}

\begin{table}[htp]
\caption{Comparison of \citet{suh-etal-2025-language} (SubPOP-FT), \citet{cao-etal-2025-specializing}, and \algname.}
\label{tab:sft_comparison}
\centering\small
\begin{tabular}{@{}>{\raggedright\arraybackslash}p{2.5cm}*{3}{>{\raggedright\arraybackslash}p{3.3cm}}@{}}
\toprule
 & SubPOP-FT & Cao et al. & \algname \\
\midrule
Target populations & 22 U.S.\ demographic subgroups & 46 country populations & Defined by the given survey \\ \addlinespace[3pt]
Applying to a new population & New training corpus and fine-tuning & Locale supplied as model input & Re-run the pipeline (new survey) or refit Stage~2 weights (same survey); locale conditioning not required (Section~\ref{sec:erroranalysis}) \\ \addlinespace[3pt]
Adaptation & LoRA fine-tuning; GPU & Fine-tuning; GPU & Inference-time regression; CPU \\ \addlinespace[3pt]
What varies, training to test & Fielding period, respondents, topic & Topic only & Questions only \\ \addlinespace[3pt]
Temporal drift & Not modelled & Not modelled & Weights refit per survey occasion \\
\bottomrule
\end{tabular}
\end{table}

\subsection{Data Selection and Active Learning}
To increase preference coverage while operating under budget constraints, our active endowment generation pipeline draws inspiration from machine learning paradigms, particularly active learning.

Traditional data selection methods aim to improve the data efficiency of supervised deep learning models by identifying the most informative or important training examples for generalization~\cite{Coleman2020Selection,paul2021deep}. Selection via Proxy (SVP)~\cite{Coleman2020Selection} uses a lightweight proxy model to estimate sample utility at a fraction of the cost, while GraNd and EL2N scores~\cite{paul2021deep} identify important training examples early in training.

Similarly, deep active learning applies query strategies to iteratively select the most informative samples from a large pool of data to be added to the training dataset for retraining or fine-tuning~\cite{astorga2024partially,hubotter2025efficiently,Li_deepactivelearning_2025}. Active learning strategies balance exploitation, focusing on uncertain or high-impact samples, with exploration, seeking diverse and novel samples to broaden coverage.

In our system, we apply this logic to guide both endowment generation and question patching. Specifically, low-entropy responses are treated as indicators of low coverage or model uncertainty. We refer to question patching as a boosting strategy—not in the ensemble learning sense, but to distinguish it from the main adaptive sampling process—since it targets underperforming questions by allocating additional sampling to enhance expressivity. Meanwhile, the exploitation–exploration mechanism used to sample across existing modes reflects the same active learning principle, but with a different emphasis: prioritizing high-utility regions while maintaining representational diversity. Together, these components form an active learning–inspired loop that incrementally expands coverage of the preference space while minimizing redundant agent generation.

\subsection{Connections to Econometrics}

\algname is related to three lines of econometric work. Synthetic control \cite{abadie2003economiccosts, Abadie2010syntheticcontrol} estimates the counterfactual outcome of a treated unit as a weighted combination of untreated donor units, with weights fit to reproduce the treated unit's pre-treatment outcomes and judged afterwards. In a similar spirit, \algname fits weights over proxy agents using training questions and evaluates them on held-out questions, although the agents are actively generated rather than taken from a pre-existing set. Indicator saturation \cite{santos_automatic_2008, johansen2009analysis, castle2015sis} mitigates outliers and structural breaks in a time series by including a full set of dummy variables spanning the modeled time period and then selecting among them. \algname likewise generates more candidate agents than it retains, with Stage~1 aiming to span the latent preference space and Stage~2 selecting a compact ensemble from the pool. Viewed as a basis expansion whose size can grow, \algname's ensemble approach is akin to a sieve approximation \cite{chen2007sieveestimation} to the population response distribution. We leave a formal treatment of these connections to future work.

\clearpage

\section{Scope of Generalization}\label{sec:generalization}

\paragraph{What generalization means in \algname} \algname treats every survey as a snapshot of a particular set of preferences, representing a specific population, on a given topic, at a specific moment in time. It reconstructs these preferences only as identified through the survey—it does not recover broad-sense context-free preferences. As the right panel of Figure~\ref{fig:preference_model} illustrates, any observation set is a many-to-one, imperfect gauge of a lower-dimensional preference space, where more observations sharpen identification without removing the reduction. Our vantage is representation learning rather than psychological recovery: observed responses serve as anchors for learning an agent-ensemble representation of the human preferences that gave rise to them, and validity is only asserted where it can be tested, on held-out responses. Accordingly, generalization in \algname is to unseen questions within the same survey, with the population and topic scope held fixed. This scoping is consistent with established findings in social science: survey responses are shaped by the instrument itself (e.g., wording, format,
and context)~\citep{schwarz_self-reports_1999} and instruments sharing a topic label need not probe the same construct~\citep{hulleman2010metaanalytic}. The econometrics of extrapolation makes the same point from the other side: transporting estimates to a new population requires explicit identifying assumptions~\citep{hotz2005predicting}, while extrapolation over time is fragile to shifts in the local data-generating process (LDGP)~\citep{clements1999forecasting}. The ensemble \algname constructs recovers the LDGP of a given survey-population-moment.

\paragraph{Beyond the scope} Applying the method beyond its generalization scope—by extending an agent ensemble, fit to a particular population and context, across topics, surveys, or time—calls for extra caution. We maintain healthy scepticism about methods claiming to recover general preferences, as that claim itself is a measurement claim, valid only if data supports the scope. Zero-shot transfer to a new topic, for instance, is a domain-shift problem, inherently harder than generalizing to questions within the same topic, which is why the comparison with \citet{cao-etal-2025-specializing} is framed as a stress test (Section~\ref{sec:stresstest}). For \algname, we make the distinction between the generated candidate pool (output of Stage~1, represented using endowments) and the final agent ensemble (output of Stage~2, represented using endowments and weights), and observe that the former can be used across populations and moments. Accordingly, the locale transfer (Section~\ref{sec:erroranalysis}), the subgroup prediction (Section~\ref{sec:subpop}) and the temporal re-anchoring experiment (Section~\ref{sec:temporal}) are transfer learning in the standard sense: a candidate pool generated for a specific survey is reused by refitting its weights for a new target—the familiar pattern of a frozen representation with a target-supervised head.

\clearpage
\section{Limitations and Future Work}\label{sec:futurework}

\paragraph{Question weighting and regression loss}\label{sec:weighting}

By design, {\texttt{P2P}\xspace} converts all multiple choice questions into binary questions before fitting a regression model. For example, a question with five options will turn into five questions with binary options. While this design ensures that regression can run smoothly with heterogeneous question types, it steers the model to prioritize fitting questions with more answer options, e.g., a 10-point Likert scale. Put differently, information contained in questions with fewer answer options, e.g., Yes/No, gets diluted after the binarization. Whether or not this design is desirable depends on the user's assumption of the relationship between preference signal and choice granularity. The current design implicitly posits a positive relationship between the two—i.e., the finer the scale, the better a signal on the latent preference. For other assumptions, question reweighting or format transformation need to be done before the regression step. \algname uses option-level weighting as the default, but also offers block weighting as an option.

\paragraph{Survey data in, text out}

The current implementation of {\texttt{P2P}\xspace} is limited to structured response settings: Given question and options, the aligned agent ensemble expresses its preference by offering a probability distribution over the given options. A key future research direction is to extend the use case to text completion, e.g., drafting resolutions, offering suggestions and answering freeform survey questions. How to aggregate individual agent textual responses into a representative aggregate response remains an under-explored area of research. Auction design \citep[cf.][]{auctiondesign} can be a promising aggregation algorithm in this case.

\paragraph{{\texttt{P2P}\xspace} as a benchmark for model steerability}

In this study our primary goal is understanding \textit{preference reconstruction theory} and its implementation through the two-stage alignment framework. While our focus is on the design aspects of the system, we acknowledge that the backend model's steerability is also a determinant of {\texttt{P2P}\xspace}'s empirical performance. The ablation study using different backends offers first insights on this aspect. It also indicates {\texttt{P2P}\xspace}'s potential of being transformed into a benchmark pipeline for the study of model steerability in pluralistic alignment.

\paragraph{Theoretical guarantees and finite-sample behavior}

Our theoretical formulation characterizes preference reconstruction at the level of learnability and expressive feasibility, but it does not yet offer finite-sample error bounds or convergence guarantees for the full \algname pipeline. Deriving these theoretical insights is non-trivial as proxy agents are actively constructed using a battery of adaptive sampling strategies rather than sampled from a fixed, well-specified distribution, placing the procedure outside standard statistical learning settings. Our simulation studies offer controlled evidence that reconstruction improves with proxy-pool expressiveness, and our empirical ablations show stable performance under moderate pool sizes, but these results should be interpreted as empirical validation rather than formal guarantees. Developing error bounds that relate reconstruction quality to proxy expressiveness, calibration-target question shift, finite sample size, and regularization constitutes an important direction for future work.

\paragraph{Temporal Alignment and Non-stationary Preference} Any method calibrated at a single moment risks becoming stale as attitudes and beliefs shift. This also affects \algname when considering using a fitted ensemble in future surveys. Our view is that synthetic agents should be periodically re-anchored using more recent, easier-to-collect data (e.g., shorter or cheaper surveys) and then tested on more demanding items. Architecturally, \algname is set up for this. The endowment generation and regression stages can be rerun on new waves, and one can track how learned weights and entropy patterns evolve, or when reconstruction quality begins to deteriorate. Appendix~\ref{sec:temporal} demonstrates such re-anchoring across two WVS waves. While \algname provides the machinery, systematic studies of preference shift and alignment across time and topics remain open.

\paragraph{Anchor-Centered Stage~2} The anchors of Appendix~\ref{sec:stage2_config} and the drift refit of Appendix~\ref{sec:temporal} inspire a more general redesign of Stage~2: fitting deviations from a reference distribution rather than response distributions directly. Given an anchor distribution $a_{ik}$, we can subtract it from both the targets and every agent's answers before fitting:
\begin{equation}
r_{ik} - a_{ik} = \sum_j w_j \big(d_{j,ik} - a_{ik}\big) + \varepsilon_{ik}, \text{ equivalently } \hat r_{ik} = \Big(1 - \sum_j w_j\Big)\, a_{ik} + \sum_j w_j\, d_{j,ik}. \label{eq:anchor_centered}
\end{equation}
As centered answers sum to zero within each question, any weights yield predictions that sum to one, removing the need for a mass penalty. The $\ell_1$ penalty also becomes genuine selection. In the current design, weights summing to one fix $\|\mathbf{w}\|_1$ up to the negative weight, so the penalty mainly prices negative weight. With centering, shrinking the weights shrinks the prediction toward the anchor, a valid distribution. When no agents are selected, Stage~2 returns the anchor, while each selected agent explains a deviation from it. The anchor can be chosen to match the setting: the uniform distribution, recovering the reference respondent with an unpenalized weight $1 - \sum_j w_j$; the previous wave's marginals, recovering the drift refit when $\sum_j w_j = 0$; the overall population's marginals when predicting subgroups; or the mean of the agents' answers, so that Stage~2 shrinks toward the simple average. We leave the implementation and evaluation of this design to future work.

\paragraph{Short-cut Learning and Generalization} Overfitting to survey artifacts is a key concern for \textit{any} survey-based evaluation. This concern is closely related to what the robustness literature calls \textit{shortcut learning} or \textit{Clever Hans} behavior, where models exploit superficial regularities of a dataset instead of learning the underlying concept \citep{geirhos_shortcut_2020, Lapuschkin2019CleverHans}. Our empirical evaluation across 14 ATP waves is designed to mitigate the risk that our findings hinge on idiosyncrasies of a single wave, and the additional WVS experiments further mitigate the risk of ATP-specific shortcutting. While these steps serve as initial evidence against program-specific shortcutting, they do not constitute a complete solution to the generalization problem. A full multi-survey study (e.g., across GlobalOpinionQA and other polling corpora) is beyond the scope of this paper, but \algname’s modular pipeline readily supports this kind of analysis. Fundamentally, we see this as partly a survey design challenge: if we want synthetic agents to learn stable preferences rather than survey-specific shortcuts, future work may need survey instruments that are explicitly constructed with learnability and cross-topic generalization in mind. Moreover, we suggest a semantic question module for \algname to quantify the expressiveness of survey questions, which could serve as a parallel line of work to mitigate shortcutting on the model-design side. Taken together, these directions highlight an important and promising interdisciplinary research frontier in our pursuit of alignment science.

\section{Broader Impacts and Ethical Considerations}
\label{sec:broader-impacts}

The work endeavors to lower the fixed cost of entry for preference alignment and computational social science research. By requiring only API inference, publicly available social survey datasets, and standard CPU-based regression, \algname enables researchers without access to GPU clusters, proprietary individual-level datasets, or sensitive demographic profiles to study population-level preference reconstruction. As a zero-demographic method, it also serves as a diagnostic tool for testing when finer demographic or locale conditioning is genuinely necessary for faithful emulation.

At the same time, the method relies on LLM backends that may propagate social biases, stereotypes, or culturally skewed response patterns, and some generated endowment profiles may contain opinionated or sensitive content. Our goal is methodological—studying preference reconstruction at the aggregate level rather than individual personalization—which reduces profiling risks. We see potential for positive impact on pluralistic alignment and principled use of AI in social science research, but stress that the agent ensemble is an auxiliary tool for understanding human choice patterns and should not replace humans in critical policy-making scenarios.



\end{document}